\documentclass[10pt,twocolumn,letterpaper]{article}

\usepackage[pagenumbers, algorithms]{wacv}      % To produce the REVIEW version for the algorithms track
\usepackage{multirow}

\definecolor{wacvblue}{rgb}{0.21,0.49,0.74}
\usepackage[pagebackref,breaklinks,colorlinks,allcolors=wacvblue]{hyperref}

\usepackage{xcite}
\usepackage{xr}
\makeatletter
\newcommand*{\addFileDependency}[1]{% argument=file name and extension
  \typeout{(#1)}% latexmk will find this if $recorder=0 (however, in that case, it will ignore #1 if it is a .aux or .pdf file etc and it exists! if it doesn't exist, it will appear in the list of dependents regardless)
  \@addtofilelist{#1}% if you want it to appear in \listfiles, not really necessary and latexmk doesn't use this
  \IfFileExists{#1}{}{\typeout{No file #1.}}% latexmk will find this message if #1 doesn't exist (yet)
}
\makeatother

\newcommand{\ModelName}{NoHumansRequired}

\newcommand{\DatasetName}{NoHumansRequired Dataset}
\newcommand{\DatasetNameShort}{NHR-Edit}
\newcommand{\BagelNHR}{\textsc{Bagel-NHR-Edit}}
\usepackage{algorithm}
\usepackage{algorithmic}
\usepackage{amsmath}

\usepackage{algorithm}
\usepackage{algorithmic}
\usepackage{newfloat}
\usepackage{float}
\usepackage{listings}
\usepackage{array}
\usepackage{placeins}
\usepackage{graphicx}
\usepackage{makecell}

\newcommand{\algheader}[2]{%
  \hrule\hrule%
  \vspace{0.5ex}%
  \noindent\textbf{Algorithm #1: #2}%
  \vspace{0.5ex}%
  \hrule%
  \vspace{1.5ex}%
}

\newfloat{listing}{tb}{lst}{}
\floatname{listing}{Listing}

\makeatletter
\newif\iflisting@breakable
\let\orig@listing\listing
\let\endorig@listing\endlisting

\renewenvironment{listing}[1][]{%
  \if@twocolumn
    \listing@breakabletrue
    \par\addvspace{\intextsep}%
    \begingroup
      \let\old@captype\@captype
      \def\@captype{listing}%
      \hrule \vspace{.5\baselineskip}%
  \else
    \listing@breakablefalse
    \orig@listing[#1]%
  \fi
}{%
  \iflisting@breakable
      \vspace{.5\baselineskip}\hrule%
    \endgroup
    \par\addvspace{\intextsep}%
  \else
    \endorig@listing
  \fi
}
\makeatother

\makeatletter
\newcommand{\linknum}[1]{\hyperref[#1]{\ref*{#1}}}

\makeatother

\usepackage{siunitx}
\def\wacvPaperID{2465} % *** Enter the WACV Paper ID here
\def\confName{WACV}
\def\confYear{2026}

\title{\ModelName: Autonomous High-Quality Image Editing Triplet Mining}

\author{%
Maksim Kuprashevich \qquad Grigorii Alekseenko \qquad Irina Tolstykh \qquad Georgii Fedorov\\
Bulat Suleimanov \qquad Vladimir Dokholyan \qquad Aleksandr Gordeev\\[0.3em]
R\&D Department, SALUTEDEV \\
\url{https://riko0.github.io/No-Humans-Required/}
}

\begin{document}

\maketitle
\begin{abstract}
Recent advances in generative modeling enable image editing assistants that follow natural language instructions without additional user input. Their supervised training requires millions of \textit{triplets} $\langle$original image, instruction, edited image$\rangle$, yet mining pixel-accurate examples is hard. Each edit must affect only prompt-specified regions, preserve stylistic coherence, respect physical plausibility, and retain visual appeal. The lack of robust automated edit-quality metrics hinders reliable automation at scale. We present an automated, modular pipeline that mines high-fidelity triplets across domains, resolutions, instruction complexities, and styles. Built on public generative models and running without human intervention, our system uses a task-tuned Gemini validator to score instruction adherence and aesthetics directly, removing any need for segmentation or grounding models. Inversion and compositional bootstrapping enlarge the mined set by $\approx 2.6\times$, enabling large-scale high-fidelity training data. By automating the most repetitive annotation steps, the approach allows a new scale of training without human labeling effort. To democratize research in this resource-intensive area, we release \textbf{\DatasetNameShort}, an open dataset of 720k high-quality triplets, curated at industrial scale via millions of guided generations and validator passes, and we analyze the pipeline's stage-wise survival rates, providing a framework for estimating computational effort across different model stacks. In the largest cross-dataset evaluation, it \textbf{surpasses all public alternatives}. We also release \textbf{Bagel-NHR-Edit}, a fine-tuned Bagel model with state-of-the-art metrics. 
\end{abstract}

\section{Introduction}
Recent acceleration in generative modeling has facilitated image-editing assistants that follow natural language instructions. Creating such editors is a multi-stage process, starting with \textbf{foundational pre-training} on large, often noisy datasets (e.g.,~\citet{brooks2023instructpix2pixlearningfollowimage, yu2025anyedit, zhao2025ultraedit, wei2024omniedit, ge2024seed, hui2024hq, ye2025imgedit, zhang2023magicbrush}). This stage adapts a base text-to-image model to execute diverse edits and preserve unedited regions. Next, \textbf{initial SFT} on smaller, curated datasets elevates performance on specific tasks; ObjectDrop~\citep{burroni2024objectdrop_arxiv} and OmniPaint~\citep{shin2025omnipaint_arxiv} have shown that as few as \num{2500}-\num{3300} pairs of real photos can teach a model to remove shadows and reflections in object removal task. The third stage, \textbf{continual supervised fine-tuning (SFT) and preference optimization}~\citep{wallace2023diffusiondpo, rafailov2024ipo}, handles more complex edits and improves quality but presents a data bottleneck. It is constrained by reliance on human annotators to review millions of pixel-level edits, which is not the best use of expert attention. 

Existing large-scale data collection methods have fundamental drawbacks. Cascades of external tools, e.g., for grounding~\citep{liu2023groundingdino}, segmentation~\citep{kirillov2023sam}, and inpainting~\citep{suvorov2021lama}, create visual artifacts and can corrupt the data --- if an imperfect ``remove'' edit with inpainting artifacts is inverted into an ``add'' operation, the model may learn to use artifacts as spatial cues rather than understanding the instruction’s semantics, effectively \textbf{poisoning} the training data. Approaches like 3D rendering~\citep{cheng2024aurora} lack realism and scalability, while video frame extraction~\citep{liu2025step1x} depends on complex, error-prone auxiliary models. A lack of reliable validation metrics for detecting subtle defects persists; although MLLMs are now used as evaluators~\citep{wu2023viescore, wei2024omniedit, ye2025imgedit}, we found even top models like Gemini 2.5 Pro~\citep{google2024gemini25pro} insufficient, and we therefore fine-tuned a Gemini-2.0-flash~\citep{google2024gemini2flash} validator on human scoring data (\cref{sec:meth:valid}).

\begin{figure*}[!t]
    \centering
    \includegraphics[width=0.85\linewidth]{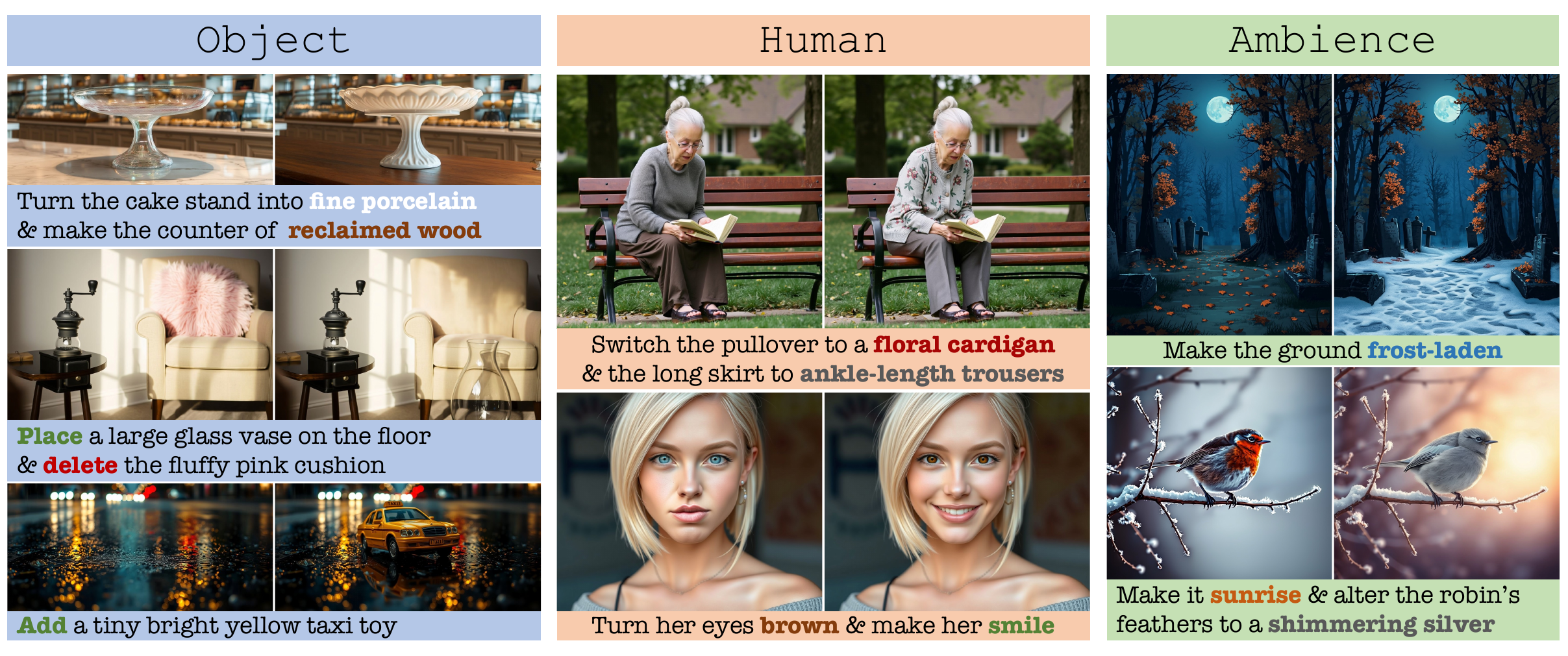}
    \caption{High-quality samples from our \textbf{\DatasetNameShort{}} dataset.}
    \label{fig:representative_results}
\end{figure*}
% \begin{figure}[!h]
%     \centering
%     \includegraphics[width=0.5\textwidth]{Figures/pipeline_final.pdf}
%     \caption{Proposed \ModelName{} framework.}
%     \label{fig:arch}
% \end{figure}

We posit that the potential of a model after initial SFT is under-exploited. By utilizing its new abilities and sensitivity to stochastic initialisation, the editor itself can generate unlimited high-quality synthetic data. To realize this, we introduce an end-to-end triplet-mining pipeline. For each instruction, the framework generates multiple candidate edits. These are pre-filtered, then judged by our fine-tuned validator, which selects the single best edit that meets our strict quality standards (\cref{algo:triplet_mining}). This self-contained framework unlocks several capabilities for continual learning:
\begin{itemize}
\item \textbf{Direct complexity measurement for curricula:} Instruction difficulty for the \textit{current} model is quantified by counting attempts for a successful edit, providing a direct signal for an \textbf{easy-to-hard learning curriculum}.
\item \textbf{Targeted weakness correction:} Rare successes on complex tasks can be mined by running the model repeatedly to harvest a targeted dataset that fixes that weakness.
\item \textbf{Compositional edit synthesis:} Complex training data can be created by combining multiple instructions. For example, a single instruction can execute two additions, one deletion, and a global style change in one pass.
\item \textbf{Flexible input sourcing:} The framework uses real and synthetic inputs. Real images provide authentic scenarios, while synthetic images enable exploration of the long-tail, including \textbf{impossible-to-photograph scenarios} (e.g., a corgi in a spacesuit on a rocket).
\item \textbf{Unparalleled simplicity and flexibility:} The framework is model-agnostic and requires no external specialist models for segmentation, depth estimation, or grounding.
\end{itemize}

To demonstrate effectiveness, we release \textsc{\DatasetName} (\textbf{\DatasetNameShort}), a public dataset of 720k rigorously validated triplets (for representative samples, see \Cref{fig:representative_results} and Figures~\ref{fig:collage1}-\ref{fig:collage8} in Appendix). Building on this data, we release \textbf{\BagelNHR{}}, a LoRA-tuned BAGEL~\citep{deng2025emerging} variant trained on \DatasetNameShort{} that surpasses the base model on two benchmarks. Our primary contribution is this end-to-end pipeline, a powerful engine for advancing research in self-improving generative models~\citep{chen2024selfplay, zhang2024selfimproving}.

\section{Related Work}

Our research builds upon two main pillars of generative modeling: methodologies for creating instruction-based editing data and the paradigm of model self-improvement through preference optimization.

\subsection{Methodologies for Editing Data Generation}
Creating high-quality editing data is a foundational challenge, with existing approaches presenting unique trade-offs.

\paragraph{Pipelines on Real-World Data.} A common strategy is a cascade of models to edit real images, like in AnyEdit~\citep{yu2025anyedit} and ImgEdit~\citep{ye2025imgedit}, which use pipelines for detection~\citep{liu2023groundingdino}, segmentation~\citep{kirillov2023sam}, and inpainting~\citep{suvorov2021lama}. Each stage can propagate errors, and global edits struggle to preserve details. Video-based methods like Step1X-Edit add complexity with pipelines for motion estimation and background filtering~\citep{zheng2024bilateralreferencehighresolutiondichotomous}. These approaches can also suffer from dataset bias (\citet{schuhmann2022laion5bopenlargescaledataset}).

\paragraph{Fully Synthetic Generation.} Synthetic generation offers more control but has its own drawbacks. Methods range from 3D rendering~\citep{cheng2024aurora}, which is labor-intensive and lacks photorealism, to diffusion-based techniques~\citep{zhao2025ultraedit, hui2024hq, ge2024seed} that can introduce artifacts, alter details, or generate data misaligned with real-world distributions.

\paragraph{Specialist Models.} OmniEdit~\citep{wei2024omniedit} trains specialized models for each task (e.g., inpainting, attribute modification) integrated into similar pipelines. While ensuring quality for simple tasks, this inherits cascade complexity and error propagation issues and cannot handle complex, compositional instructions.

Our work differs by using the editor model itself as the data source, creating a simple framework that bypasses complex pipelines and specialist models.

\subsection{The Metric Gap in Image Editing}
Evaluation is a key challenge, as traditional, reference-based metrics (e.g., LPIPS~\citep{zhang2018lpips}, DINO~\citep{ruiz2023dreamboothfinetuningtexttoimage}, CLIPScore~\citep{hessel2022clipscorereferencefreeevaluationmetric}) correlate poorly with human preference and are unsuitable for our generative framework. While MLLM-based reward models have emerged in related fields (IQA, T2I, T2V)~\citep{wu2023humanpreferencescorev2,zhang2025vqinsightteachingvlmsaigenerated,wu2025visualqualityr1reasoninginducedimagequality}, their use in editing was pioneered by VIEScore~\citep{wu2023viescore}, which showed GPT-4o judgments align well with human preferences. Subsequent work like OmniEdit and ImgEdit built on this by distilling judgments or fine-tuning MLLMs. However, curating data for SFT demands higher precision. We found that even top models like Gemini 2.5 Pro~\citep{google2024gemini25pro} are unreliable for detecting subtle editing flaws (\cref{fig:gpt4o_failure}). We therefore developed a specialized validator by fine-tuning Gemini-2.0-flash~\citep{google2024gemini2flash} on human preference data to achieve the necessary sensitivity.

\subsection{Self-improvement and Iterative Learning}
A model generating its own data for self-refinement is a highly effective concept, proven in NLP~\citep{touvron2023llama2, rafailov2024ipo} and extended to generative vision~\citep{yuan2024selfplayfinetuningdiffusionmodels}. Our framework is an automated engine applying these preference alignment techniques to image editing. Algorithms like DPO~\citep{wallace2023diffusiondpo} and KTO~\citep{bhardwaj2024diffusionkto} require scalable preference-labeled data, which our pipeline automatically provides.
By solving the data generation and labeling bottleneck, our work enables applying these powerful self-improvement techniques to instruction-based image editing.

\section{Methodology}

This section details our autonomous triplet-mining pipeline, which comprises four modules: (i) a prompt engineer for generating consistent text-to-image (T2I) and image-to-image (I2I) instructions; (ii) a T2I generator; (iii) an instruction-guided image editor; and (iv) a multi-stage validation stack.
\subsection{Automated Mining Pipeline}

\Cref{fig:arch} and \cref{algo:triplet_mining} overview the pipeline (full prompts can be found in \cref{appendix:prompts}). The process starts with initial constraints (e.g., topic, style) which are used by a prompt engineering module (\cref{algo:triplet_mining}a) to produce a T2I prompt ($p_{\mathrm{t2i}}$) and corresponding edit instructions ($\{p_e\}_k$), as shown in \cref{listings:prompt_example}. While supplied manually here, these constraints could be automated.

For each T2I prompt, the pipeline generates $N$ candidate source images ($I_0$) using different random seeds (Algorithm 1b).
Each source image undergoes $M$ edit attempts for every instruction $p_e$.
This yields a large pool of candidate triplets $\langle I_0, p_e, I_e \rangle$, which are subjected to a coarse pre-filtering step before final validation (see~\cref{sec:meth:valid}).
In the final stage, for each unique pair $\langle I_0, p_e \rangle$, the highest-quality edited image $I_e^\star$ is selected by maximizing the geometric mean of its scores ($\sqrt{s_{\text{aes}} \cdot s_{\text{adh}}}$, see \cref{algo:triplet_mining}). We chose this metric because it enforces a balance between aesthetic quality and instruction adherence, proving particularly robust for highly imbalanced scores where a candidate excels on one criterion but fails on the other. This prevents the selection of, for instance, a visually pleasing but semantically incorrect edit. The winning image is added to the final dataset $\mathcal{D}$ only if both of its scores exceed predefined quality thresholds.
\begin{listing}[!h]
\caption{Example of a generated T2I prompt and its corresponding edit instructions.}
\label{listings:prompt_example}
\begin{lstlisting}[basicstyle=\ttfamily\fontsize{9.5pt}{11.5pt}\selectfont]
\\ T2I prompt
"prompt": "A living room with a large window: a small cactus on the windowsill, a half-eaten bowl of cereal on the coffee table, a remote control, a crocheted blanket, and a dog toy on the rug.", 
\\ I2I prompts for editing
"edits": [
    "Get rid of that cactus.",
    "Remove the cereal bowl.",
    "No remote control, thanks.",
    "Lose the crocheted blanket.",
    "Eliminate the dog toy.",
    "Remove the cactus, cereal, remote, blanket, and toy"
]
\end{lstlisting}
\end{listing}

\begin{algorithm*}[!t]

\begin{minipage}{\linewidth}
    % --- LEFT COLUMN (Algorithm 1 and 3) ---
    \begin{minipage}[t]{0.47\linewidth}
        % --- Algorithm 1 ---
        \algheader{1a}{SamplePromptsDesign}
        \begin{algorithmic}[1]
            \setlength{\itemsep}{0.5ex} % Adds space between lines
            \REQUIRE Task description in $\mathcal{P}_{\ref{prompt:t2i}}$
            \ENSURE Set $\mathcal{P} = \bigl\{(p_{\mathrm{t2i}},\{p_e\}_k)\bigr\}_{m}$
            \STATE $\mathcal{P}\gets\text{OpenAI o3}\bigl(\mathcal{P}_{\ref{prompt:t2i}}\bigr)$
            \STATE \textbf{return} $\mathcal{P}$
        \end{algorithmic}

        \vspace{2.5ex} % Vertical space between the two algorithms

        % --- Algorithm 3 ---
        \algheader{1c}{Autonomous Triplet-Mining Pipeline}
        \begin{algorithmic}[1]
        \raggedright
            \setlength{\itemsep}{0.5ex} % Adds space between lines
            \REQUIRE Task description in $\mathcal{P}_{\ref{prompt:t2i}}$, parameters $N,M$, $T_{\text{aes}},T_{\text{adh}}$ 
            \ENSURE Final dataset $\mathcal{D}$
        
            \STATE $\mathcal{D}\gets\emptyset$, $\textit{Pool}\gets\emptyset$
            \STATE $\mathcal{P}\gets$\textsc{SamplePromptsDesign}$(\mathcal{P}_{\ref{prompt:t2i}})$ \COMMENT{1a}
            \FORALL{$(p_{\mathrm{t2i}},\{p_e\}_k)\in\mathcal{P}$}
                \STATE $\textit{Pool}\gets\textit{Pool}\cup$\textsc{TripletMining}$(p_{\mathrm{t2i}},\{p_e\}_k,N,M)$ \COMMENT{1b}
            \ENDFOR
        
            \FORALL{distinct $\langle I_0,p_e\rangle$ in \textit{Pool}}
                \STATE $\mathcal{S}\gets\{I_e\mid\langle I_0,p_e,I_e\rangle\in\textit{Pool}\}$
                \STATE $s_{\text{aes}}(I_e),\,s_{\text{adh}}(I_e) \gets \text{Gemini}\bigl(I_0,p_e,I_e,\mathcal{P}_{\ref{prompt:eval}}\bigr)$ \textbf{for every} $I_e\in\mathcal{S}$
                \STATE $\mathcal{S}\gets\{I_e\in\mathcal{S}\mid s_{\text{aes}} \ge T_{\text{aes}}\land s_{\text{adh}} \ge T_{\text{adh}}\}$
                \IF{$\mathcal{S}\neq\emptyset$}
                    \STATE $I_e^\star\gets\displaystyle\arg\max_{I_e\in\mathcal{S}}\sqrt{s_{\text{aes}}(I_e)\,s_{\text{adh}}(I_e)}$
                    \STATE $\mathcal{D}\gets\mathcal{D}\cup\{\langle I_0,p_e,I_e^\star\rangle\}$
                \ENDIF
            \ENDFOR
        
            \STATE $\mathcal{D}\gets\mathcal{D}\cup$\textsc{ApplyInversions}$(\mathcal{D})$~\ref{sec:meth:inv}
            \STATE $\mathcal{D}\gets$\textsc{BCFilter}$(\mathcal{D},T_{\text{inv,aes}},T_{\text{inv,adh}})$~\ref{sec:meth:back}
            \STATE $\mathcal{D}\gets\mathcal{D}\cup$\textsc{ApplyBootstraps}$(\mathcal{D})$~\ref{sec:meth:boot}
        
            \STATE \textbf{return} $\mathcal{D}$
        \end{algorithmic}
    \end{minipage}
    \hfill
    % --- RIGHT COLUMN (Algorithm 2) ---
    \begin{minipage}[t]{0.47\linewidth}
        % --- Algorithm 2 ---
        \algheader{1b}{TripletMining}
        \begin{algorithmic}[1]
        \raggedright
            \setlength{\itemsep}{0.5ex} % Adds space between lines
            \REQUIRE T2I prompt $p_{\mathrm{t2i}}$, edits $\{p_e\}_k$, parameters $N,M$, global GPU-hour budget $\texttt{Budget}$
            \ENSURE Candidate pool $\mathcal{C}$
          
            \STATE $\mathcal{C}\gets\emptyset$, $\textit{Jobs}\gets\emptyset$
            \FOR{$i\gets1$ \textbf{to} $N$}
                \STATE $\text{seed}_i \gets \text{Random}(i)$
                \STATE $I_0\gets\text{FLUX.1-schnell}(p_{\mathrm{t2i}},\text{seed}_i)$
                \IF{\textbf{not}\;Qwen$_{7\mathrm{B}}\bigl(I_0,p_{\mathrm{t2i}},\mathcal{P}_{\ref{prompt:flux_check}}\bigr)$}
                    \STATE \textbf{continue}
                \ENDIF
          
                \FORALL{$p_e\in\{p_e\}_k$}
                    \FOR{$j\gets1$ \textbf{to} $M$}
                        \STATE $\textit{Jobs}\gets\textit{Jobs}\cup\{(I_0,p_e,\text{Random}(j))\}$
                    \ENDFOR
                \ENDFOR
            \ENDFOR
          
            \WHILE{$\textit{Jobs}\neq\emptyset$ \textbf{and} $\texttt{GPU\_hours}<\texttt{Budget}$}
                \STATE \textbf{sample} $(I_0,p_e,s)\sim\text{Uniform}(\textit{Jobs})$
                \STATE $\textit{Jobs}\gets\textit{Jobs}\setminus\{(I_0,p_e,s)\}$
                \STATE $I_e\gets\text{I2I DiT (internal)}(I_0,p_e,s)$
                \STATE $(s_{\text{aes}},s_{\text{adh}})\gets \text{Qwen}_{72\mathrm{B}}\bigl(I_0,p_e,I_e,\mathcal{P}_{\ref{prompt:eval}}\bigr)$
          
                \IF{$s_{\text{aes}}\ge T_{\text{aes}} \;\textbf{and}\; s_{\text{adh}}\ge T_{\text{adh}}$}
                    \STATE $\textit{check}_p \gets \text{Qwen}_{72\mathrm{B}}(I_0,p_e,I_e,\mathcal{P}_{\ref{prompt:unw_mod}},\mathcal{P}_{\ref{prompt:vis_aes}})$
                    \STATE $\textit{check}_l \gets \text{LowLevelCheck}(I_0,I_e)$
          
                    \IF{$\textit{check}_p$ \textbf{and} $\textit{check}_l$}
                        \STATE $\mathcal{C}\gets\mathcal{C}\cup\{\langle I_0,p_e,I_e\rangle\}$
                    \ENDIF
                \ENDIF
            \ENDWHILE
            \STATE \textbf{return} $\mathcal{C}$
        \end{algorithmic}
    \end{minipage}
\end{minipage}
\caption{Pipeline Pseudocode}\label{algo:triplet_mining}
\end{algorithm*}

\subsection{Validation Framework}
\label{sec:meth:valid}
Robust validation is a key challenge in automated triplet mining. Our two-stage process uses a \textbf{Qwen-VL 72B} pre-filter to discard obvious failures, reducing calls to the more expensive final validator. While this open-source model cannot filter all noise, it is effective. The second stage uses a specialized \textbf{Gemini 2.0 Flash} model, fine-tuned on a curated corpus, to assign final aesthetic and instruction adherence scores.

\paragraph{Validator threshold.}
We set the validator thresholds using an \emph{a priori} rule grounded in the survival curve \(S(T)\) (\cref{fig:thresholds} in Appendix). The curve shows a gradual decline up to \(\approx 4.3\) and then enters a broad cliff over \(T\in[4.4,4.9]\) with pronounced drops at \(T=4.5\) (\(-62.1\%\) of the initial pool) and \(T=4.9\) (\(-84.0\%\)). To avoid operating exactly at a discontinuity while staying before the collapse regime, we choose the point that maximizes the minimum distance to these two knees. This midpoint yields \(T=4.7\). Additionally, an independent 3 raters audit of \num{1000} randomly sampled items further indicates that the \emph{residual} errors, i.e., cases where the hard-filter validator makes mistakes, as any model can --- are dispersed at high scores and frequently lie at \(\geq 4.7\); items that pass \(T=4.6\) typically receive very high scores (\(\geq 4.8\)). Consequently, raising the threshold from \(4.7\) to \(4.8\) removes almost no additional erroneous samples while shrinking the dataset. We therefore adopt the first reliable operating point before the collapse region, \(T=4.7\).
We note that an exact operating point could, in principle, be obtained only through a thorough manual audit, ideally yielding \emph{per-category} thresholds. However, such curation is labor-intensive and beyond scope. The survival-curve rule above provides a sufficient and stable choice for our application, as supported by the results in subsection \textbf{Human manual audit} and cross-dataset comparison in \cref{sec:cross_dataset_comp}.

\paragraph{Low-level check.} \label{sec:meth:low} The absolute-difference image \(D = \lvert I_e - I_0 \rvert\) is thresholded (\(>40\)) and analysed with \texttt{ConnectedComponents} using 4-connectivity and 32-bit labels; a triplet is discarded if the largest connected component covers \(<0.5 \%\) of all pixels flagged as changed. This purely heuristic, optional filter empirically outperforms a raw image-difference threshold. Cutoff level was also found during the threshold analysis of \(T\).

\label{sec:human_audit}
\paragraph{Human manual audit.} In a blinded audit of \(n=300\) accepted triplets (\cref{tab:appendix_humanaudit} in Appendix), residual issues were low: 5.0\% T2I-inherited imperfections, 4.3\% difficult removals under complex lighting or occlusion, 3.3\% small residuals after deletion, and 1.6\% minor inpainting near the edit area.

\subsection{Gemini Validator}
\label{sec:gemini}
While many pipelines use general-purpose models like \textit{GPT-4o}~\citep{hui2024hq, wei2024omniedit, wu2023viescore} for evaluation, they are not optimized for fine-grained \emph{pixel-level} changes (see \cref{fig:gpt4o_failure} in Appendix). To obtain reliable estimates, we fine-tuned a \texttt{Gemini-2.0-flash}~\citep{gemini} model on a dedicated human-annotated corpus.
This corpus was meticulously constructed to cover a wide spectrum of edit qualities, using a combination of an in-house DiT editor and proprietary models like Grok~\citep{grok_xai} and Gemini.
This diverse sourcing ensures the assessor was trained on a broad distribution of potential successes and failures, preventing overfitting.
Following HQ-Edit~\citep{hui2024hq}, OmniEdit~\citep{wei2024omniedit} and AnyEdit~\citep{yu2025anyedit},
each image is rated on two five-point scales: (i) \textbf{Instruction} score and (ii) \textbf{Aesthetics} score.
The collected set contains \num{2998} training and 827 validation examples;
every example is judged by two to four independent raters.
Inter-rater reliability, as mean pair-wise Spearman correlation, is
\(\rho=0.41\pm0.09\) for \emph{Aesthetics} and
\(\rho=0.64\pm0.05\) for \emph{Instruction},
corresponding to \emph{moderate} and \emph{substantial} agreement.
The higher consistency on the instruction axis is expected, as semantic correctness is less subjective than aesthetics.
To aggregate scores, each rating is first normalized by subtracting the annotator’s bias, computed relative to the same triplets they rated. The bias $b_j$ for each rater $j$ is
\begin{equation}
\label{eq:rater_bias}
b_j
= \underbrace{\frac{1}{|N_j|}\sum_{i \in N_j} s_{i,j}}_{\text{Rater $j$'s mean score}}
  - \underbrace{\frac{1}{|N_j|}\sum_{i \in N_j} \bar{s}_i}_{\text{Mean score of triplets rated by $j$}}
\end{equation}
where $N_j$ is the set of triplets rated by rater $j$, $R_i$ is the set of all raters for triplet $i$, and
$\bar{s}_i = \frac{1}{|R_i|}\sum_{k \in R_i} s_{i,k}$ denotes the mean score of triplet $i$.

The final score $S_i$ for a triplet is then the mean of the bias-corrected scores:
\begin{equation}
\label{eq:final_score}
S_i = \frac{1}{|R_i|} \sum_{j \in R_i} \bigl(s_{i,j} - b_j\bigr)
\end{equation}

Using this annotated validation set, we benchmarked our task-specific, fine-tuned \texttt{Gemini 2.0-flash} model against its original version, the larger \texttt{Gemini 2.5-pro}~\citep{gemini}, and \texttt{Qwen 2.5 72B}.
\Cref{tab:assessor_quality} compares the mean absolute error (MAE) and Spearman~$\rho$.
Vanilla checkpoints suffer from calibration error, whereas finetuning halves the MAE and boosts rank correlation on the instruction axis from $0.36$ to $0.82$, outperforming even the larger 2.5-pro model.
Notably, the fine-tuned model provides high-quality scores directly, without a costly chain-of-thought step, confirming a specialized assessor is a more efficient paradigm for large-scale filtering.
To further validate our assessor's robustness, we benchmarked it against the publicly available ImgEdit validator~\citep{ye2025imgedit} on a per-category basis.
Overall, our assessor nearly doubles the rank correlation (overall \(\rho = 0.79\) vs. \(0.41\)). Category-level breakdowns --- including large gains on \textit{Replace} and \textit{Compose} are provided in Appendix \cref{tab:spearman_comparison}.

\begin{table}[!htbp]
\centering
\caption{Quality metrics of the assessor model on validation data. \textit{I --- Instruction, A --- Aesthetic. }}
\begin{tabular}{lcccc}
\toprule
Model & I MAE $\downarrow$ & I $\rho$ $\uparrow$ &
A MAE $\downarrow$ & A $\rho$ $\uparrow$ \\
\midrule
Qwen 2.5 72B              & 0.961 & 0.551 & 0.839 & 0.361 \\
Gemini-2.5-pro              & 0.869 & 0.609 & 0.915 & 0.523 \\
Gemini-2.0-flash            & 1.241 & 0.359 & 1.063 & 0.245 \\
\makecell[l]{\bf Gemini-2.0-flash \\ \bf (finetune)} & \textbf{0.503} & \textbf{0.815} & \textbf{0.568} & \textbf{0.631} \\
\bottomrule
\end{tabular}
\label{tab:assessor_quality}
\end{table}

\subsection{Image Editing Backbone}
\label{sec:editor}

Our framework requires an instruction-guided image-to-image (I2I) model that takes a source image $I_0$ and prompt $p_e$ to produce an edited image $\hat I_e$. We use a proprietary, internal diffusion-based editor but treat it as a black box. This modular design ensures no component depends on the editor's internals, allowing it to be swapped with any other I2I model. The external validation stack reinforces this modularity.

\subsection{Implementation Details}
\label{sec:meth:details}

\paragraph{Component specification.}
Our pipeline is fully modular; each block can be replaced by any compatible alternative. Unless otherwise noted, we use the following defaults:

{
    \setlength{\leftmargini}{\parindent} % Mimics leftmargin=*
    \setlength{\topsep}{0pt}            % Removes space before/after the list
    \setlength{\itemsep}{0pt}           % Removes space between items
    \setlength{\parsep}{0pt}            % Removes space between paragraphs in items
    \setlength{\partopsep}{0pt}         % Removes extra top space if list follows a paragraph

    \begin{itemize}
    \item \textbf{Prompt engineer.} We query the reasoning-centric \emph{OpenAI~o3} model~\citep{openaio3} with the template~\ref{prompt:t2i} to jointly emit a text-to-image (T2I) prompt and a set of $k$ logically consistent edit instructions.
    \item \textbf{T2I generator.} Source images are synthesised with \emph{FLUX.1-schnell}~\citep{flux1schnell} at a random resolution (long side $\in[860,\num{2200}]$\,px; aspect ratio bounded by $1{:}6 \leq \mathrm{AR} \leq 6{:}1$) using 4 steps.
    \item \textbf{Plausibility gate.} We retain only sample seeds whose captions pass a plausibility check by \emph{Qwen2.5-VL-7B}~\citep{qwen25vl7b} using (Appendix, Prompt~\ref{prompt:flux_check}).
    \item \textbf{Instruction-guided editor.} By default we employ our internal I2I DiT model with 18-28 diffusion steps.
    \item \textbf{Soft pre-validation filter.} Candidate edits first pass a coarse screen with \emph{Qwen2.5-VL-72B} using (Appendix, Prompts~\ref{prompt:eval},~\ref{prompt:unw_mod},~\ref{prompt:vis_aes}).
    \item \textbf{Hard validation filter.} The fine-tuned Gemini validator (\cref{sec:meth:valid}) runs at temperature $0.0$ with (Appendix, Prompt~\ref{prompt:eval}).
    \end{itemize}
}

All Qwen-VL calls use the HuggingFace \texttt{transformers} default configuration with temperature $10^{-6}$.

\paragraph{Configuration.}
The optimal counts for T2I seeds ($N$) and edit retries ($M$) depend on prompt difficulty and represent a fundamental trade-off between dataset diversity, success rate, and computational cost. While a larger $M$ helps with harder samples by trading compute for success probability, a larger $N$ improves diversity. Our choice of $N=10$ and $M=5$ was a cost-effective balance for our specific model stack and should not be considered a universal optimum. Practitioners should tune these values based on their editor's capabilities and instruction complexity. For instance, a less capable model may require a higher $M$ to achieve a reasonable success rate. Validation thresholds are fixed at $T_{\mathrm{aes}} = T_{\mathrm{adh}} = 4.7$.

\paragraph{Budget-aware random scheduler.}
This scheduler allows practitioners to cap total expenditure. It works by enumerating all potential seed-instruction pairs ($N \times k \times M$), queuing those that pass a plausibility test, and then drawing jobs uniformly without replacement until a predefined limit is exhausted. This limit, denoted as \texttt{Budget}, is a user-specified cap in GPU-hours (or API-seconds). The final compute, quality, and dataset yield are therefore dictated by this budget, not by the nominal $(N,M)$ values. In future work, this could be extended to adaptive sampling, such as prioritizing difficult categories or continuing retries until a pre-filter success.

\subsection{Data Augmentation}
\label{sec:meth:aug}

The dataset is further refined and expanded through post-processing and augmentation.

\label{sec:meth:inv}
\paragraph{Semantic Inversion.} Any edit can be inverted by rewriting the instruction into its logical inverse using Gemini 2.5 Flash and Prompt~\ref{prompt:inverse_inst}. Crucially, access to the original T2I prompt allows preserving details for a high-quality learning signal. For the example in \cref{listings:prompt_example}, the inverse of the composite deletion is not a simple addition but a fully specified prompt: ``Add a small cactus on the windowsill, a half-eaten bowl of cereal on the coffee table, a remote control, a crocheted blanket, and a dog toy on the rug.''

\begin{figure}[h]
  \centering
  \includegraphics[width=0.98\linewidth]{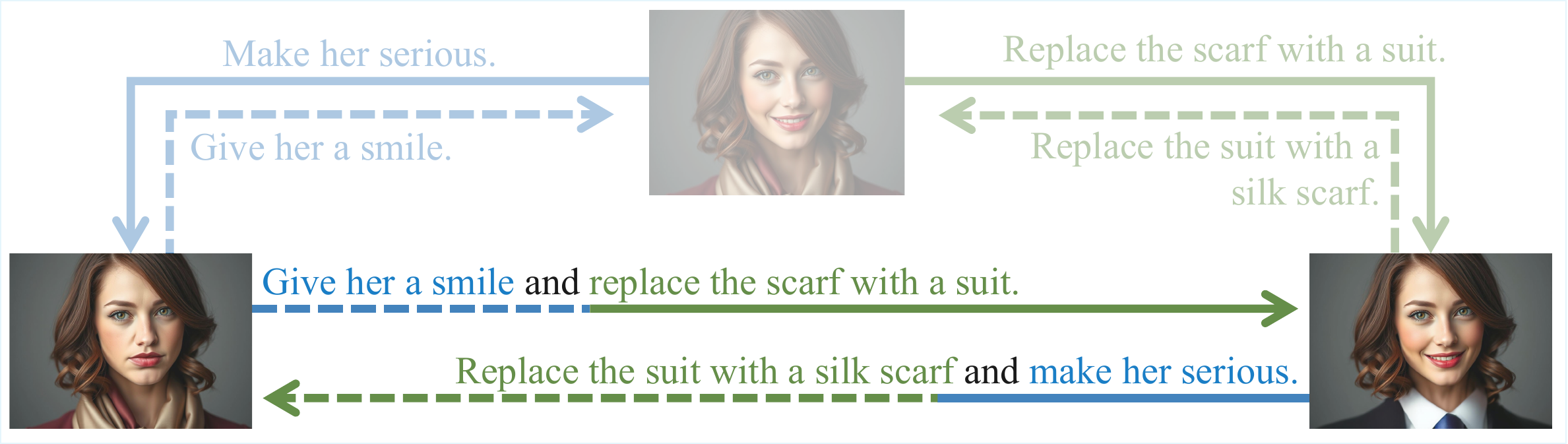}
  \caption{Solid arrows represent forward instructions, and dashed arrows represent their semantic inversions. Instructions for compositional triplets are aggregated from both forward instructions and inversions.}
  \label{fig:composite}
\end{figure}

\label{sec:meth:boot}
\paragraph{Bootstrap Composition.} Since each source image $I_0$ can be successfully edited into multiple distinct images ($I_{e1}$, $I_{e2}$, etc.), new triplets can be constructed. Given two successful edits, a new instruction $p_{e2}'$ can be formulated to transform $I_{e1}$ into $I_{e2}$, yielding a novel compositional triplet $\langle I_{e1},p_{e2}',I_{e2}\rangle$ (demonstrated in \cref{fig:composite}).
\label{sec:meth:back}
\paragraph{Backward Consistency filter.} Semantic inversion guards against trivial forward successes when the T2I misses an object. If the inverse instruction (e.g., “add the cat on the sofa”) receives a low score, we drop both the forward and inverse triplets. This optional check depends on the T2I and the validator and serves as an extra quality assurance layer.

\subsection{\DatasetName{}}
The final pipeline yields a dataset of \num{720088} high-quality triplets. \Cref{tab:filter_rates} provides a detailed breakdown of data volume changes. Initial generation and editing phases have survival rates of 44\% and 43\% respectively, with subsequent filtering further refining the set. Augmentation through inversion and composition increases the dataset size by 94.88\% and 30.65\%.

% As shown in \cref{fig:categories:overall}, \cref{fig:categories:complex} and \cref{fig:categories:composite} in Appendix, the focus is on object removal, as successful inversions provide challenging object addition examples crucial for improving modern editors. The dataset also includes a variety of other operations (\cref{fig:categories:complex} and \cref{fig:categories:composite}), obtained during bootstrapping (\cref{sec:meth:boot}). The data spans diverse styles, perspectives, and aspect ratios (\cref{fig:styles}, \cref{tab:ar_distr}), ensuring real-world diversity. This framework enables scalable triplet generation and supports \emph{weakness-targeted mining}: the pipeline can focus on simpler, invertible operations to address more complex weaknesses, functioning as a self-correcting loop without human intervention.

\DatasetNameShort{} presents a variety of editing categories, while also spanning diverse styles, perspectives, and aspect ratios:
{
    \setlength{\leftmargini}{\parindent} % Mimics leftmargin=*
    \setlength{\topsep}{0pt}            % Removes space before/after the list
    \setlength{\itemsep}{0pt}           % Removes space between items
    \setlength{\parsep}{0pt}            % Removes space between paragraphs in items
    \setlength{\partopsep}{0pt}         % Removes extra top space if list follows a paragraph

    \begin{itemize}
    \item \textbf{Removal ($\approx 227k$) and Addition ($\approx 225k$).} The focus is on object removal, as successful inversions provide challenging object addition examples, crucial for improving modern editors (\cref{fig:categories:overall}).
    \item \textbf{27 more diverse operations ($\approx 103k$).} These include complex object manipulations (reshape, change color or texture, degrade and restore), ambience (change background, time of day, weather, season), and human-related editing (emotion, haircut, clothes, accessories) --- see \cref{fig:categories:complex}.
    \item \textbf{Almost 300 composite categories ($\approx 165k$).} Bootstrap composition (\cref{sec:meth:boot}) allows the construction of multi-operation editing triplets, invaluable as complex training data (\cref{fig:categories:composite}).
    \item \textbf{96 various styles.} Spanning from photographic compositions (e.g., DSLR, panorama, wide-angle, aerial) --- to specific artistic choices (oil painting, sketch, anime, crochet, minimalist, etc.) (\cref{fig:styles}).
    \item \textbf{26 aspect ratios.} From \(640 \times 1600\) portraits to \(1600 \times 640\) panoramas. Every image is a well-established composition, generated and edited in its native aspect ratio. The distribution and samples are shown in \cref{tab:ar_distr}.
    \end{itemize}
}

\begin{table*}[!h]
\centering
\caption{Each stage statistics for \num{63292} prompts. Taking \num{3072385} generation attempts, the survival rate can be estimated as $15.3\%$, excluding the squeezing step.}
\begin{tabular}{llll}
\toprule
\textbf{Processing Stage} & \textbf{Method / Model} & \textbf{$\Delta$ (\%)} & \textbf{Remaining Vol.} \\
\midrule
Initial Generation & FLUX.1-schnell & --- & \num{1171773} \\
Generation Filtering & Qwen-7B & $-56.00$ & \num{515584} \\
Editing Generation & In-house DiT & $+495.90$ & \num{3072385} \\
Editing Filtering & Qwen-72B (Pre-Filter) & $-57.00$ & \num{1321126} \\
Low Level Check & Connected Component Analysis & $-3.00$ & \num{1281492} \\
Quality Scoring & Gemini Validator (Hard Filter) & $-63.21$ & \num{471523} \\
Final Selection & ArgMax Selection & $-31.01$ & \num{325287} \\
\midrule
Inversion & Gemini 2.5 Flash & $+94.88$ & \num{633904} \\
Composition & Bootstrap \& Concatenation & $+30.65$ & \num{828212} \\
Backward Consistency Filtering & Gemini Validator (Hard Filter) & $-13.06$ & \num{720088} \\
\bottomrule
\end{tabular}
\label{tab:filter_rates}
\end{table*}

\begin{table*}[!h]
  \centering
  \setlength{\tabcolsep}{12pt}
  \caption{Quality metrics across editing datasets, sorted in ascending order by geometric mean. The 'Type' column indicates the generation method: \textbf{A} for Automatic and \textbf{M} for Manual. The asterisk (*) denotes a highly curated automatic dataset.}
  \begin{tabular}{p{3.0cm}lccc}
    \toprule
    \textbf{Dataset} & \textbf{Type} & 
    \textbf{Instr.} $\uparrow$ & 
    \textbf{Aesth.}  $\uparrow$ & \textbf{Geom.} $\uparrow$ \\
    \midrule
    UltraEdit        & A    & 2.67 & 3.30 & 2.92 \\
    Seed Part 2      & M    & 3.20 & 3.03 & 3.09 \\
    Seed Unsplash    & A    & 3.01 & 3.84 & 3.28 \\
    InstructPix2Pix  & A    & 3.17 & 3.58 & 3.30 \\
    MagicBrush       & A    & 3.62 & 3.27 & 3.38 \\
    AnyEdit          & A    & 3.39 & 3.64 & 3.44 \\
    HQ\mbox{-}Edit   & A    & 2.90 & 4.21 & 3.45 \\
    ImgEdit          & A    & 3.26 & 3.91 & 3.49 \\
    Seed OpenImages  & A    & 3.42 & 3.86 & 3.50 \\
    Seed Part 3      & M    & 4.06 & 4.37 & 4.13 \\
    OmniEdit         & A*   & 4.21 & 4.35 & 4.23 \\
    \midrule
    \textbf{\DatasetNameShort{}} & A & \textbf{4.56} & \textbf{4.52} & \textbf{4.53} \\
    \bottomrule
  \end{tabular}
  \label{tab:dataset_quality}
\end{table*}

\label{sec:cross_dataset_comp}
\subsection{Cross-dataset comparison.}

We compare our dataset quality against public benchmarks by using our fine-tuned assessor to score \num{5000} random samples from each. \Cref{tab:dataset_quality} reports the mean \emph{Instruction}, \emph{Aesthetics}, and (following OmniEdit) geometric mean scores. With a geometric mean of 4.53, \DatasetNameShort{} establishes a new state-of-the-art, significantly outperforming existing datasets, including those with manual curation. This validates that our automated methodology can produce a corpus whose quality is superior to existing benchmarks.

\textit{Method note.} To justify using our assessor for cross-dataset ranking, we ran a targeted human cross-check on a \emph{sentinel} panel spanning the spectrum in \cref{tab:dataset_quality}: the lowest-ranked (UltraEdit), a mid-ranked set (HQEdit), and the two highest-ranked (OmniEdit, \DatasetNameShort{}). For each dataset we sampled \(n=80\) items and obtained 3 independent crowd annotations under the same instructions as the assessor. \Cref{tab:gemini_vs_human_geom} reports dataset-level geometric means with 95\% bootstrap intervals. Across this sentinel panel, assessor and humans induce the \emph{same} ordering (UltraEdit $<$ HQEdit $<$ OmniEdit $<$ \DatasetNameShort{}), with substantial interval overlap in $3/4$ cases and both assigning the top rank to \DatasetNameShort{}. This probes potential misorderings at the bottom, middle, and top regimes and provides sufficient evidence that the assessor preserves dataset-level rank; we therefore use it to score \num{5000} samples per dataset in \cref{tab:dataset_quality}. Minor numerical differences between assessor means in \cref{tab:dataset_quality} and ~\cref{tab:gemini_vs_human_geom} arise from the \(n=80\) subsampling.

\begin{table}[t]
  \centering
  % \footnotesize
  \setlength{\tabcolsep}{4pt}
  \caption{Gemini (assessor) vs.\ Human geometric mean (Geom.), shown as mean $\pm$ half-width of the 95\% nonparametric bootstrap CI ($B=\num{2000}$) over $n=80$ items per dataset (3 raters/item), recomputing $Geom.$ per resample.}
  \begin{tabular}{lcc}
    \toprule
    \textbf{Dataset} & \textbf{Gemini Geom.} $\uparrow$ & \textbf{Human Geom.} $\uparrow$ \\
    \midrule
    UltraEdit   & 3.00 $\pm$ 0.14 & 3.05 $\pm$ 0.15 \\
    HQEdit      & 3.52 $\pm$ 0.15 & 3.54 $\pm$ 0.15 \\
    OmniEdit    & 4.30 $\pm$ 0.16 & 4.50 $\pm$ 0.15 \\
    \midrule
    \DatasetNameShort{} & \textbf{4.54} $\pm$ \textbf{0.12} & \textbf{4.75} $\pm$ \textbf{0.09} \\
    \bottomrule
  \end{tabular}
  \label{tab:gemini_vs_human_geom}
\end{table}

\begin{table*}[!h]
  \centering
  \caption{Overall results comparing our \BagelNHR{} with the baseline. We report mean \(\pm\) standard deviation and [\(95\%\) confidence intervals] computed from 3 inference runs using different random seeds. The best results based on the mean are in \textbf{bold}. Per-category breakdowns appear in Appendix \cref{tab:imgedit} and \cref{tab:gedit}.}
  \begin{tabular}{lccc}
    \toprule
    \textbf{Benchmark} & \textbf{Metric(s)} & \textbf{BAGEL} & \textbf{\BagelNHR{}} \\
    \midrule
    ImgEdit-Bench & Overall & 3.30 \(\pm\) 0.03 [3.23, 3.36] & \textbf{3.33 \(\pm\) 0.02 [3.28, 3.38]} \\
    \midrule
    % GEdit-Bench   & SC / PQ / O & 7.61 \(\pm\) 0.15 [7.23, 7.98] / & \textbf{7.80 \(\pm\) 0.07 [7.63, 7.97]} / \\
    %               &             & 6.18 \(\pm\) 0.15 [5.82, 6.55] / & \textbf{6.56 \(\pm\) 0.08 [6.37, 6.75]} / \\
    %               &             & 6.53 \(\pm\) 0.14 [6.19, 6.87]  & \textbf{6.80 \(\pm\) 0.07 [6.63, 6.98]} \\
    GEdit-Bench   & SC & 7.61 \(\pm\) 0.15 [7.23, 7.98] & \textbf{7.80 \(\pm\) 0.07 [7.63, 7.97]} \\
                  & PQ & 6.18 \(\pm\) 0.15 [5.82, 6.55]
    & \textbf{6.56 \(\pm\) 0.08 [6.37, 6.75]} \\
                  & O & 6.53 \(\pm\) 0.14 [6.19, 6.87]  
    & \textbf{6.80 \(\pm\) 0.07 [6.63, 6.98]} \\
    \bottomrule
  \end{tabular}
  \label{tab:main_overall}
\end{table*}

\section{Experiments}\label{sec:exp}

This section investigates if \DatasetName\ can improve an existing edit method's performance.

\subsection{Experimental Setup}

We use BAGEL~\citep{deng2025emerging}, a 14B-parameter open-source multimodal foundation model with a Mixture-of-Transformer-Experts architecture. We performed parameter-efficient adaptation only to the generation expert’s attention and feed-forward projection layers using LoRA~\citep{hu2021lora} (rank = 16, alpha = 16, dropout = 0.05, bias = ``none'', batch size = 16 (it is dynamic, on average 2 per gpu), lr = 2e-5). We refer to this fine-tuned variant as \BagelNHR{}. Other BAGEL components are frozen to preserve the model’s pretrained capabilities. We chose LoRA for its training stability and substantially lower computational cost compared to full fine-tuning. All BAGEL and \BagelNHR{} runs use matched batch size, optimizer, learning rate schedule, precision, and data augmentations.

\subsection{Benchmarks and Metrics}
We evaluate \BagelNHR{} against the BAGEL baseline on GEdit-Bench~\citep{liu2025step1x} and ImgEdit-Bench~\citep{ye2025imgedit}, \emph{strictly following the authors’ official evaluation protocols}. For \textbf{GEdit-Bench}, we use the VIEScore setup with GPT-4o~\citep{gpt4} to report Semantic Consistency (\emph{SC}, 0-10), Perceptual Quality (\emph{PQ}, 0-10), and Overall (\emph{O}). For the \textbf{ImgEdit-Bench} evaluation, we adopt the original authors’ protocol: GPT-4o is used to score edited images across several criteria, each rated on a 1-to-5 scale.

\subsection{Results}

\Cref{tab:main_overall} reports mean, standard deviation, and \(95\%\) confidence intervals calculated from 3 inference runs with different seeds for each model. \BagelNHR{} improves over the baseline on the mean scores for both benchmarks: on \emph{ImgEdit-Bench}, the overall score increases from 3.30 to \(\mathbf{3.33}\) \((+0.03)\); on \emph{GEdit-Bench}, the SC/PQ/O scores improve from \(7.61/6.18/6.53\) to \(\mathbf{7.80/6.56/6.80}\), with deltas of \((\Delta {+}0.19/{+}0.38/{+}0.27)\) respectively. Detailed per-category results are in Appendix ~\cref{tab:imgedit} and \cref{tab:gedit}.

\section{Conclusion}

We propose an automated end-to-end pipeline to mine high-quality triplets for instruction-guided image editing. A pretrained editor generates candidate edits and we retain only successful ones after strict filtering. Instruction inversion and compositional editing produce semantically rich, diverse triplets. Integrating a T2I model broadens stylistic coverage and mitigates overfitting. The pipeline is self-improving: as the editor advances it yields better triplets, creating a feedback loop. We release \BagelNHR{}, a LoRA-tuned BAGEL variant that outperforms its baseline on public benchmarks, and \texttt{\DatasetNameShort{}} to support future research in text-based editing.

\subsubsection*{Limitations}

Our framework is bounded by its component models: it cannot produce triplets for operations the base editor cannot perform, a limitation only partly mitigated by multi-seed sampling. Data quality also depends on the T2I generator and instruction LLM, which can introduce biases from templates or priors. LLM-written instructions may diverge from real user phrasing, though diverse prompting reduces this gap.

Reporting absolute GPU-hours would be misleading as costs depend on chosen models and API pricing. Instead, we provide stage-wise survival rates in \cref{tab:filter_rates} to help estimate required generations and costs for a given model stack.

\textbf{Ethics \& Societal Impact.} \DatasetNameShort{} contains only \emph{synthetic} images generated with FLUX.1-schnell from ChatGPT o3 prompts; no photographs of real people are used, so consent/privacy risks tied to real-person imagery are not implicated (though incidental resemblance is possible). We rely on provider safeguards and automated post-filters to reduce NSFW or biased samples, but filtering is imperfect and no manual curation was performed, so some undesirable cases may remain. Because editing models can be misused, the dataset is released for research use only. Prompt diversity was encouraged, yet representation biases may persist; downstream users should assess content, apply safety filters, and comply with applicable laws and policies before deployment.

{
    \small
    \bibliographystyle{ieeenat_fullname}
    \bibliography{main}
}

\FloatBarrier % (из placeins) вытащить все накопившиеся флоты
\clearpage
\appendix
\counterwithin{figure}{section}
\counterwithin{table}{section}
\counterwithin{listing}{section}
\counterwithin{algorithm}{section}
\counterwithin{equation}{section} % если нужно

\renewcommand{\thefigure}{\thesection.\arabic{figure}}
\renewcommand{\thetable}{\thesection.\arabic{table}}
\renewcommand{\thelisting}{\thesection.\arabic{listing}}
\renewcommand{\thealgorithm}{\thesection.\arabic{algorithm}}

\onecolumn
\twocolumn[
  \centering
  {\LARGE\bfseries Supplementary \par}
  \vspace{2.5em}
]

\section{Prompts}

\label{appendix:prompts}

\begin{listing}[tb]
\caption{Samples Design Prompt}
\label{prompt:t2i}
\begin{lstlisting}
[WARN] ABSOLUTE BAN: The model must never run Python, or any other executable code, while thinking.
It must compose prompts with its own knowledge only.

---------------------------------------
1. HIGH-LEVEL PRINCIPLES
---------------------------------------
1. Natural-language first – Full phrases beat comma-separated keyword lists.
2. Specificity over brevity – Vague prompts yield "average" images; be precise.
3. One coherent vision – Avoid conflicting or scatter-shot modifiers.
4. Layered thinking – Describe foreground -> mid-ground -> background in order.
5. Active, sensory wording – "Swirls", "emerges", "diffused glow" enrich texture & motion.
---------------------------------------
2. CORE PROMPT TEMPLATE  (use as prose; brackets describe purpose)
---------------------------------------
[TECH / STYLE TAG]: [SUBJECT + ACTION], [ENVIRONMENT / CONTEXT], [COMPOSITION & CAMERA],  
[LIGHTING], [COLOUR & MOOD].
(Optional) [TEXT ELEMENTS].

Example  
DSLR photograph on Nikon Z8 with 85 mm f/1.4:  
A red fox pauses atop a snow-dusted log in a quiet boreal forest, captured at eye-level;
shallow depth-of-field isolates the fox. Soft overcast light yields gentle shadows;
a muted winter palette of whites, greys and russets conveys tranquillity.
---------------------------------------
3. DETAILED COMPONENT GUIDE
---------------------------------------
- Subject & focal point – species, character, or object with defining traits  
- Action / interaction – dynamic verb or relationship  
- Environment / setting – location, era, weather, cultural cues  
- Composition / lens – shot type, framing, spatial layout, focal length  
- Lighting – source, quality, direction, time-of-day  
- Colour palette – dominant hues, contrasts, transitions  
- Mood / atmosphere – emotional tone, sensory adjectives  
- Art / render style – medium, artist, movement  
- Technical descriptors – camera body, film stock, HDR, focus stacking, 8-K 
and related specs  
- Text integration – exact wording, font, placement, effect

---------------------------------------
4. LAYERED & SPATIAL CONTROL
---------------------------------------
Describe layers in order (foreground -> mid -> background) or label them explicitly.
Use spatial cues ("above", "to the left", "half-submerged") so FLUX can reason about position.
---------------------------------------
5. ADVANCED TECHNIQUES
---------------------------------------
- Contrast / dual aesthetics – Define clear borders & transitions (day/night split, joy/sorrow).
- See-through materials – Clarify front/behind & distortion ("rain-soaked glass distorts neon...").
- Spotlighting – Bracket clause or write "strong emphasis on ..." for key elements.
- Text-rich posters & UI – Specify font family, size, orientation; keep text short and unique.
---------------------------------------
6. DOS & DON'TS
---------------------------------------
[OK] Use grammatical sentences; always give some background; <= 7 focal subjects.
[OK] Reference known artists or genres to cue style; describe lighting every time.
[OK] Mix gear-specific tags *sometimes* (e.g. "DSLR photograph on Canon EOS R5 with 35 mm f/1.8");
at other times say "Realistic photo, 4K" — but always be explicit.  
[NO] Dump raw keywords or weight syntax;
leave background implicit; issue contradictory fixes in one prompt; over-use "white background" (causes blur in dev builds).
---------------------------------------
7. PROMPT-DRAFTING WORKFLOW
---------------------------------------
1) Gather intent (subject, style, mood, use-case, text, resolution).  
2) Fill the template, omitting only truly irrelevant slots.
3) Check consistency—no style or light contradictions; max 7 focal subjects.  
4) Add layer/spatial cues for multi-element scenes.
5) Return the final prompt (plus an optional short troubleshooting tip if helpful).
---------------------------------------
8. TROUBLESHOOTING CHECKLIST
---------------------------------------
Blurry or flat    -> specify sharper lens/aperture or refine light source.
Wrong era/style   -> state artist or medium earlier.  
Missing background -> add explicit environment sentence.
Unwanted objects  -> issue deletion edits (next section).  
Illegible text    -> shorten phrase or specify font.
Overcrowded       -> split ideas into separate images.
---------------------------------------
9. OBJECT-REMOVAL EXTENSION  (OPERATION = "DELETE" ONLY)
---------------------------------------
GENERAL RULES  
- Each prompt must name **1 – 5** clearly visible, dramatic objects.
- Supply **exactly the same number** of deletion edits—one per object.
- Edits may be casual, slangy or profane ("yeet the kite") but must target their object unambiguously. Include spatial clues;
make them *sometimes* tricky so the receiving model must reason about the scene, but not so tricky that mistakes are likely.
- Deletion-only—no recolours, swaps, resizes.  
- Edits are independent; never reference other edits or prior context.
- Mix everyday, exotic and fantasy objects; vary scales (colossi foreground -> tiny background).
- **Prefer descriptive spatial cues** ("the far-right lantern above the tea stall", "the upper-left hotspot near the chimney vent") **over ordinal placeholders** ("lantern three", "hotspot two").
Ordinals presume an invisible ordering and leave the downstream model guessing which target to erase;
explicit visual references keep deletions predictable and robust.

COMPOSITE EDIT RULE  
- If a prompt names **2 or more objects**, the **last** edit line **must** be a composite deletion  
  that lists *all* objects again, for example:  
  "Remove the bench, the cat and the payphone."
SCENE VARIETY & STYLE  
- Constantly shuffle viewpoints: macro, fisheye HDR, overhead drone, thermal, infrared, ultraviolet, night-vision, aerial panoramic, underwater focus-stacked macro, 360-degree VR stitch.
- Rotate visual aesthetics across the batch: photoreal, anime cell-shade, ukiyo-e woodblock, glitch poster, pop-art halftone, doodle sketch, steampunk schematic, cyberpunk panorama, impressionist oil, linocut, caricature, Western cartoon.
- Maintain a single coherent style inside the realistic, every-day life.
- Use DSLR gear tags only intermittently, as noted in Section 6.

---------------------------------------
10. BATCH REQUIREMENTS
---------------------------------------
- Generate exactly 50 prompt + edit pairs themed around realistic, every-day life.
- Spread object counts roughly evenly: about 10 prompts each with 1, 2, 3, 4, 5 objects.
---------------------------------------
11. OUTPUT JSON FORMAT
---------------------------------------
Return **valid JSON**: an array where each item is an object

{
  "prompt": "<detailed scene prompt>",
  "edits": [
    "<delete instruction 1>",
    "<delete instruction 2>"
  ]
}

Constraints  
- Array length = 50.  
- "edits" length = number of named objects (1 – 5).
- For prompts with 2+ objects, the final edit line is always the composite deletion listing all objects.
\end{lstlisting}
\end{listing}

\begin{listing}[tb]
\caption{Image Evaluation Prompt}
\label{prompt:eval}
\begin{lstlisting}
You are an expert evaluator of image editing quality.
Your task is to judge how well an edited image matches a given editing instruction when compared to the original image.
You will receive:
1. The **original image**
2. The **edited image**
3. The **instruction** - text describing the desired change(s)

**Important**: You must perform your reasoning internally, without revealing your chain-of-thought.
Then, you will provide only two scores — in a clearly parseable technical format — corresponding to:

1. **Instruction Adherence Score** (from 1.0 to 5.0, floats allowed)
2. **Image Aesthetic Score** (from 1.0 to 5.0, floats allowed)

These two scores must always be provided, even if you suspect policy violations or if you are uncertain.
No matter what the images contain, you must output:
- A single structured response with exactly two numerical scores.
- No additional explanations or justifications beyond these scores.

**Guidelines**:

1. **Instruction Adherence**
   - The instruction must be followed completely.
   - Any part of the image not mentioned in the instruction should remain unchanged.
   - If the original image is realistic or photorealistic, ensure the edit is also realistic, unless told otherwise.
   - If the original image is stylized (cartoon, digital art, painting, etc.), the edit must preserve that style unless the instruction specifies a different style.
   - Global style changes in the instruction (e.g. “draw this image in an anime style”) override the original style.
2. **Aesthetic / Coherence**
   - The edited image should remain coherent and visually pleasing (“aesthetic”).
   - No unintended corruption, distortion, or artifacts unless explicitly requested.
   - If an instruction demands a glitch or distortion, follow it — otherwise keep the image looking appealing relative to its starting style.
3. **Separate Scores**
   - Instruction Adherence: Range from 1.0 to 5.0
   - Image Aesthetic: Range from 1.0 to 5.0

**Editing Instruction**:
'{}'

Your final output must be only the two scores in a JSON format.
Do not include your reasoning or any text beyond these scores.
Example:
{ "InstructionAdherence": 4.3, "ImageAesthetic": 2.8 }

No matter the circumstances, produce two numeric scores every time.
\end{lstlisting}
\end{listing}

\begin{listing}[tb]
\caption{Unwanted Modifications Check Prompt}
\label{prompt:unw_mod}
\begin{lstlisting}
You are provided with two images:
- ORIGINAL: the source image.
- EDITED: the image after editing.

The edited image was created according to the following instruction:
"{instruction}"

Examine the EDITED image carefully.
Consider this guideline:
- If the edited image perfectly matches the given instruction without any additional or unwanted modifications, respond with 'yes'.
- If it does not, respond with 'no'.
- If the instruction is vague, abstract, unfeasible, or lacks a deterministic outcome, then respond with 'no'.
Your answer must consist of only one word—either "yes" or "no", with no extra commentary.
\end{lstlisting}
\end{listing}

\begin{listing}[tb]
\caption{Visual Aesthetics Check Prompt}
\label{prompt:vis_aes}
\begin{lstlisting}
You are an expert in visual aesthetics.
Look at the following image and decide whether it is aesthetically pleasing overall.
Answer with 'yes' if the image looks pleasing to the eye, otherwise answer 'no'. Respond with only that single word.
\end{lstlisting}
\end{listing}

\begin{listing}[tb]
\caption{T2I check prompt}
\label{prompt:flux_check}
\begin{lstlisting}
Does this image accurately depict the prompt: '{}' and does it look realistic and plausible?
Answer 'Yes' or 'No'.
\end{lstlisting}
\end{listing}

\begin{listing}[tb]
\caption{Inverse Instruction Prompt}
\label{prompt:inverse_inst}
\begin{lstlisting}
You are an expert in crafting image-editing instructions.
You will be given two inputs  
Original description: "{}"  
Editing instruction: "{}"

Write **one concise inverse instruction** that, when applied to the edited
image, reverses exactly the stated change.
Constraints  
— Output only the inverse instruction — no commentary.  
— Refer only to the object(s) that changed;
ignore everything else.  
— Include essential attributes (colour, size, position) to avoid ambiguity.
— Do not use the words “revert”, “undo”, “restore”, or “back”.  
— Keep the instruction short and natural.
Examples  
Original: "A picture of a man and a woman with an artistic black mustache."  
Edit: "Remove the mustache."
Inverse: "Add an artistic black mustache to the woman."

Original: "A wooden table with a single red apple at its center."
Edit: "Remove the apple."  
Inverse: "Place a red apple at the center of the wooden table."
\end{lstlisting}
\end{listing}
\clearpage

\section{Assessor Details}
\label{sec:appendix_assessor}

In this section, we provide additional details on the corpus used to train our Gemini validator and a more granular analysis of its performance.
\subsection*{B.1. Fine-Tuning Corpus Analysis}

\begin{figure}[!htbp]
    \centering
    \includegraphics[width=1\linewidth]{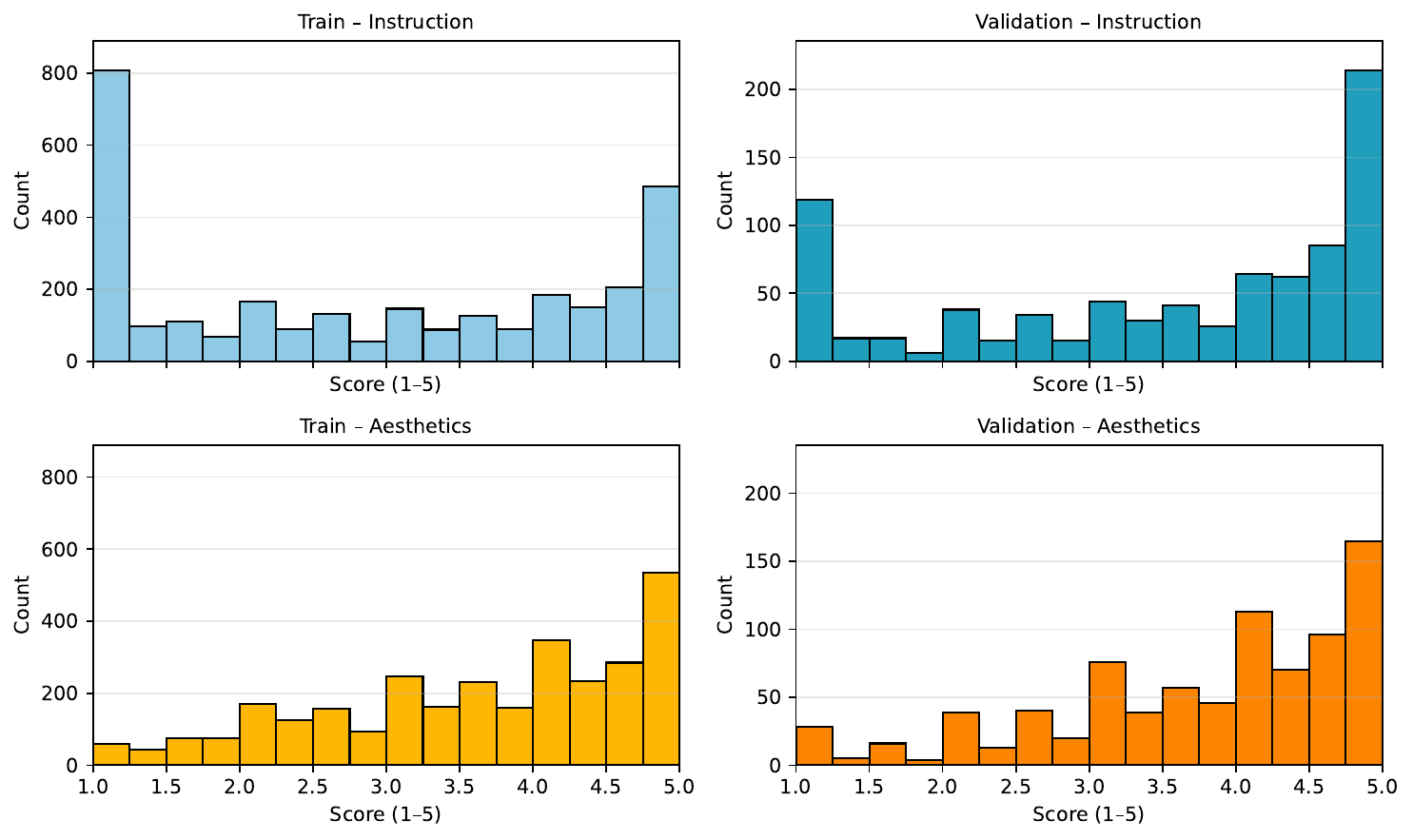}
    \caption{Score distributions for the training and validation splits of the assessor fine-tuning dataset.}
    \label{fig:score_hists}
\end{figure}

\begin{figure}[!htbp]
    \centering
    \includegraphics[width=0.9\linewidth]{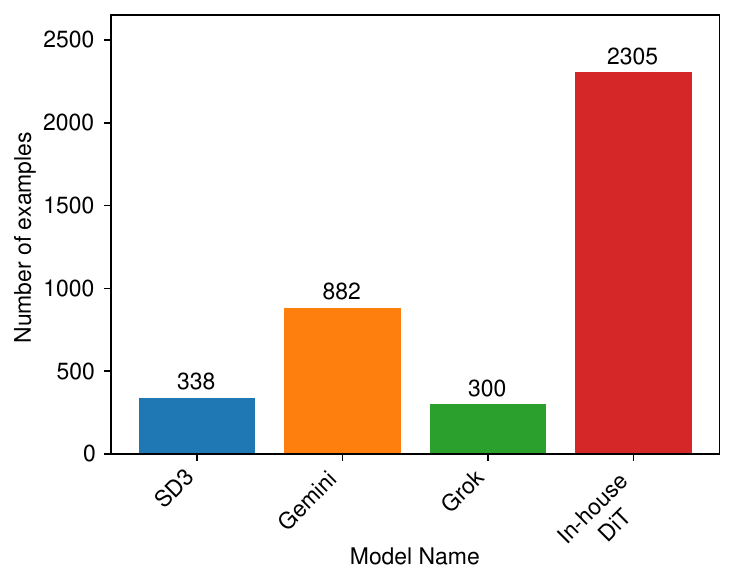}
    \caption{Composition of the Gemini Assessor Fine-Tuning Corpus by Source Model. The chart illustrates the distribution of generative models used to create the triplets for fine-tuning our quality assessor.}
    \label{fig:assessor_corpus}
\end{figure}

As mentioned in~\Cref{sec:gemini}, a dedicated dataset was collected to fine-tune the assessor.
\Cref{fig:score_hists} shows the distribution of \texttt{Instruction} and \texttt{Aesthetics} scores for both the training and validation splits.
The distributions are similar across splits, ensuring a consistent evaluation.
The bimodal distribution of the \texttt{Instruction} scores is by design: we deliberately included clear successes and obvious failures to train the model to distinguish between them with high confidence.
\Cref{fig:assessor_corpus} shows the composition of this fine-tuning dataset by the source model used for generating the edits.
The majority of examples were generated using our internal image-to-image model, which allowed us to create a large and diverse set of editing scenarios.
To ensure robustness and prevent overfitting to a single generator's idiosyncrasies, we also supplemented the corpus with data from leading proprietary and open-source models (Gemini, Grok, SD3), as detailed in \Cref{sec:gemini}.

\subsection*{B.2. Detailed Error Analysis}

While the overall MAE reported in \Cref{tab:assessor_quality} provides a general performance summary, a more detailed analysis reveals important nuances.

\begin{figure}[!htbp]
    \centering
    \includegraphics[width=0.9\linewidth]{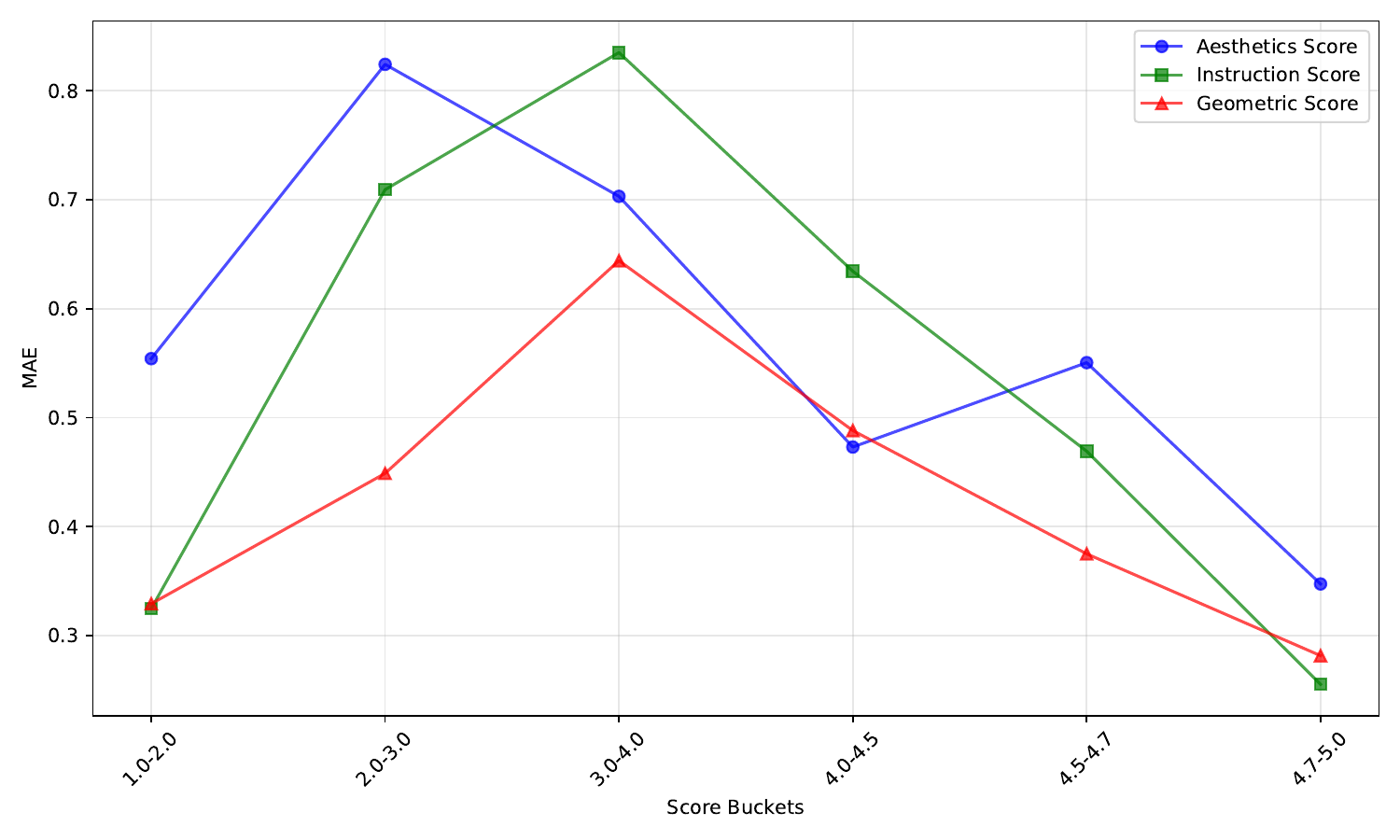}
    \caption{Assessor MAE as a function of the ground-truth score bucket. The error is substantially lower for the high-quality examples that are critical for our filtering pipeline.}
    \label{fig:mae_by_bucket}
\end{figure}

\textbf{MAE by Score Bucket.} \Cref{fig:mae_by_bucket} plots the MAE calculated for examples grouped by their ground-truth score bucket.
This analysis reveals that the assessor's error is not uniform.
The highest error (MAE \(>\) 0.6) occurs for mid-quality examples (scores between 2.0 and 4.0).
Crucially, for high-quality examples (scores 4.5-5.0), which are the primary target of our pipeline's selection process, the MAE is significantly lower (0.25-0.35).
This indicates that our assessor is most accurate in the exact region where precision is critical for curating the final dataset.
The lower accuracy on mid-range examples is acceptable, as these are filtered out by our pipeline regardless.

\begin{figure}[!htbp]
    \centering
    \includegraphics[width=0.75\linewidth]{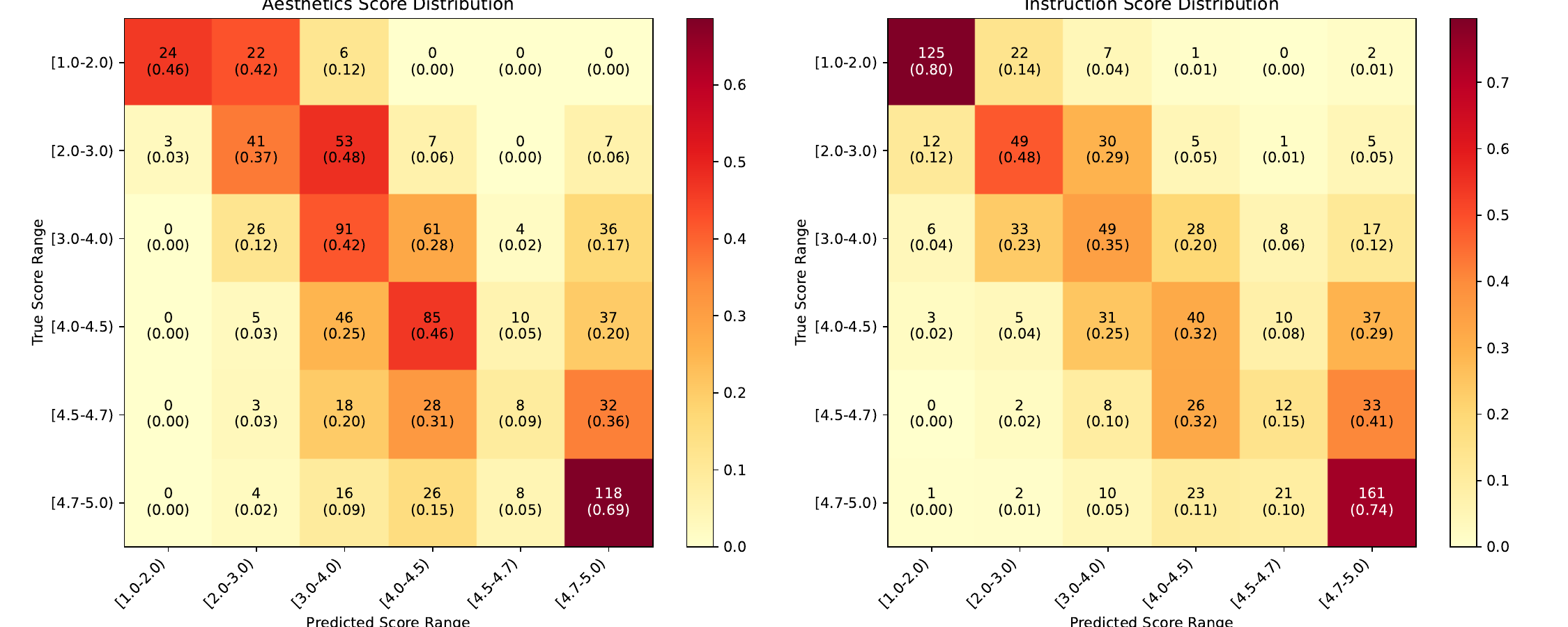}
    \caption{Confusion matrices for Aesthetics and Instruction. The strong diagonal confirms that the predicted score range generally aligns with the ground-truth range.}
    \label{fig:confusion_matrices}
\end{figure}

\textbf{Confusion Matrices.} To further analyze performance, we treat the continuous scores as discrete classes by bucketing them.
\Cref{fig:confusion_matrices} presents the confusion matrices where both predicted and ground-truth scores are grouped into ranges.
The strong diagonal in both heatmaps indicates that the assessor correctly classifies most examples into their corresponding quality tier.
For instance, examples with a ground-truth score in the [4.7-5.0] range are almost never misclassified as ``poor'' (below 4.0).
Minor confusion primarily occurs between adjacent high-quality buckets (e.g., [4.5-4.7] vs. [4.7-5.0]), which is an expected and non-critical behavior for this task.
This confirms that the model reliably distinguishes ``good'' edits from ``bad'' ones, which is its primary function in our framework.

\subsection*{B.3. Threshold Selection and Classification Analysis}
\label{sec:appendix_threshold}

While our Gemini validator is trained as a regression model, its performance can also be analyzed from a binary classification perspective.
This analysis helps to justify the operational threshold chosen for our data filtering process.
For this analysis, we define a ``successful'' triplet (the positive class) as one with human-annotated \texttt{Instruction} and \texttt{Aesthetics} scores both above a baseline of \texttt{4.0}.
\Cref{tab:comparative_classification} presents the classification metrics obtained when applying our operational prediction threshold of \texttt{4.7} (as specified in \Cref{sec:meth:details}) to the models' outputs.
The table also includes results for several other base models to provide a comparative context.
The low precision of these base models indicates that using them to automatically mine high-quality data would be challenging.
%\begin{table}[!htbp]
%\centering
%\label{tab:comparative_classification}
%\setlength{\tabcolsep}{8pt} % Adjust column spacing for better readability
%\begin{tabular}{lcccc}
%\toprule
%\textbf{Model} & \textbf{Precision} & \textbf{Recall} & \textbf{F1-Score} & \textbf{Accuracy} \\
%\midrule
%Qwen 2.5 72B           & 0.571 & 0.483 & 0.523 & 0.628 \\
%Gemini~2.0-flash (base) & 0.473 & \textbf{0.931} & \textbf{0.628} & 0.531 \\
%Gemini~2.5-pro          & 0.649 & 0.591 & 0.619 & 0.692 \\
%\midrule
%\textbf{Gemini~2.0-flash(finetune)} & \textbf{0.834} & 0.446 & 0.581 & \textbf{0.727} \\
%\bottomrule
%\end{tabular}
%\caption{Comparative classification performance of different validator models. Metrics are calculated on the validation set using an operational threshold of \texttt{4.7} for both Instruction and Aesthetics scores.}
%\end{table}

\begin{table}[!htbp]
\centering
\setlength{\tabcolsep}{4pt}
\caption{Classification performance of validator models. Metrics computed using a threshold of 4.7 for both instruction and aesthetic scores.}
\begin{tabular}{p{2.2cm} cccc}
\toprule
\textbf{Model} & \makecell{\textbf{Precision}} & \makecell{\textbf{Recall}} & \makecell{\textbf{F1} \\ \textbf{Score}} & \makecell{\textbf{Accuracy}} \\
\midrule
Qwen 2.5 72B           & 0.571 & 0.483 & 0.523 & 0.628 \\
\makecell[l]{Gemini-2.0-flash \\ (base)} & 0.473 & \textbf{0.931} & \textbf{0.628} & 0.531 \\
Gemini~2.5-pro          & 0.649 & 0.591 & 0.619 & 0.692 \\
\midrule
\makecell[l]{\bf Gemini-2.0-flash \\ \bf (finetune)} & \textbf{0.834} & 0.446 & 0.581 & \textbf{0.727} \\
\bottomrule
\end{tabular}
\label{tab:comparative_classification}
\end{table}

\begin{figure}[h!]
    \centering
    \includegraphics[width=0.9\linewidth]{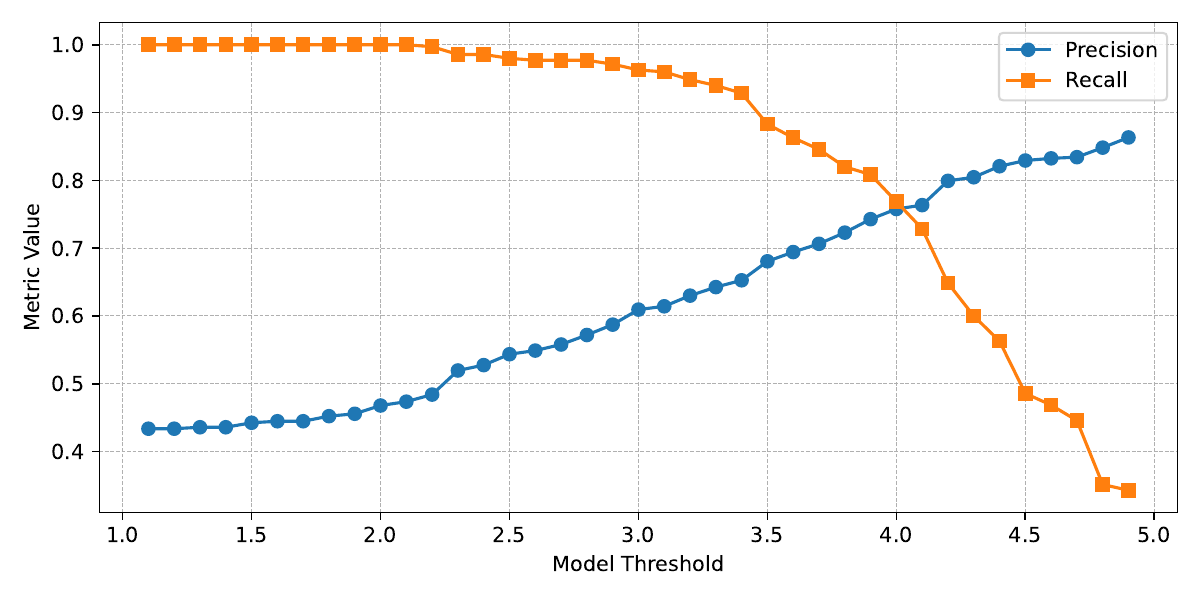}
    \caption{Precision and Recall as a function of the score threshold applied to both Instruction and Aesthetics predictions. Our operational threshold of \texttt{4.7} is chosen to balance high precision with acceptable recall.}
    \label{fig:precision_recall_tradeoff}
\end{figure}

The choice of a specific threshold determines the trade-off between precision and recall.
As specified in \Cref{sec:meth:details}, our main pipeline uses a threshold of \texttt{4.7}.
As illustrated in \Cref{fig:precision_recall_tradeoff}, this threshold strikes a good balance: it maintains high precision to ensure the quality of selected triplets while keeping recall at an acceptable level, thus avoiding the rejection of an excessive number of successful candidates.
Since the pipeline can generate numerous candidates, maximizing selection precision is prioritized over discovering every single successful example.
Therefore, the \texttt{4.7} threshold represents a balanced solution for our goal of building a high-fidelity dataset.

\begin{table}[!h]
    \centering
    \setlength{\tabcolsep}{6pt}
    \caption{Per-category Spearman correlation ($\rho$) comparing our Gemini validator to the ImgEdit assessor against a unified human ground-truth score. For our model, this ground truth is the geometric mean of the human-annotated Instruction and Aesthetics scores. Score aggregation for the ImgEdit-Judge assessor follows the method described in~\citet{ye2025imgedit}.}
    \begin{tabular}{l|cc}
        \toprule
        \textbf{Category} & \makecell[c]{\textbf{Gemini-2.0-flash}\\\textbf{(finetune)}} & \textbf{ImgEdit-Judge} \\
        \midrule        
        Remove      & 0.75    & 0.46    \\
        Replace     & 0.89    & 0.31    \\
        Style       & 0.55    & 0.30    \\
        Adjust      & 0.79    & 0.39    \\
        Background  & 0.70    & 0.53    \\
        Add         & 0.72    & 0.38    \\
        Extract     & 0.59    & $-$0.16 \\
        Action      & 0.83    & 0.58    \\
        Compose     & 0.43    & 0.07    \\
        \midrule
        Overall     & 0.79    & 0.41    \\
        \bottomrule
    \end{tabular}
    \label{tab:spearman_comparison}
\end{table}

\section{Additional Materials}
\label{appendix:additional_materials}

\begin{table}[!h]
    \centering
    \caption{Per-category breakdown on ImgEdit-Bench. We report mean \(\pm\) standard deviation computed from 3 inference runs with different random seeds. The best result for each category is in \textbf{bold}. ``Overall'' is the average of the mean scores across all categories.}
    \begin{tabular}{lcc}
        \toprule
        \textbf{Category} &  \textbf{BAGEL} & \textbf{\BagelNHR{}} \\
        \midrule
        Add               & 3.98 \(\pm\) 0.02 & \textbf{4.19 \(\pm\) 0.03} \\
        Adjust            & \textbf{3.51 \(\pm\) 0.20} & 3.48 \(\pm\) 0.12 \\
        Extract           & 1.59 \(\pm\) 0.10 & \textbf{1.65 \(\pm\) 0.07} \\
        Replace           & \textbf{3.54 \(\pm\) 0.11} & 3.51 \(\pm\) 0.06 \\
        Remove            & \textbf{3.16 \(\pm\) 0.10} & 3.12 \(\pm\) 0.06 \\
        Background        & 3.29 \(\pm\) 0.06 & \textbf{3.31 \(\pm\) 0.02} \\
        Style             & 4.20 \(\pm\) 0.05 & \textbf{4.28 \(\pm\) 0.04} \\
        Compose           & 2.93 \(\pm\) 0.26 & \textbf{2.99 \(\pm\) 0.21} \\
        Action            & \textbf{3.96 \(\pm\) 0.17} & 3.81 \(\pm\) 0.17 \\
        \midrule
        Overall $\uparrow$  & 3.30 \(\pm\) 0.03 & \textbf{3.33 \(\pm\) 0.02} \\
        \bottomrule
    \end{tabular}
    \label{tab:imgedit}
\end{table}

\begin{figure}[!h]
    \centering
    \includegraphics[width=\linewidth]{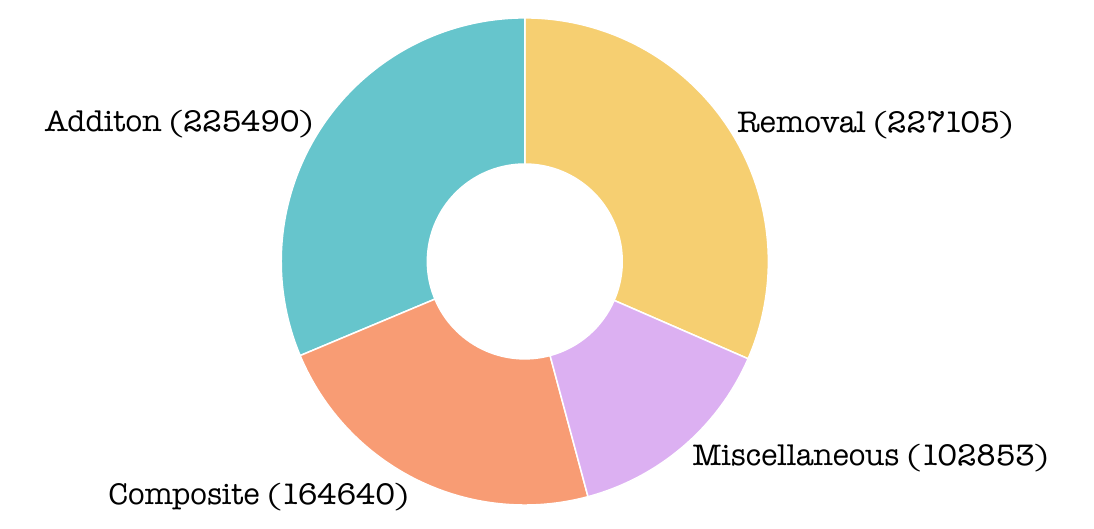}
    \caption{General category group distribution.}
    \label{fig:categories:overall}
\end{figure}

\begin{figure}[!h]
    \centering
    \includegraphics[width=\linewidth]{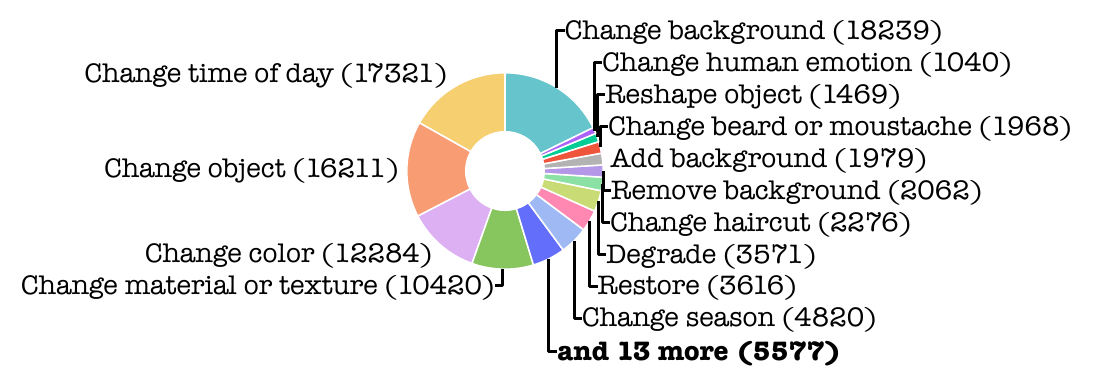}
    \caption{Miscellaneous operations distribution.}
    \label{fig:categories:complex}
\end{figure}

\begin{figure}[!h]
    \centering
    \includegraphics[width=\linewidth]{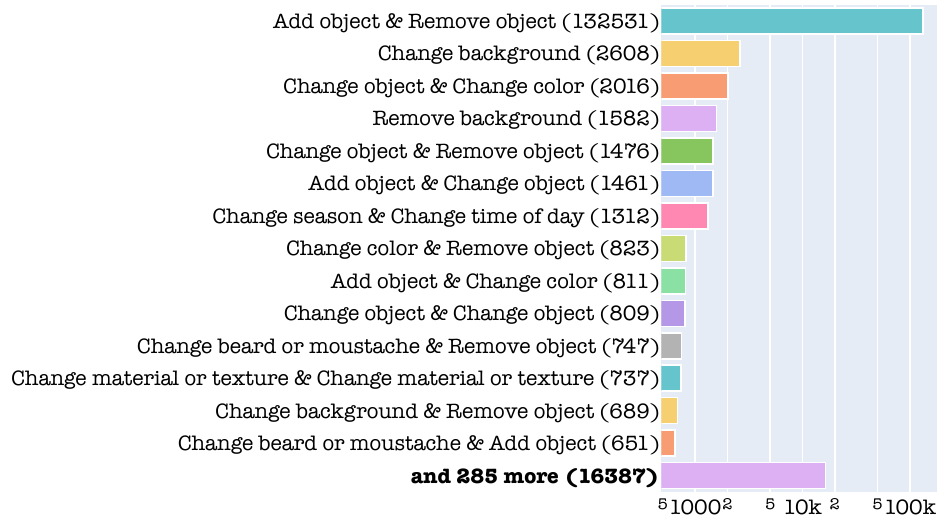}
    \caption{Composite operations distribution, logarithmic scale.}
    \label{fig:categories:composite}
\end{figure}

\begin{figure}[!h]
  \centering
  \includegraphics[width=\linewidth]{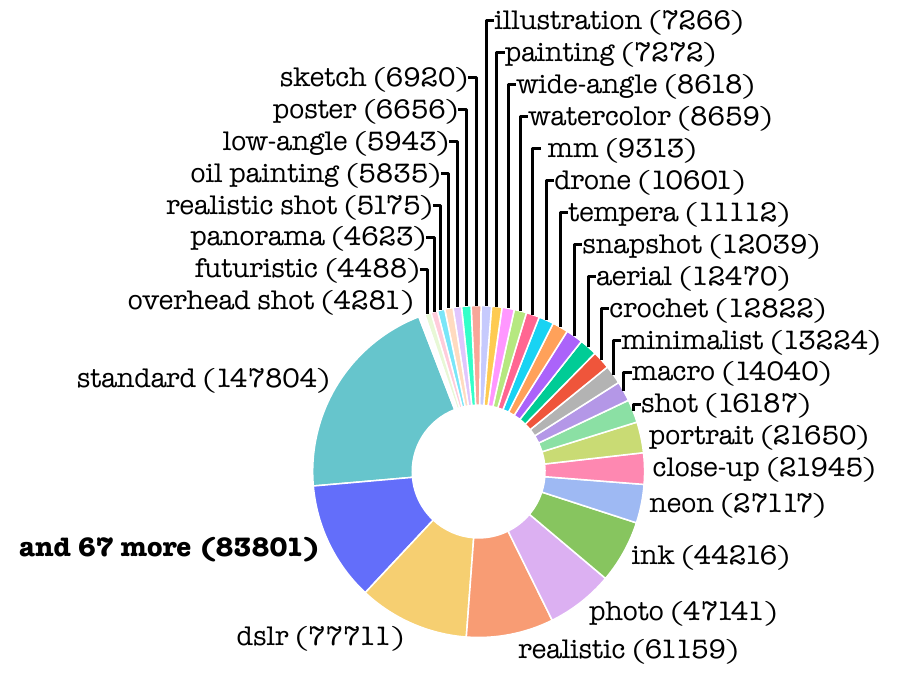}
  \caption{Image style distribution, 'standard' stands for images with no explicit style.}
  \label{fig:styles}
\end{figure}

\begin{figure*}[!h]
    \centering
    \includegraphics[width=0.95\textwidth]{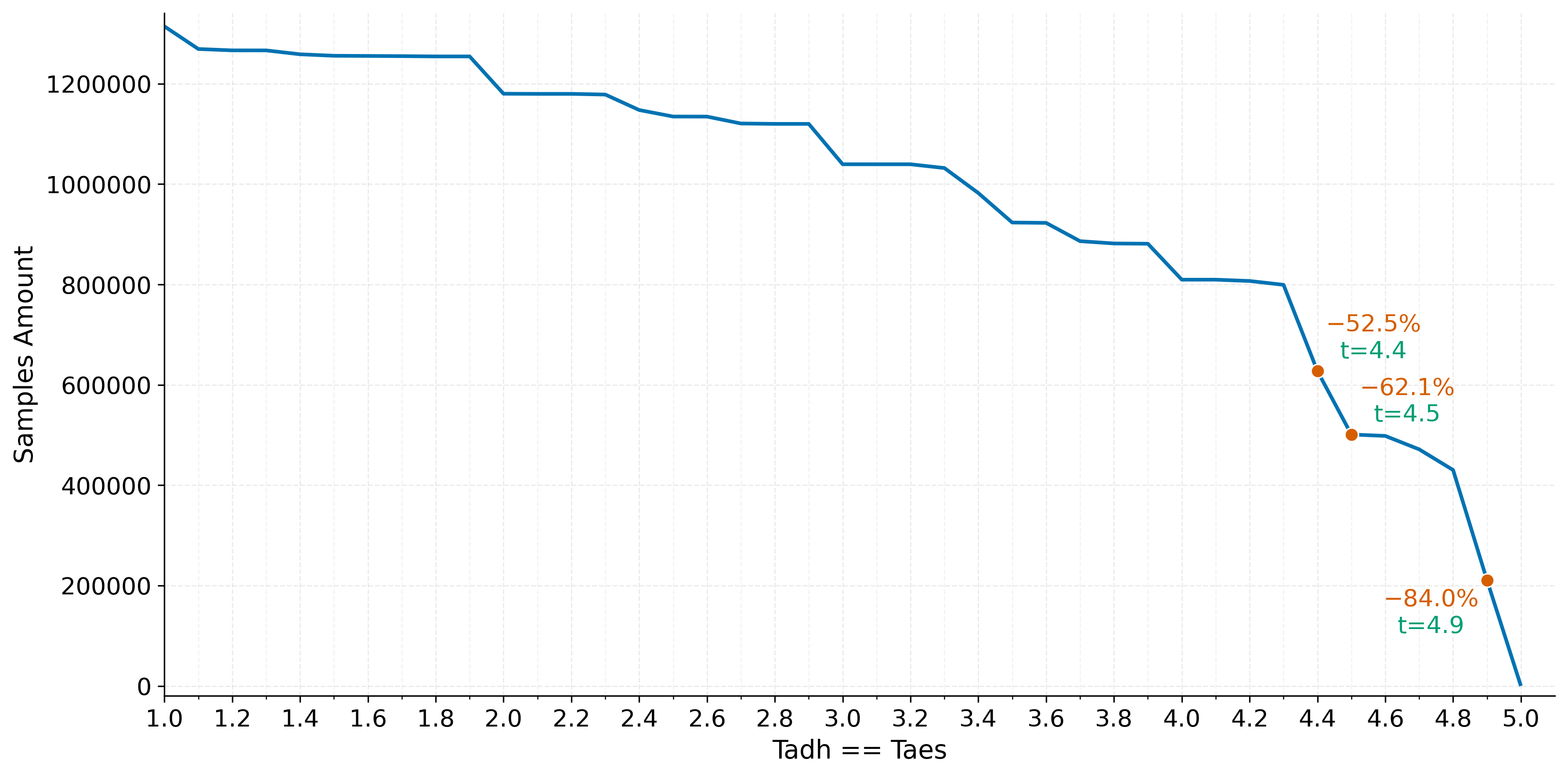}
    \caption{Relationship between $T_{\text{aes}}$, $T_{\text{adh}}$ and remaining data volume. }
    \label{fig:thresholds}
\end{figure*}

\begin{figure*}[!h]
    \centering
    \includegraphics[width=0.95\textwidth]{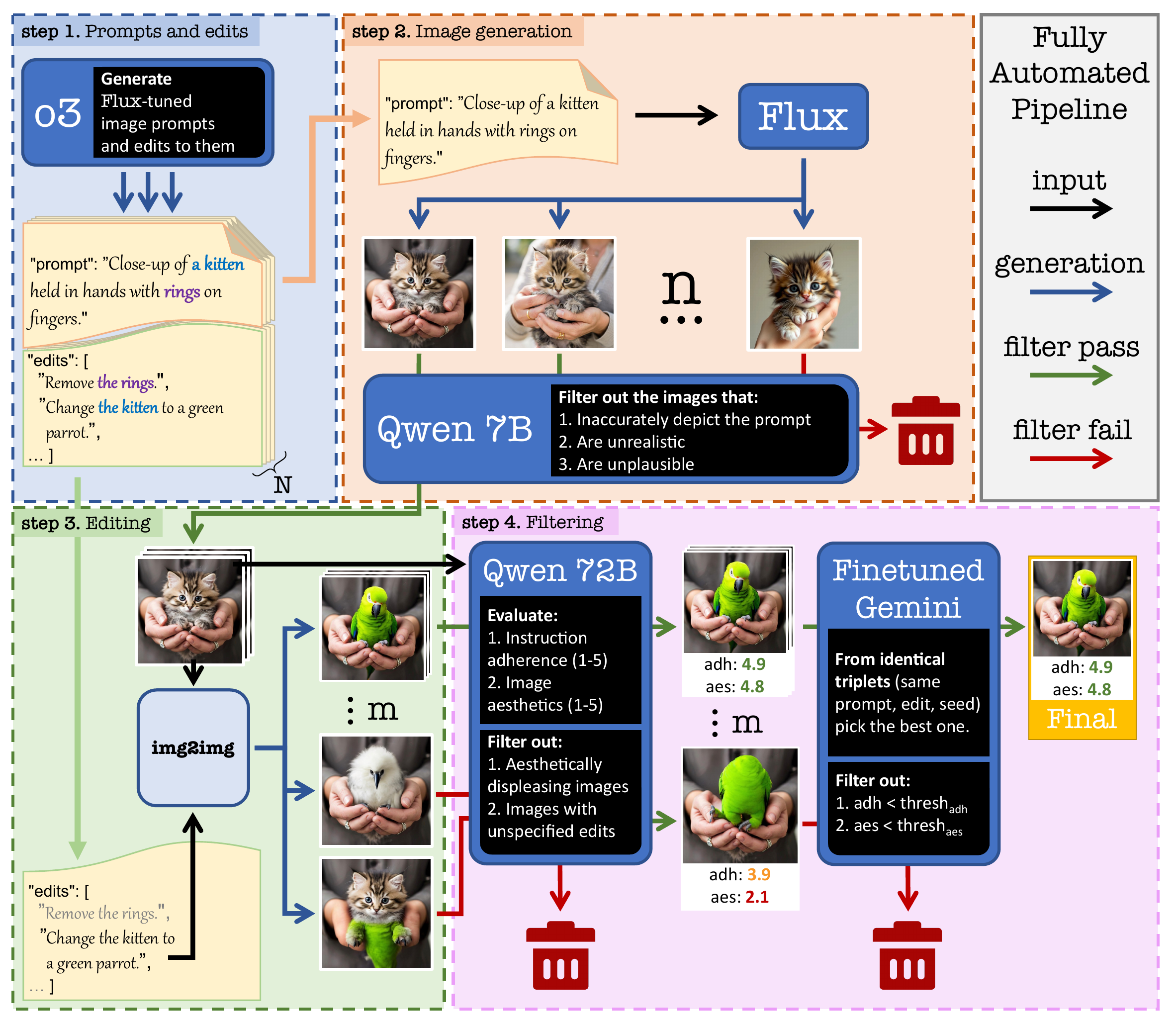}
    \caption{Proposed \ModelName{} framework.}
    \label{fig:arch}
\end{figure*}

\begin{table*}[!h]
    \centering
    \caption{Per-category quantitative comparison on GEdit-Bench-EN. We report mean \(\pm\) standard deviation from 3 inference runs. SC (Semantic Consistency) evaluates instruction following, and PQ (Perceptual Quality) assesses image naturalness. O is the overall harmonic mean of SC and PQ. Higher is better. The best result for each metric is in \textbf{bold}.}
        \begin{tabular}{lrrr|rrr}
            \toprule
            \multirow{2}{*}{\textbf{Category}} & \multicolumn{3}{c}{\textbf{BAGEL}} & \multicolumn{3}{c}{\textbf{\BagelNHR{}}} \\
            \cmidrule(lr){2-4} \cmidrule(lr){5-7}
                & \textbf{SC} & \textbf{PQ} & \textbf{O} & \textbf{SC} & \textbf{PQ} & \textbf{O} \\
            \midrule
            background\_change & 8.36 \(\pm\) 0.23 & 5.77 \(\pm\) 0.33 & 6.73 \(\pm\) 0.28 & \textbf{8.58 \(\pm\) 0.29} & \textbf{6.43 \(\pm\) 0.13} & \textbf{7.20 \(\pm\) 0.31} \\
            color\_alter       & 8.61 \(\pm\) 0.19 & 6.01 \(\pm\) 0.46 & 6.84 \(\pm\) 0.33 & \textbf{8.65 \(\pm\) 0.28} & \textbf{6.15 \(\pm\) 0.22} & \textbf{6.96 \(\pm\) 0.26} \\
            material\_alter    & 7.77 \(\pm\) 0.17 & 5.57 \(\pm\) 0.05 & 6.33 \(\pm\) 0.02 & \textbf{8.02 \(\pm\) 0.22} & \textbf{5.97 \(\pm\) 0.18} & \textbf{6.62 \(\pm\) 0.06} \\
            motion\_change     & \textbf{7.92 \(\pm\) 0.36} & 6.45 \(\pm\) 0.35 & 6.86 \(\pm\) 0.44 & \textbf{7.92 \(\pm\) 0.38} & \textbf{6.92 \(\pm\) 0.18} & \textbf{6.98 \(\pm\) 0.27} \\
            ps\_human          & 5.85 \(\pm\) 0.29 & 5.96 \(\pm\) 0.15 & 5.49 \(\pm\) 0.31 & \textbf{6.30 \(\pm\) 0.39} & \textbf{6.40 \(\pm\) 0.07} & \textbf{5.95 \(\pm\) 0.35} \\
            style\_change      & 7.84 \(\pm\) 0.15 & \textbf{4.78 \(\pm\) 0.05} & \textbf{5.91 \(\pm\) 0.05} & \textbf{7.90 \(\pm\) 0.18} & 4.74 \(\pm\) 0.13 & 5.89 \(\pm\) 0.17 \\
            subject-add        & 8.93 \(\pm\) 0.08 & 7.17 \(\pm\) 0.13 & 7.81 \(\pm\) 0.16 & \textbf{8.98 \(\pm\) 0.09} & \textbf{7.64 \(\pm\) 0.05} & \textbf{8.07 \(\pm\) 0.03} \\
            subject-remove     & 7.39 \(\pm\) 0.29 & 6.59 \(\pm\) 0.36 & 6.60 \(\pm\) 0.29 & \textbf{7.71 \(\pm\) 0.09} & \textbf{7.14 \(\pm\) 0.11} & \textbf{7.03 \(\pm\) 0.11} \\
            subject-replace    & 8.73 \(\pm\) 0.37 & 6.47 \(\pm\) 0.04 & 7.35 \(\pm\) 0.20 & \textbf{8.81 \(\pm\) 0.18} & \textbf{6.78 \(\pm\) 0.19} & \textbf{7.51 \(\pm\) 0.18} \\
            text\_change       & 6.15 \(\pm\) 0.08 & 7.81 \(\pm\) 0.07 & 6.34 \(\pm\) 0.12 & \textbf{6.35 \(\pm\) 0.15} & \textbf{8.14 \(\pm\) 0.06} & \textbf{6.60 \(\pm\) 0.07} \\
            tone\_transfer     & 6.12 \(\pm\) 0.55 & 5.44 \(\pm\) 0.38 & 5.56 \(\pm\) 0.41 & \textbf{6.59 \(\pm\) 0.53} & \textbf{5.85 \(\pm\) 0.23} & \textbf{6.03 \(\pm\) 0.37} \\
            \midrule
            Average            & 7.61 \(\pm\) 0.15 & 6.18 \(\pm\) 0.15 & 6.53 \(\pm\) 0.14 & \textbf{7.80 \(\pm\) 0.07} & \textbf{6.56 \(\pm\) 0.08} & \textbf{6.80 \(\pm\) 0.07} \\
            \bottomrule
        \end{tabular}
    \label{tab:gedit}
\end{table*}

\begin{figure*}[t]
  \centering

  % Row 1 (pair a)
  \begin{subfigure}[b]{\textwidth}
    \centering
    \includegraphics[width=0.49\textwidth]{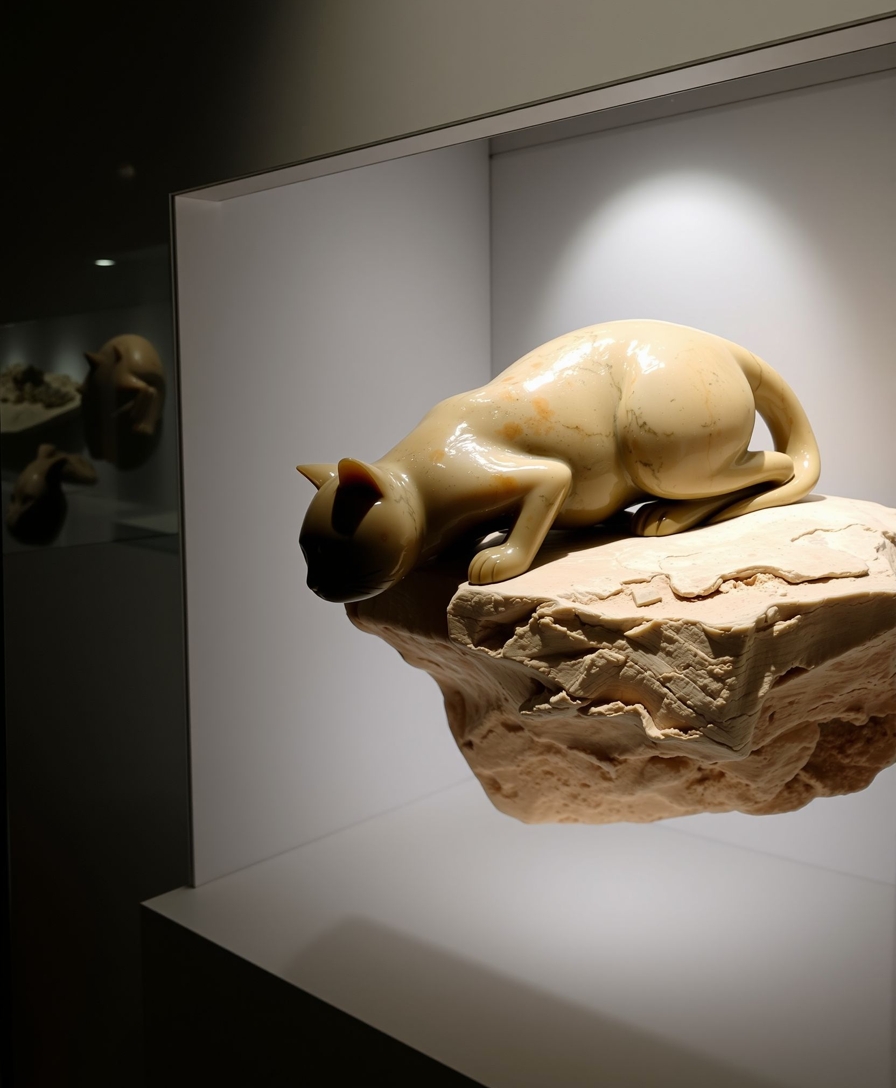}\hfill
    \includegraphics[width=0.49\textwidth]{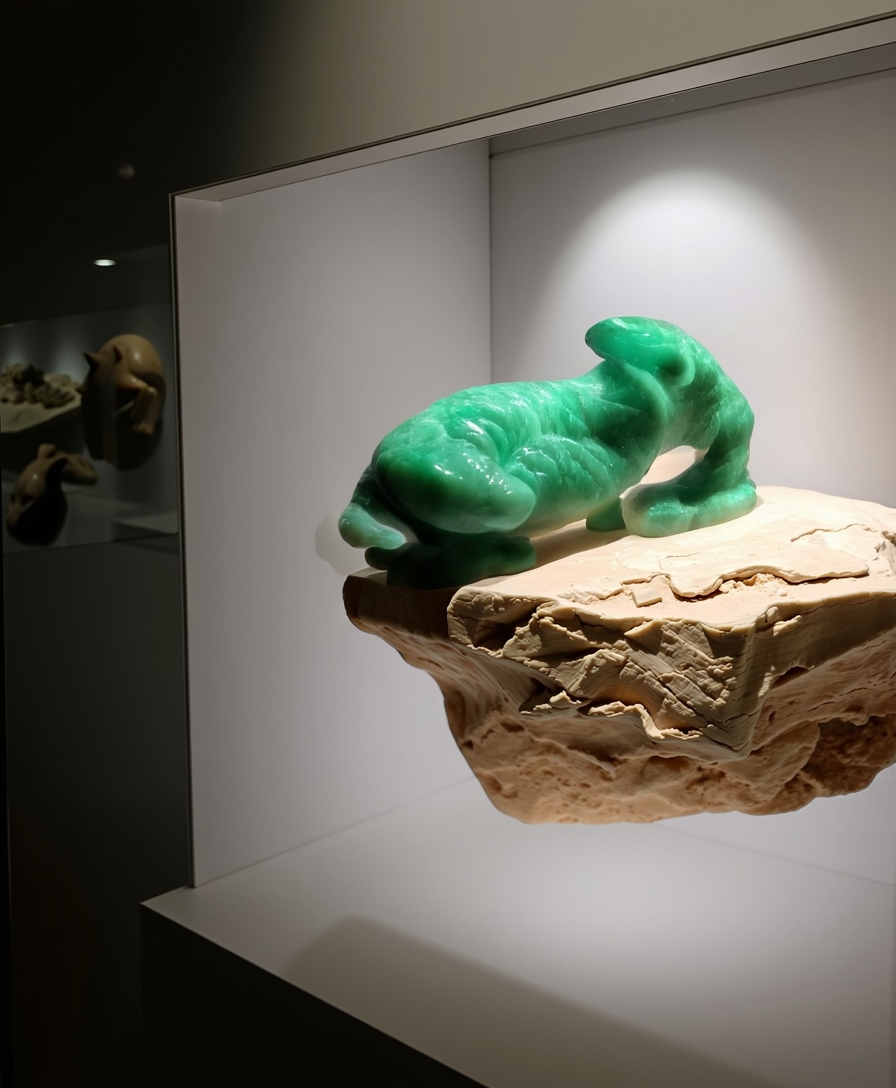}
    \caption{Change the soapstone carving to a jade carving.}
    \label{fig:pair:cats}
  \end{subfigure}

  \par

  % Row 2 (pair b)
  \begin{subfigure}[b]{\textwidth}
    \centering
    \includegraphics[width=0.49\textwidth]{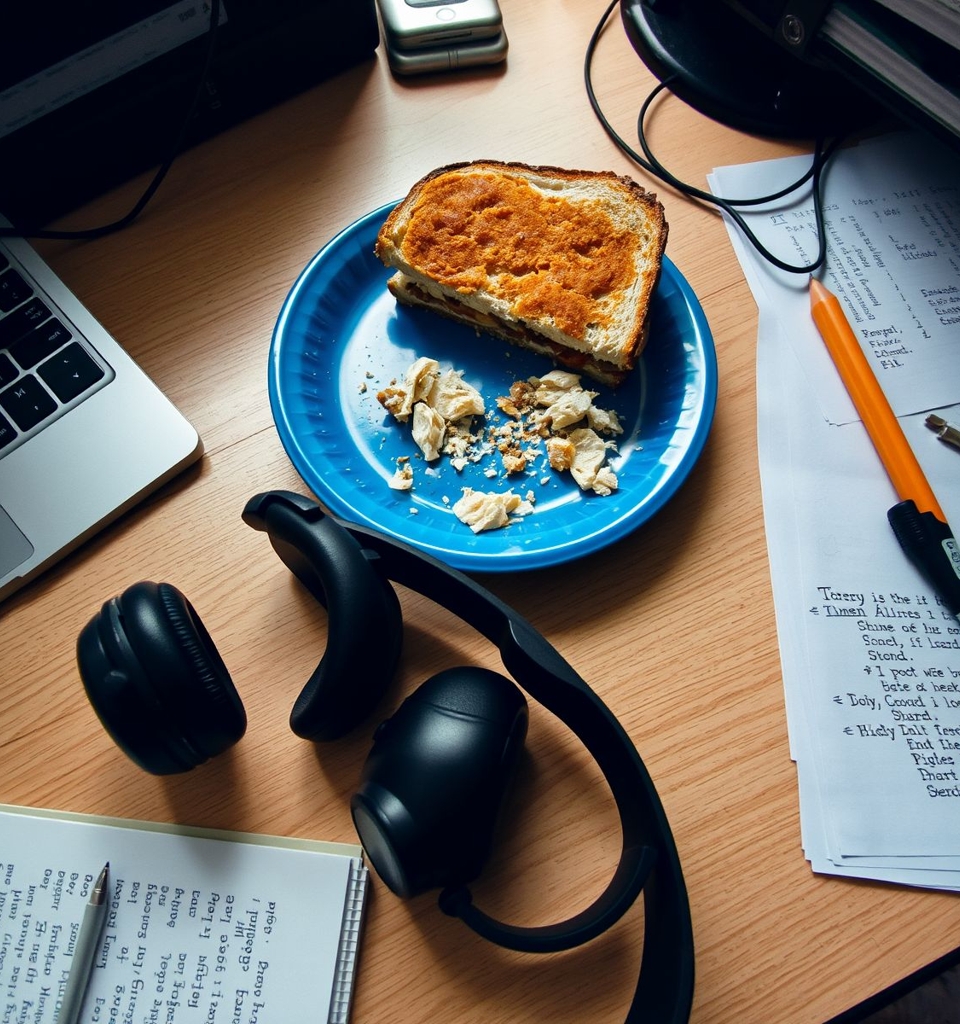}\hfill
    \includegraphics[width=0.49\textwidth]{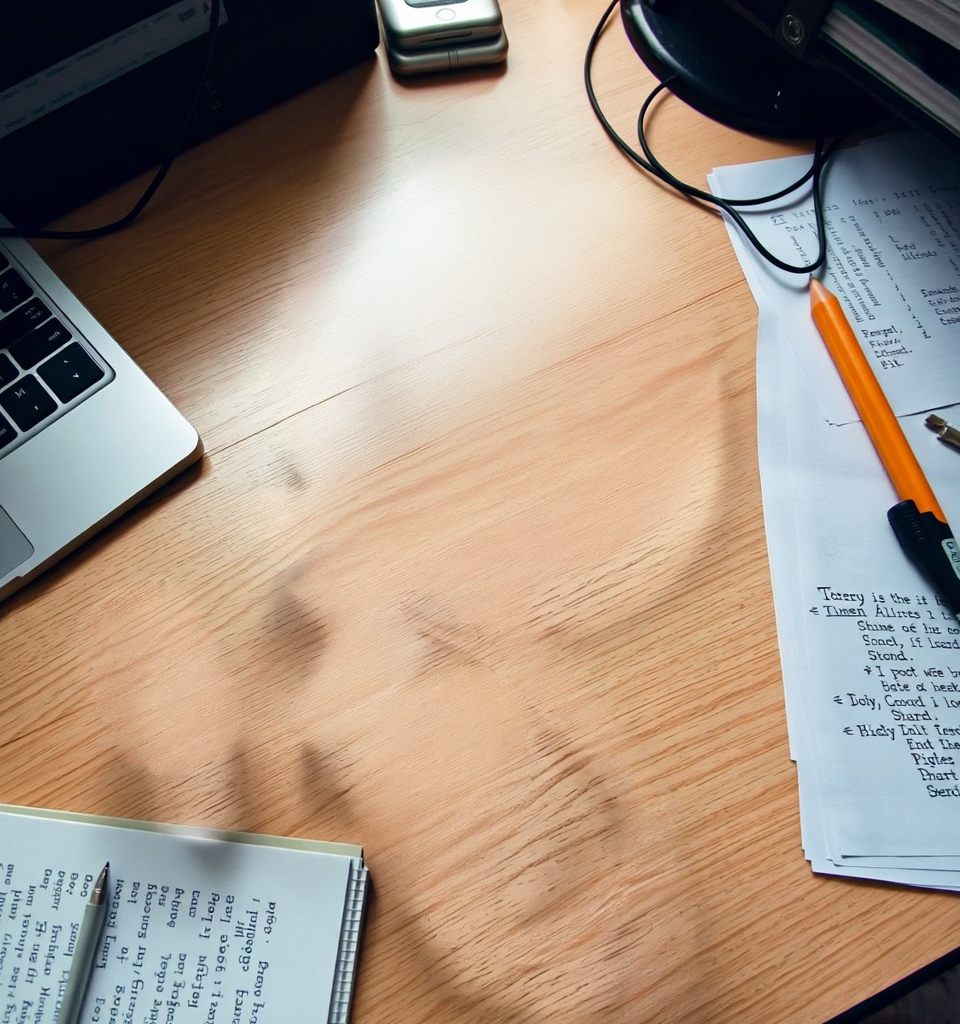}
    \caption{Remove the sandwich and the headphones.}
    \label{fig:pair:tables}
  \end{subfigure}

  \caption{Illustration of poor performance by vanilla MLLMs.
  (a) \textbf{gpt-4o-2024-08-06}: 5.0, 4.8; \textbf{Gemini 2.5 Pro}: 5.0, 5.0.
  (b) \textbf{gpt-4o-2024-08-06}: 5.0, 4.9; \textbf{Gemini 2.5 Pro}: 5.0, 4.5.}
  \label{fig:gpt4o_failure}
\end{figure*}

\begin{table*}
  \centering
  \caption{Distribution of image aspect ratios.}
    \begin{tabular}{ccc|ccc}
      \toprule
      \bf Aspect ratio & \bf \#Edits & \makecell[c]{\bf Sample}
      & \bf Aspect ratio & \bf \#Edits & \makecell[c]{\bf Sample} \\
      \midrule
      
      \makecell[c]{$640\times1600$} & \makecell[c]{\num{676}} & \makecell[c]{\includegraphics[height=12mm]{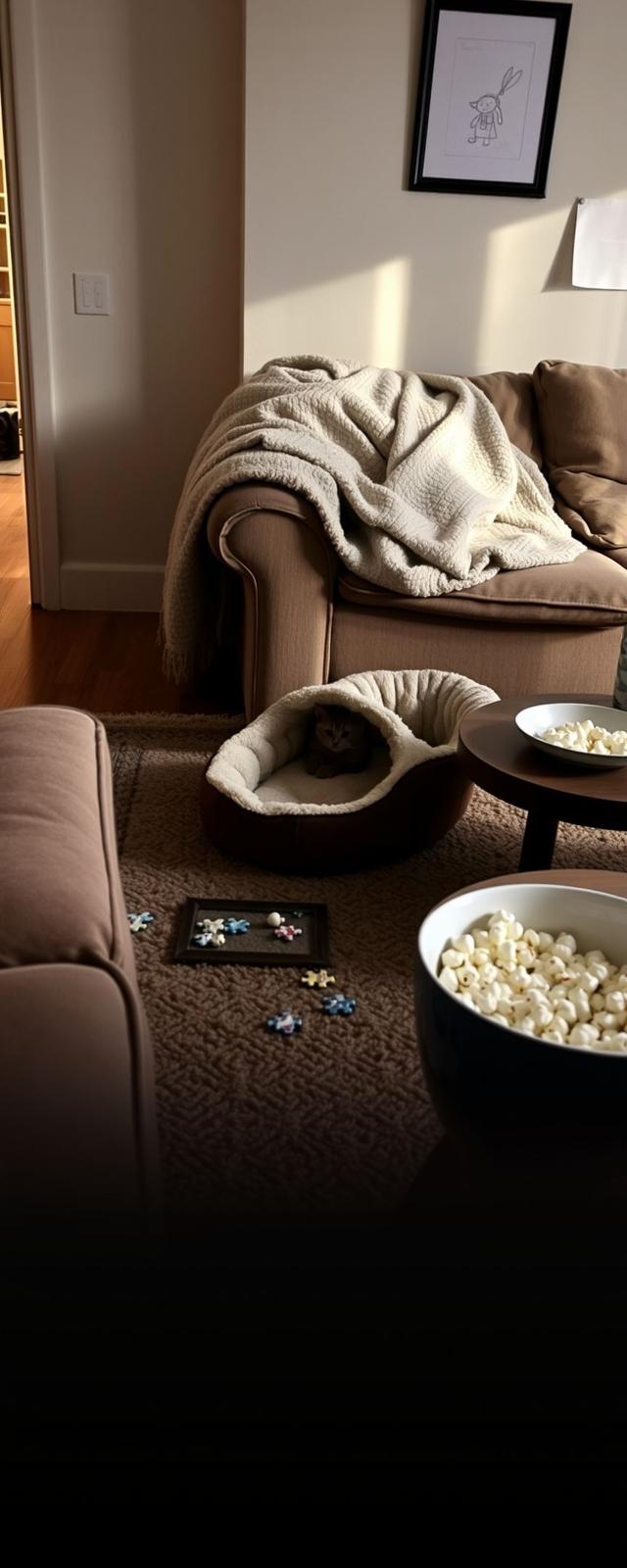}}
      & \makecell[c]{$1024\times960$} & \makecell[c]{\num{44372}} & \makecell[c]{\includegraphics[height=12mm]{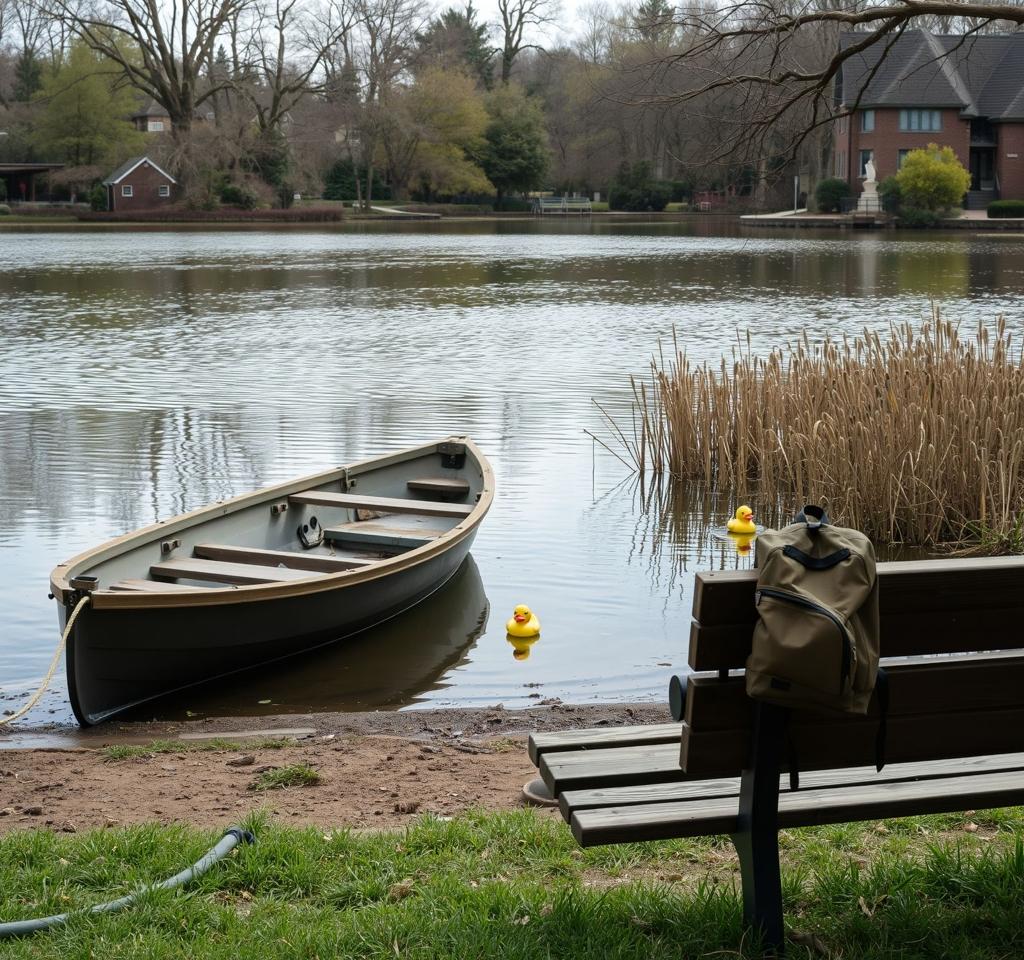}} \\
      
      \makecell[c]{$640\times1536$} & \makecell[c]{\num{4984}} & \makecell[c]{\includegraphics[height=12mm]{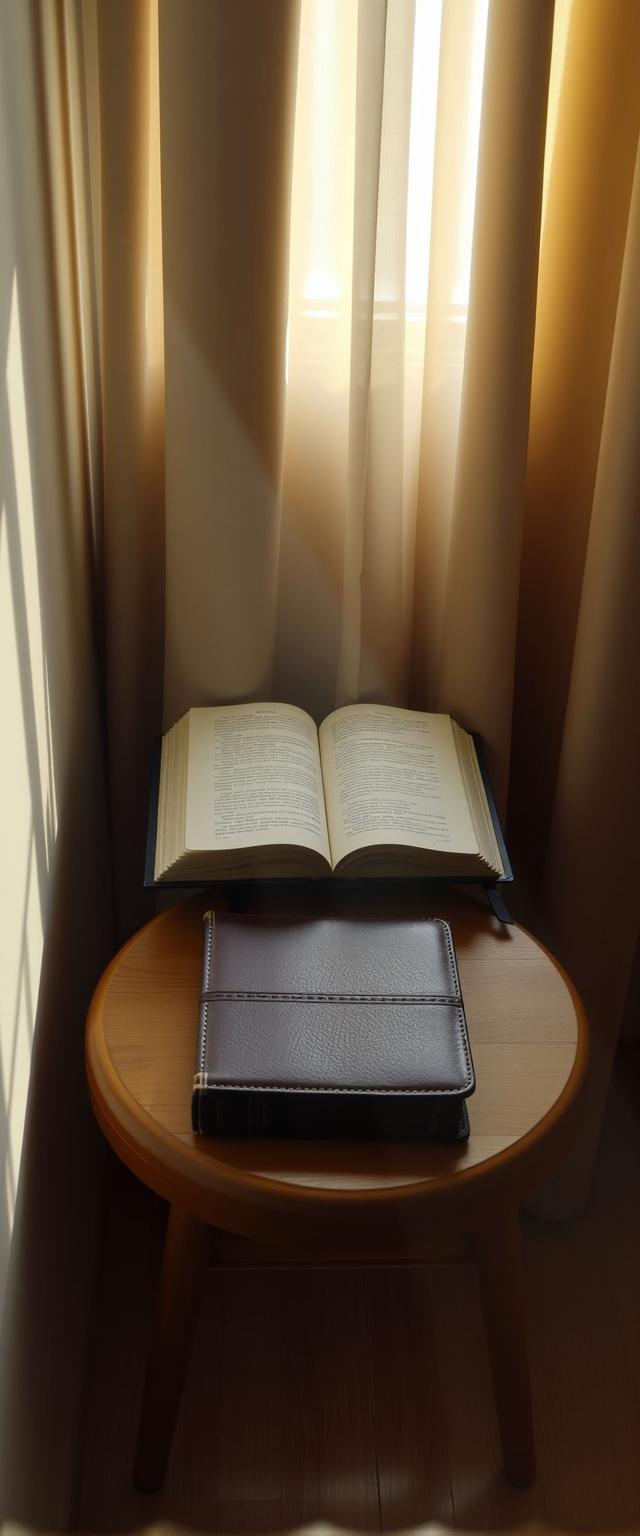}}
      & \makecell[c]{$1088\times960$} & \makecell[c]{\num{46207}} & \makecell[c]{\includegraphics[height=12mm]{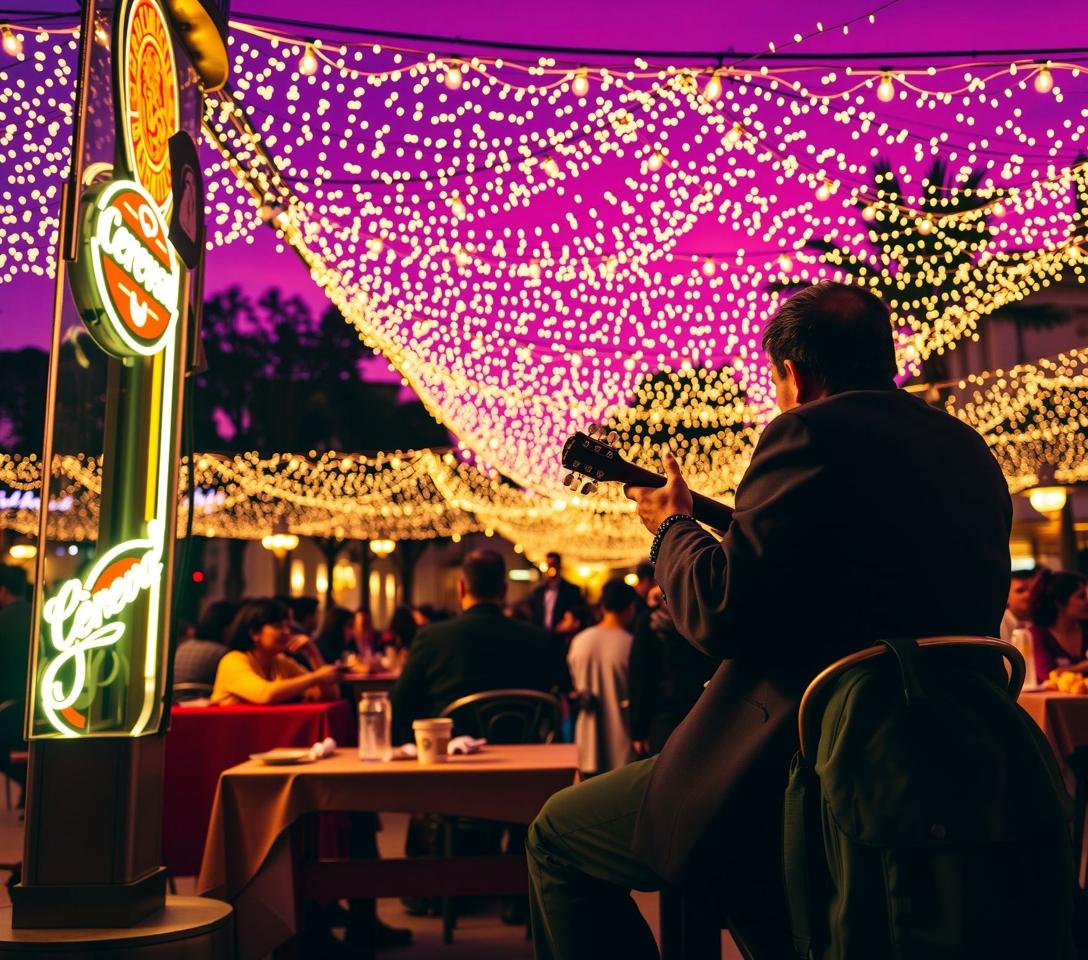}} \\
      
      \makecell[c]{$704\times1472$} & \makecell[c]{\num{11305}} & \makecell[c]{\includegraphics[height=12mm]{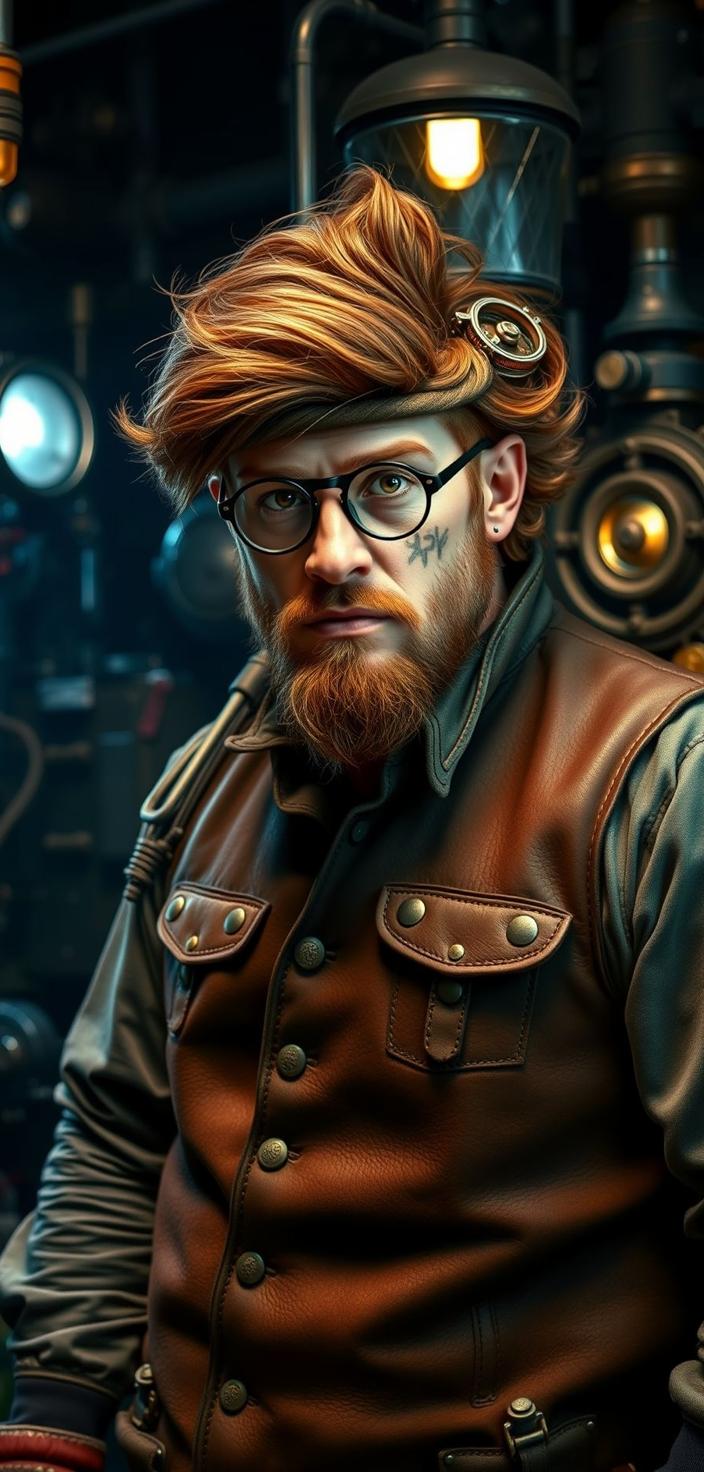}}
      & \makecell[c]{$1088\times896$} & \makecell[c]{\num{40009}} & \makecell[c]{\includegraphics[height=12mm]{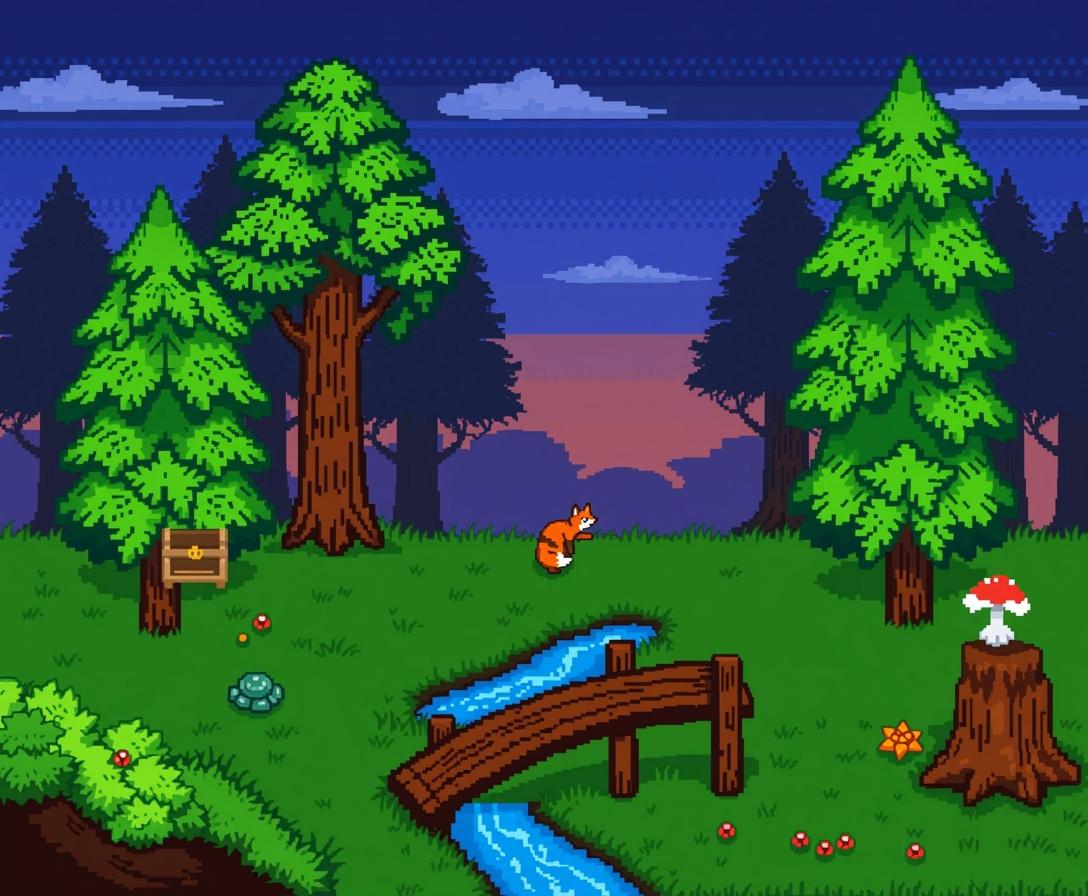}} \\
      
      \makecell[c]{$704\times1408$} & \makecell[c]{\num{15405}} & \makecell[c]{\includegraphics[height=12mm]{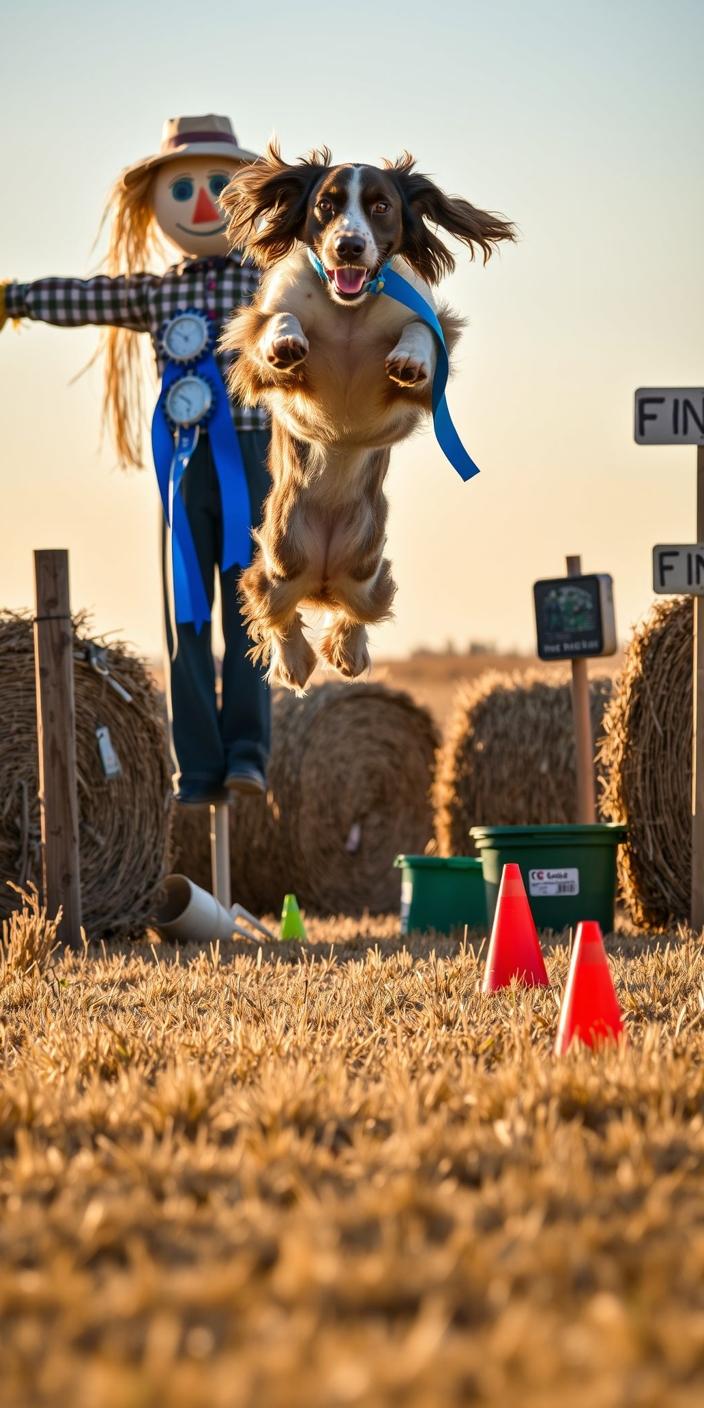}}
      & \makecell[c]{$1152\times896$} & \makecell[c]{\num{36385}} & \makecell[c]{\includegraphics[height=12mm]{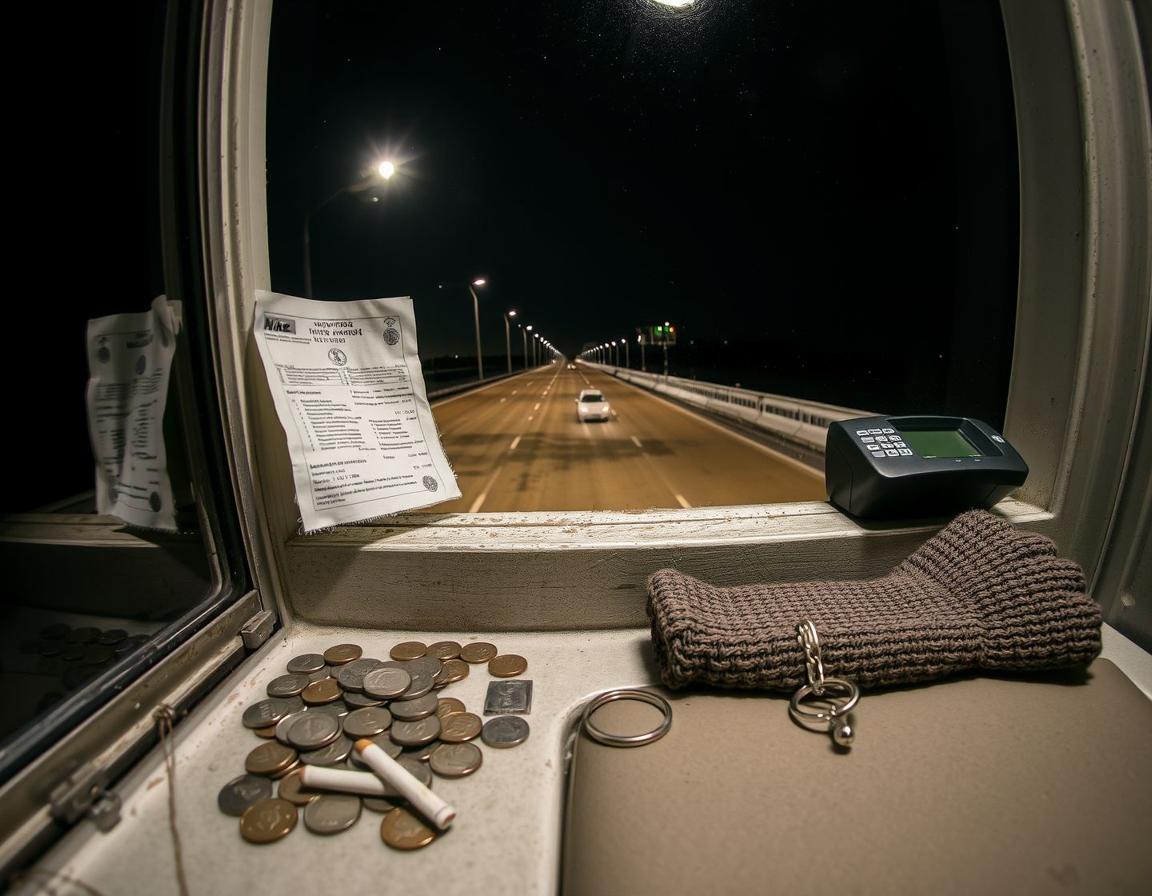}} \\
      
      \makecell[c]{$768\times1344$} & \makecell[c]{\num{23592}} & \makecell[c]{\includegraphics[height=12mm]{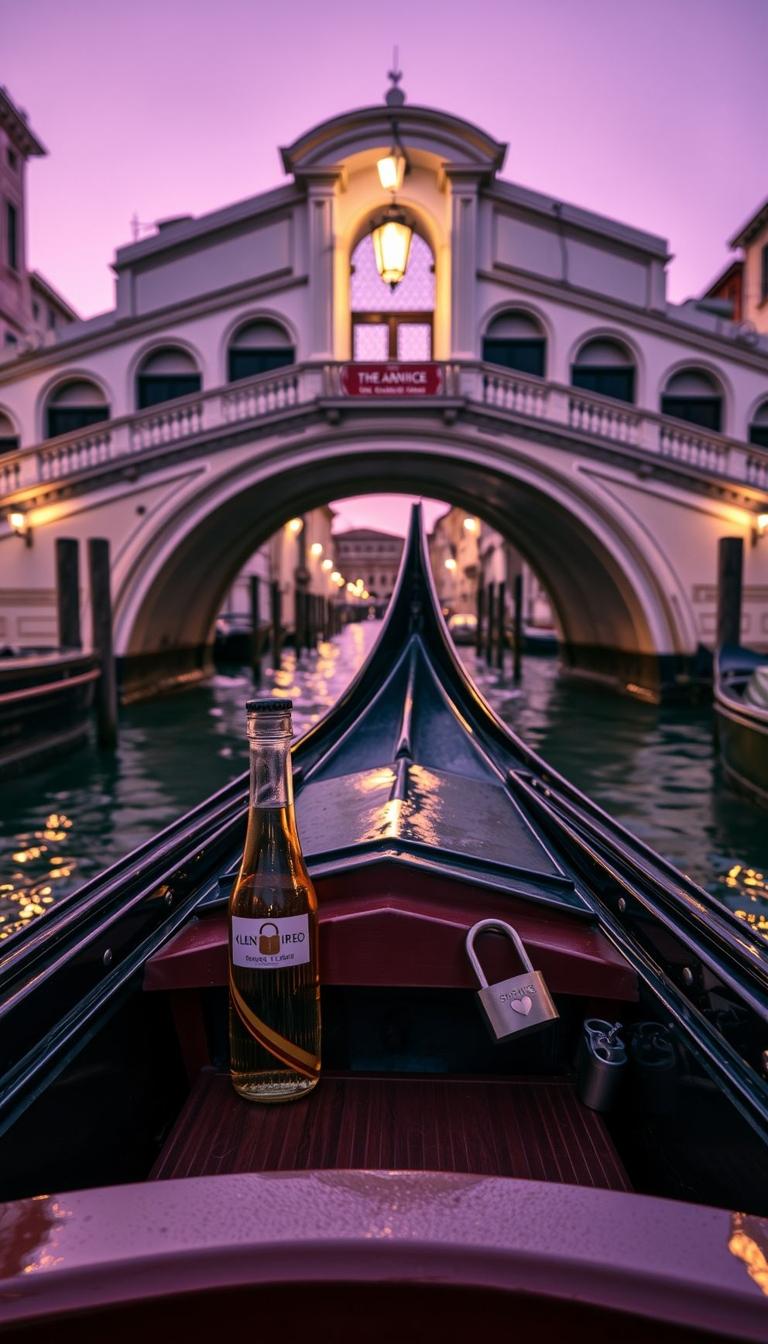}}
      & \makecell[c]{$1152\times832$} & \makecell[c]{\num{38090}} & \makecell[c]{\includegraphics[height=12mm]{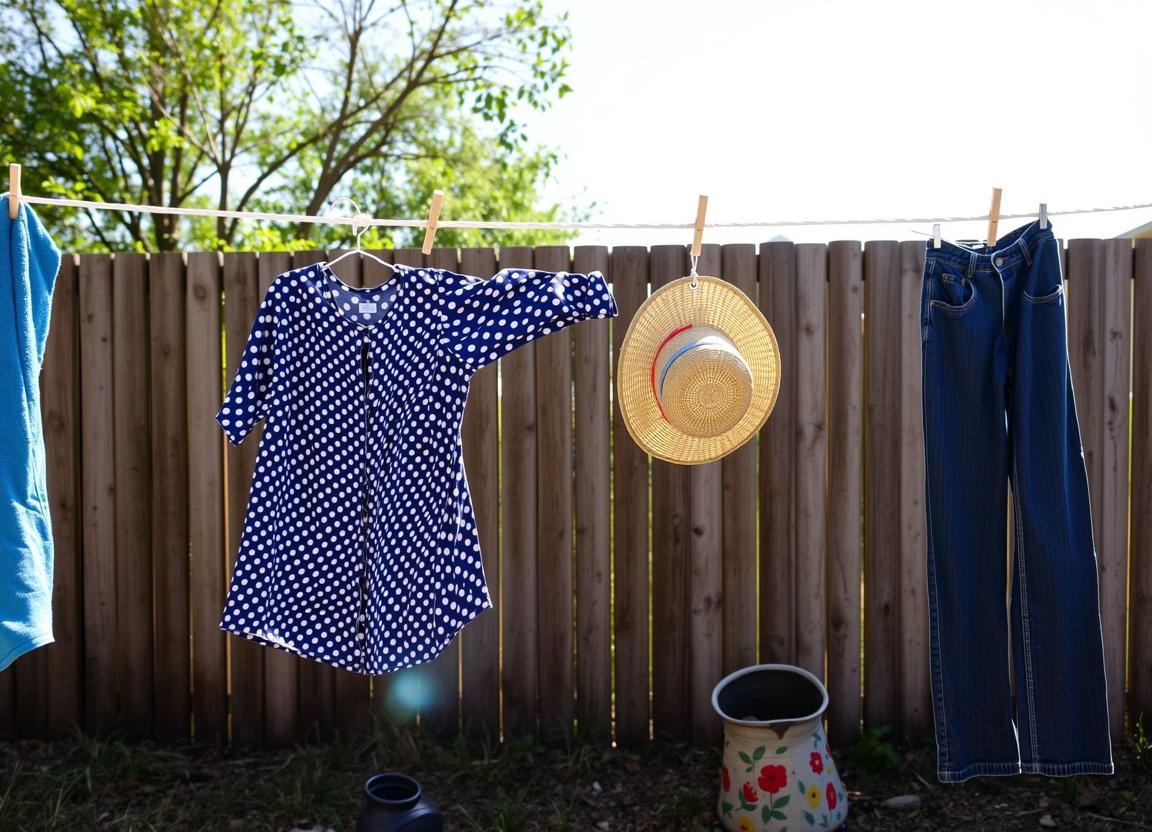}} \\
      
      \makecell[c]{$768\times1280$} & \makecell[c]{\num{30533}} & \makecell[c]{\includegraphics[height=12mm]{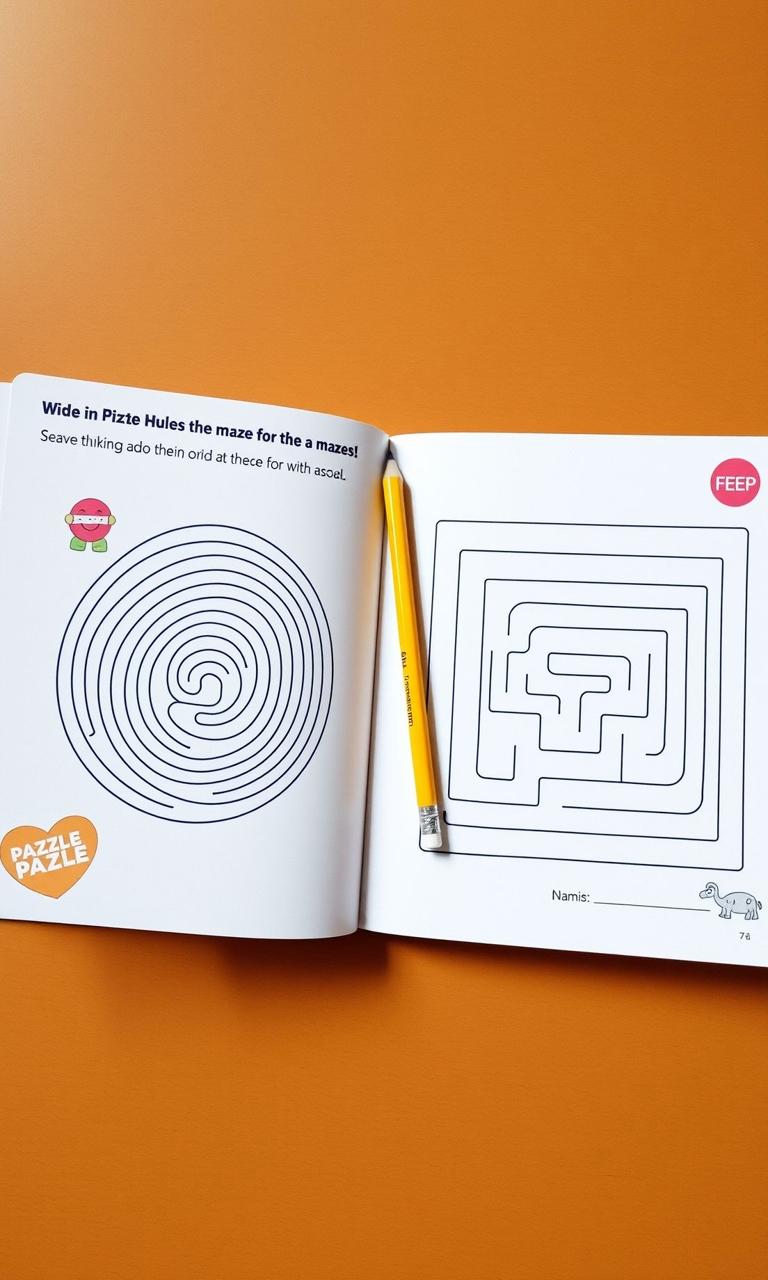}}
      & \makecell[c]{$1216\times832$} & \makecell[c]{\num{41537}} & \makecell[c]{\includegraphics[height=12mm]{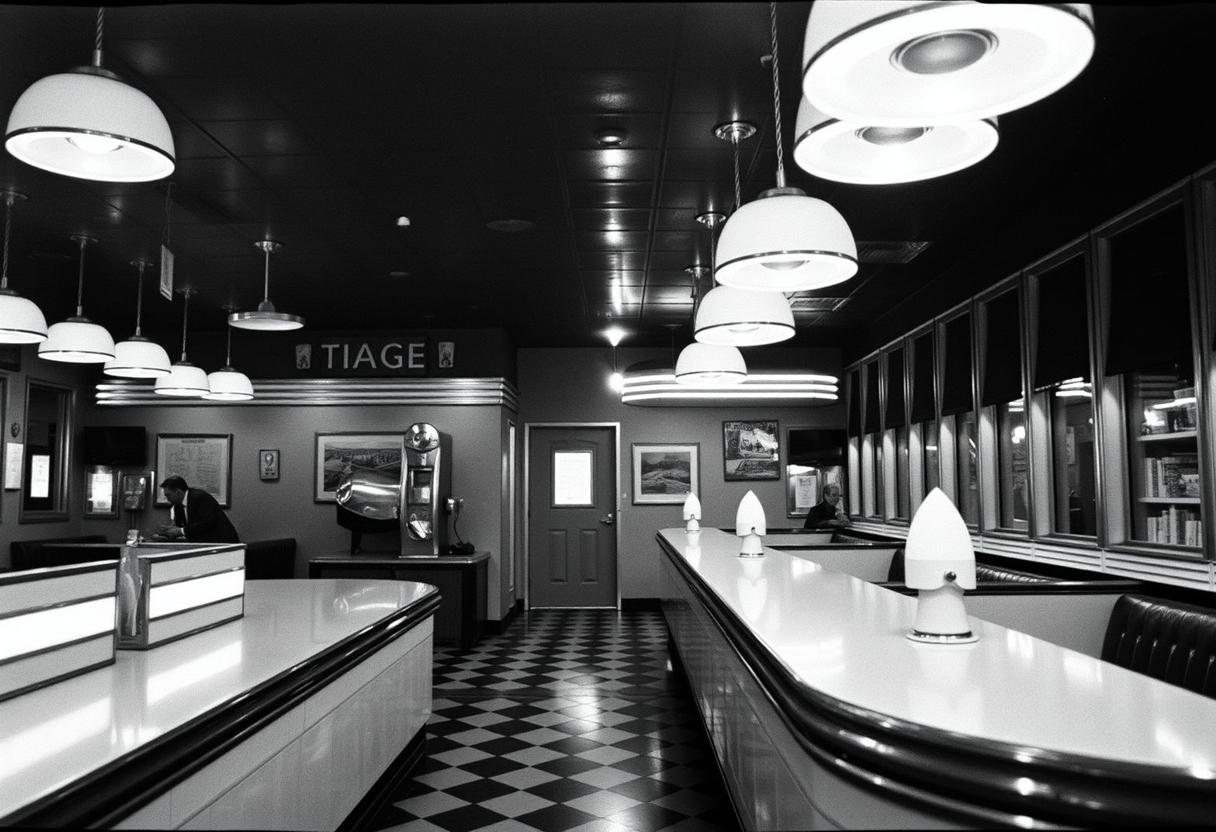}} \\
      
      \makecell[c]{$832\times1216$} & \makecell[c]{\num{43426}} & \makecell[c]{\includegraphics[height=12mm]{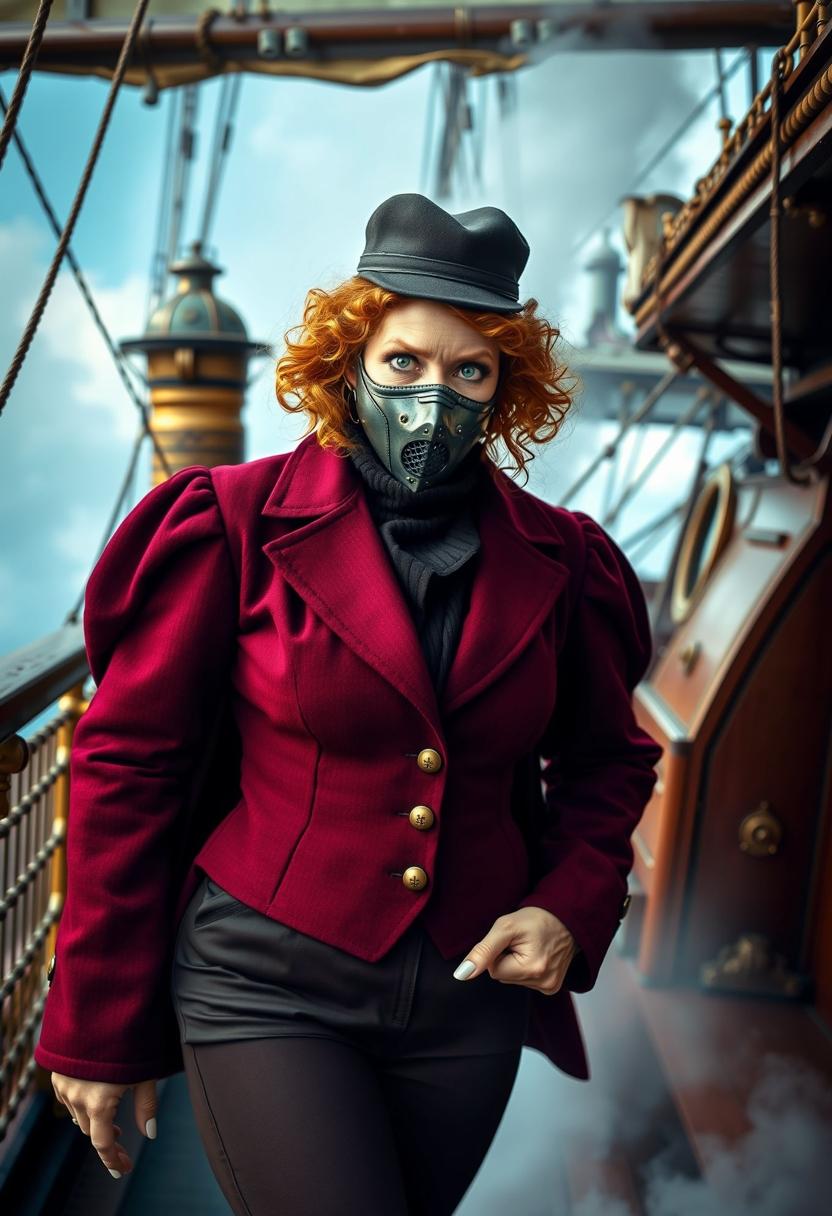}}
      & \makecell[c]{$1280\times768$} & \makecell[c]{\num{34457}} & \makecell[c]{\includegraphics[height=12mm]{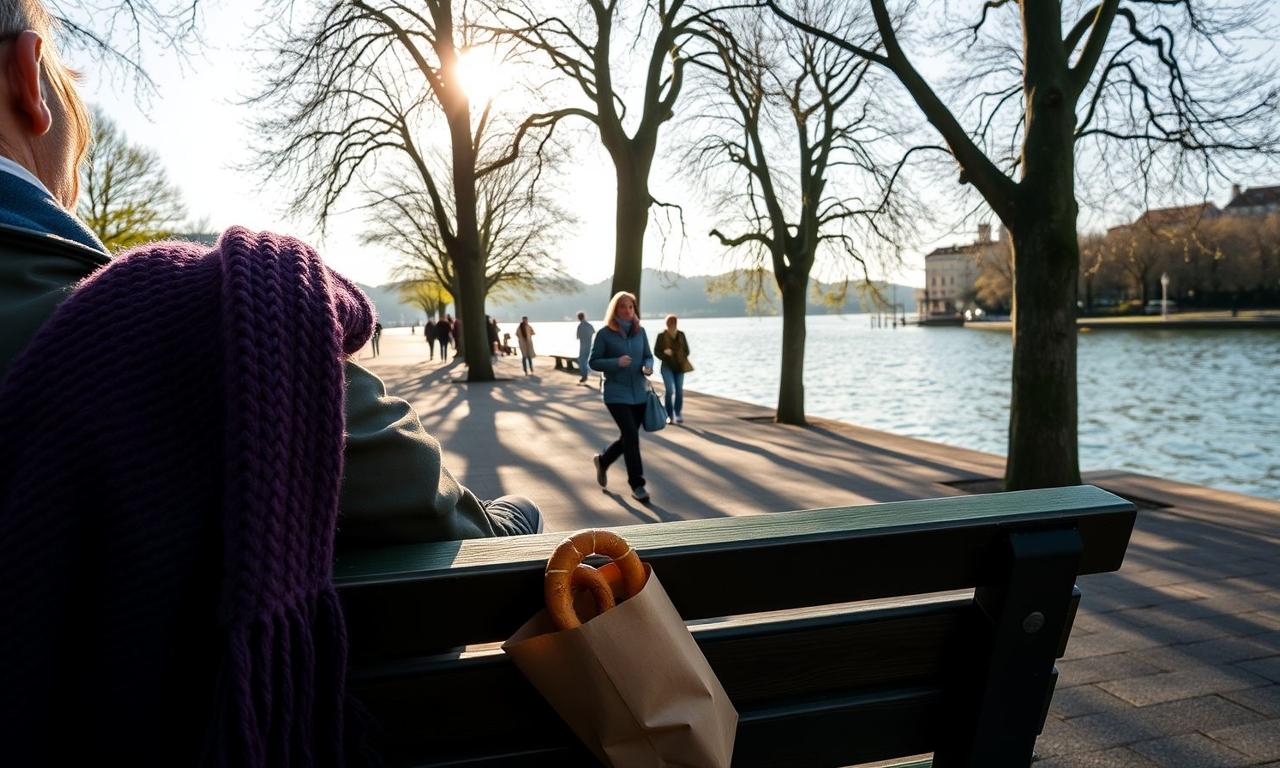}} \\
      
      \makecell[c]{$832\times1152$} & \makecell[c]{\num{32434}} & \makecell[c]{\includegraphics[height=12mm]{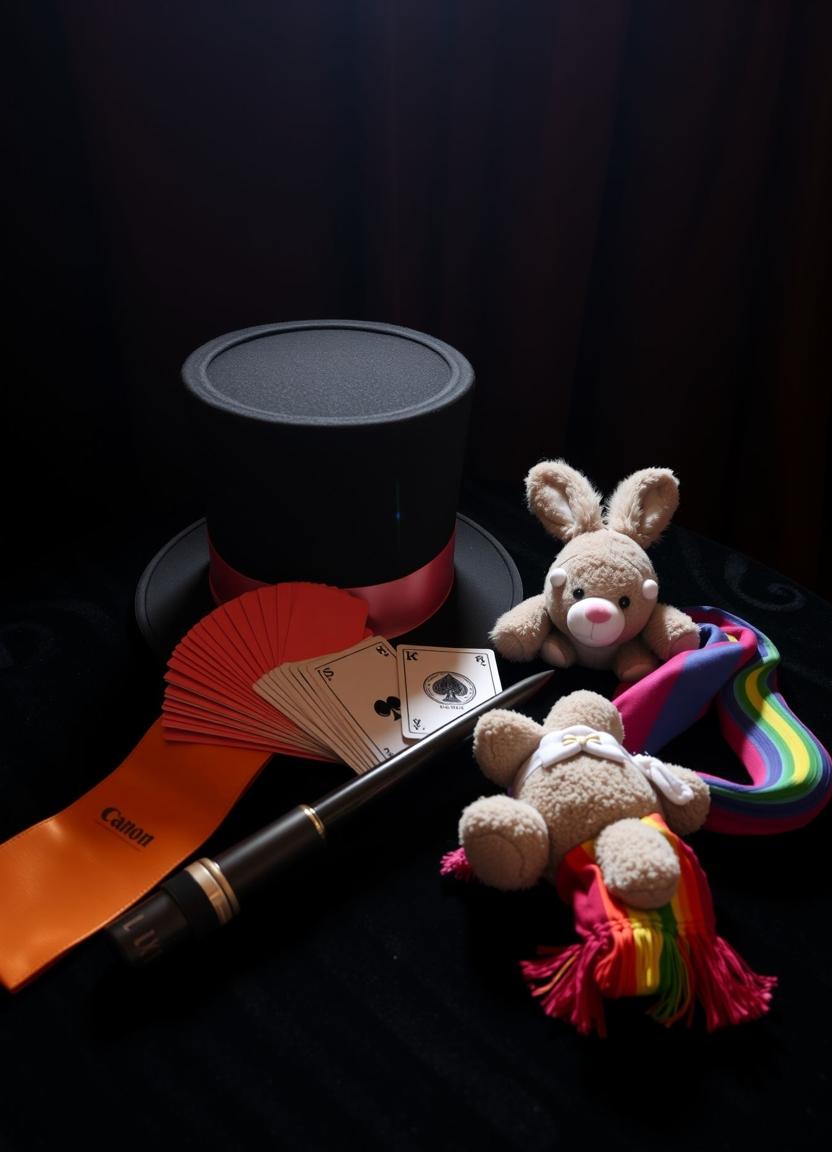}}
      & \makecell[c]{$1344\times768$} & \makecell[c]{\num{21250}} & \makecell[c]{\includegraphics[height=12mm]{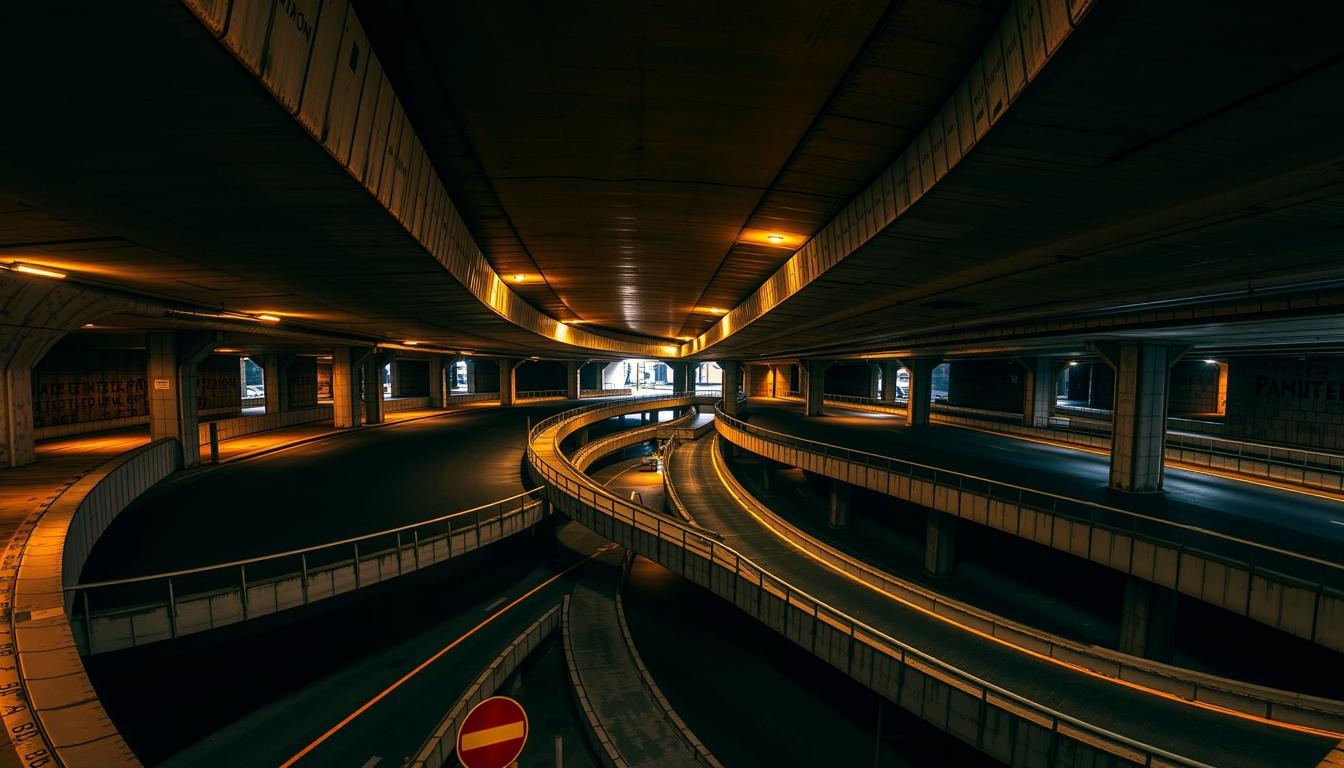}} \\
      
      \makecell[c]{$896\times1152$} & \makecell[c]{\num{37731}} & \makecell[c]{\includegraphics[height=12mm]{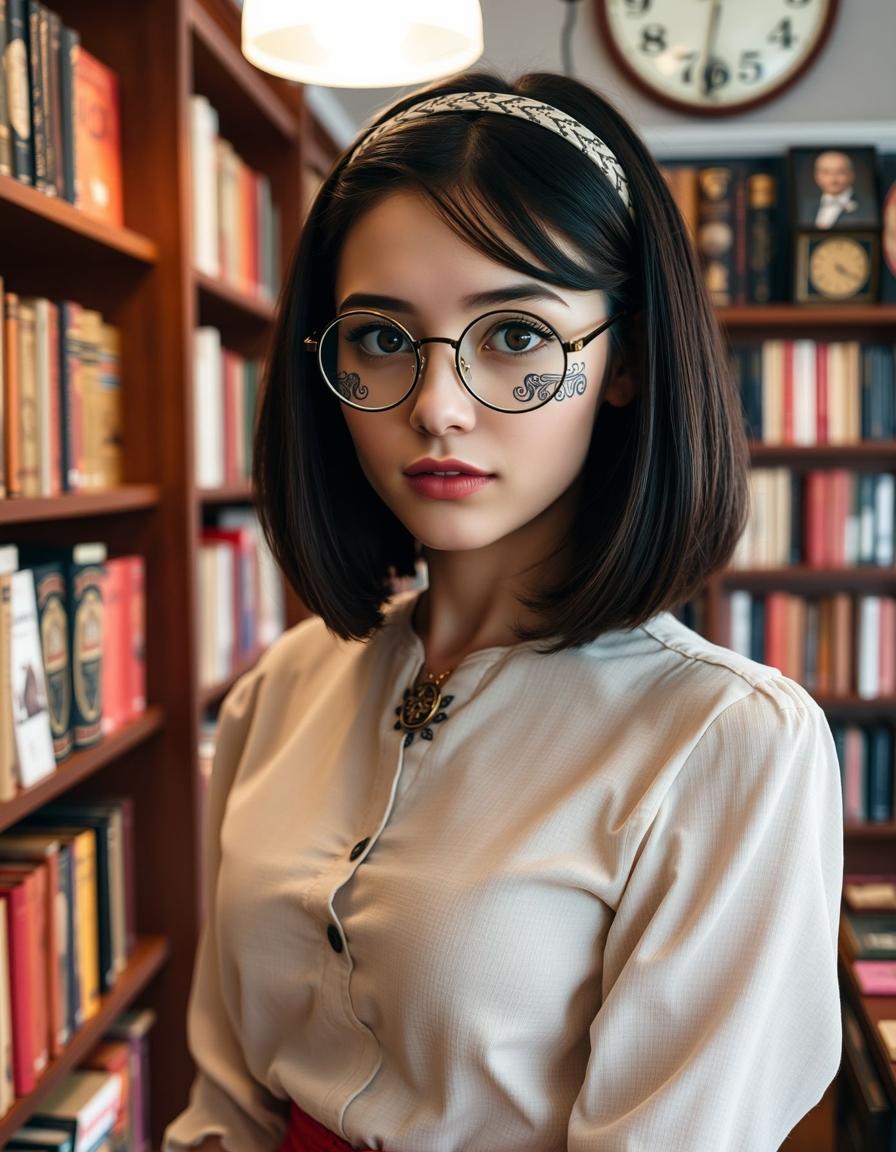}}
      & \makecell[c]{$1344\times704$} & \makecell[c]{\num{15783}} & \makecell[c]{\includegraphics[height=12mm]{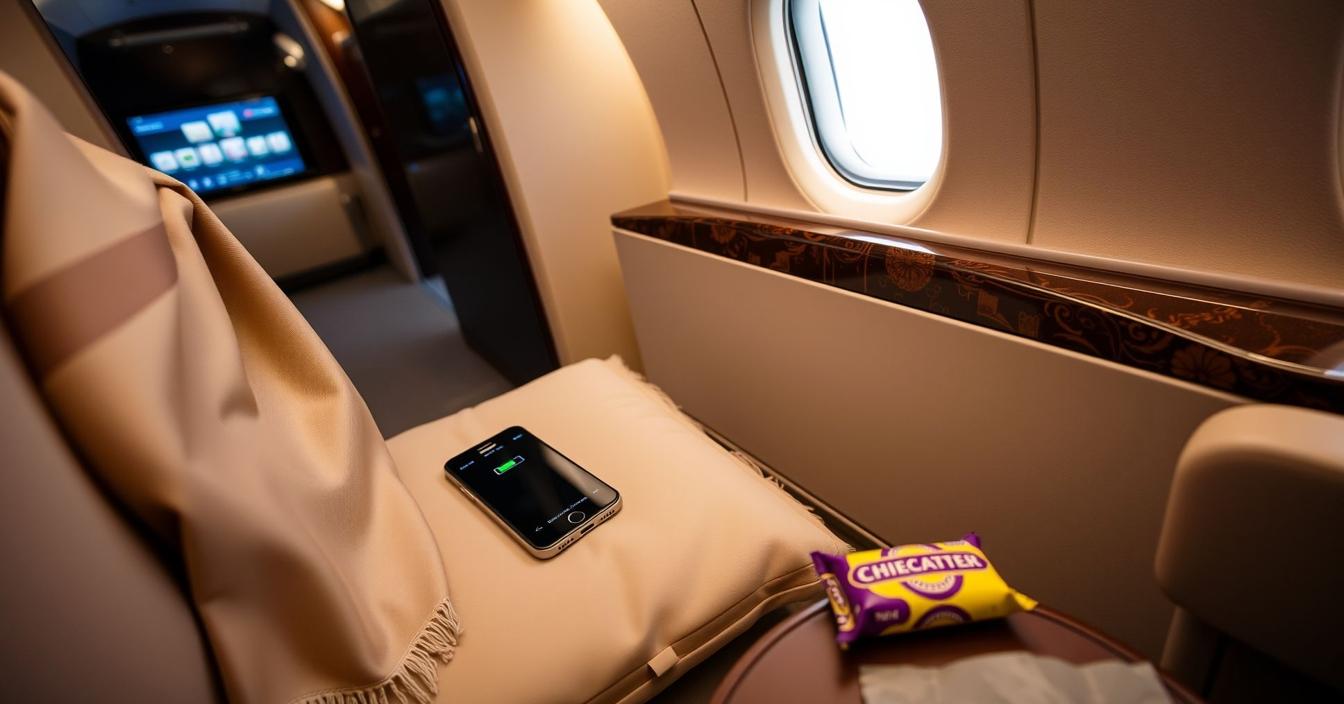}} \\
      
      \makecell[c]{$896\times1088$} & \makecell[c]{\num{43759}} & \makecell[c]{\includegraphics[height=12mm]{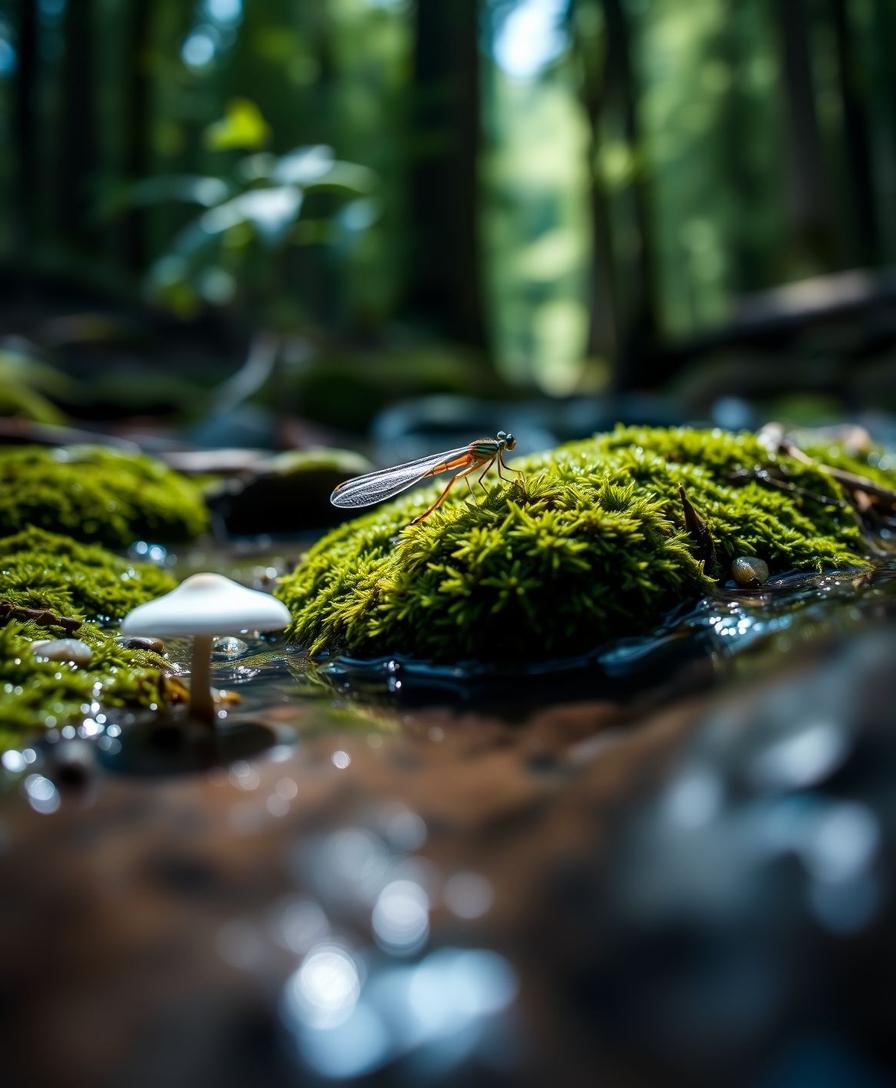}}
      & \makecell[c]{$1408\times704$} & \makecell[c]{\num{7302}} & \makecell[c]{\includegraphics[height=12mm]{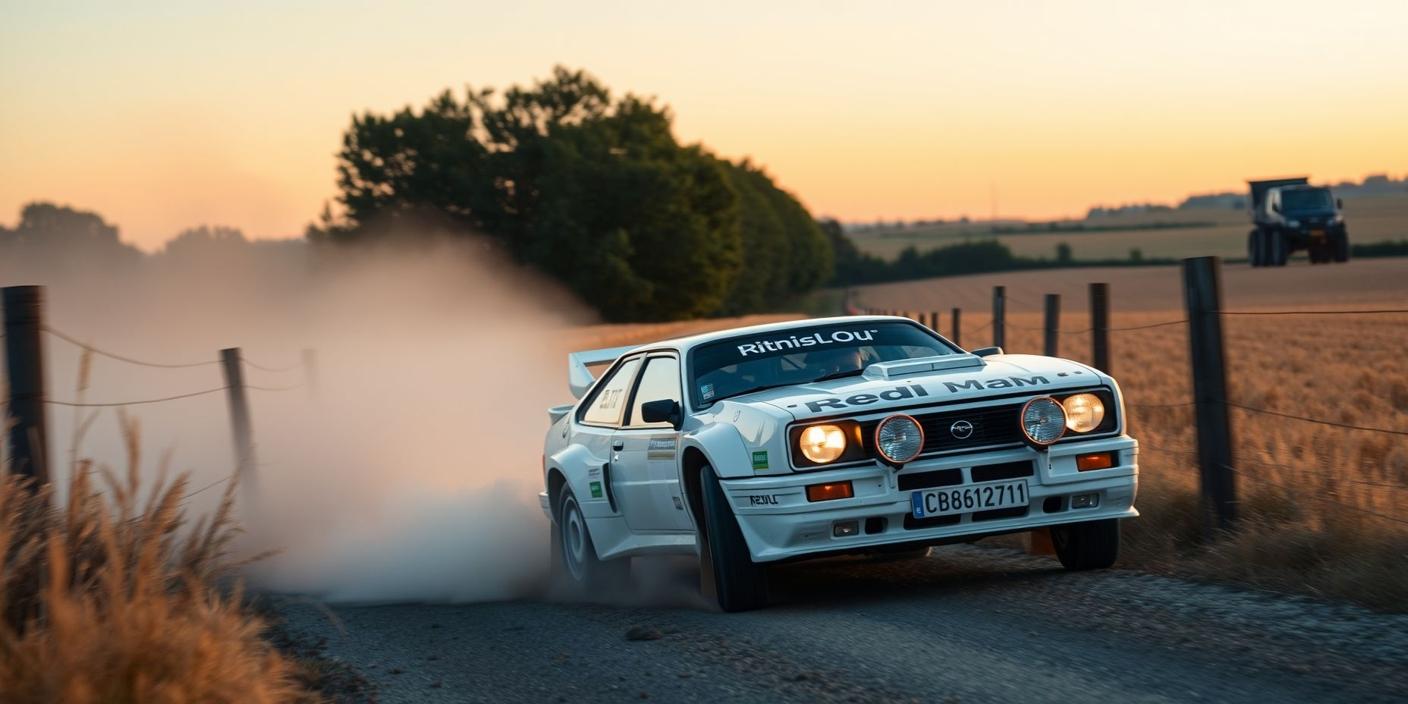}} \\
      
      \makecell[c]{$960\times1088$} & \makecell[c]{\num{42763}} & \makecell[c]{\includegraphics[height=12mm]{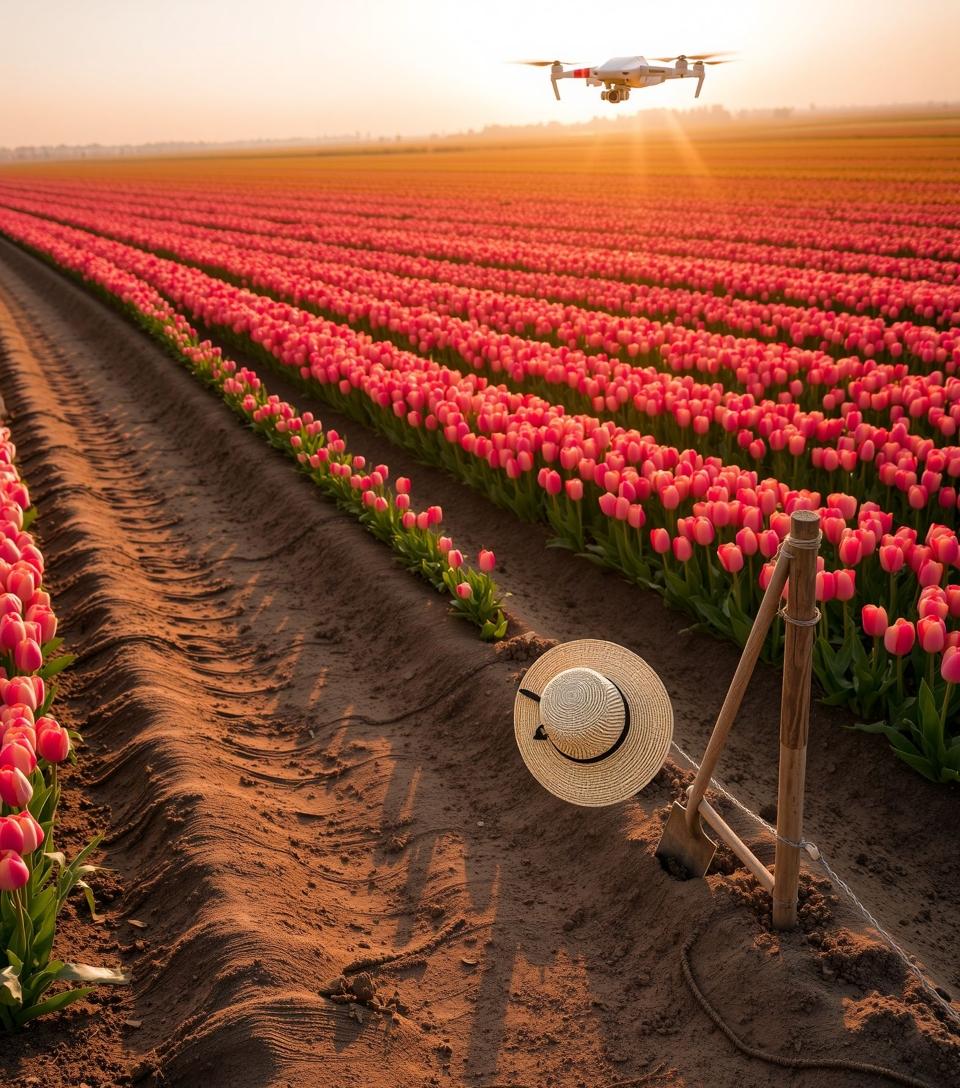}}
      & \makecell[c]{$1472\times704$} & \makecell[c]{\num{11980}} & \makecell[c]{\includegraphics[height=12mm]{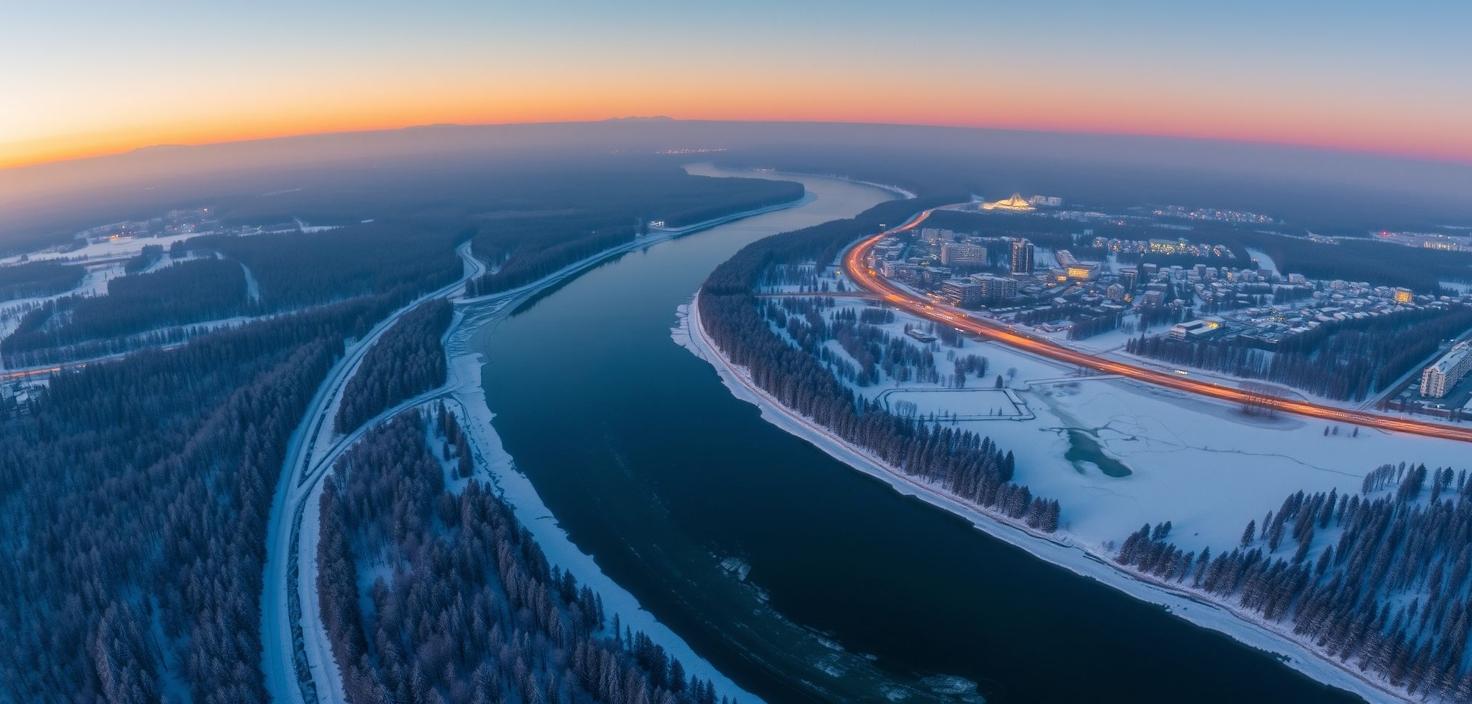}} \\
      
      \makecell[c]{$960\times1024$} & \makecell[c]{\num{42502}} & \makecell[c]{\includegraphics[height=12mm]{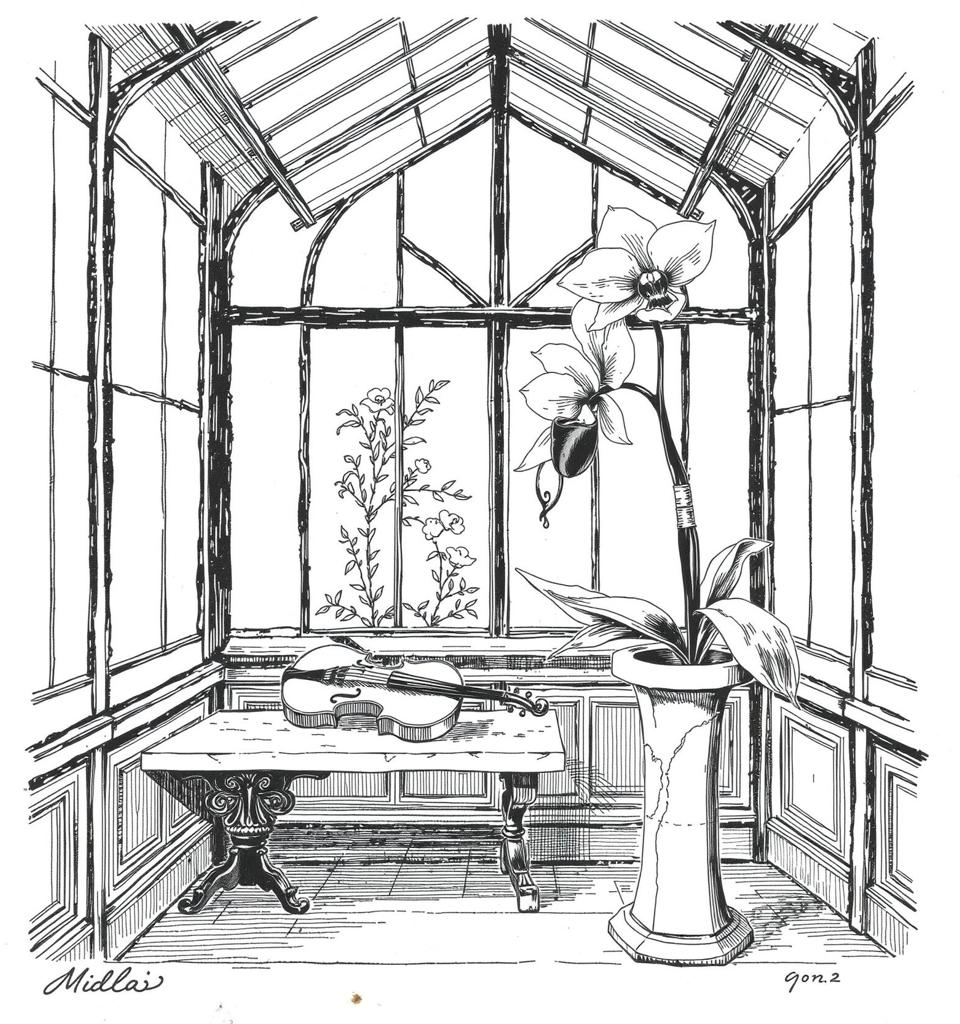}}
      & \makecell[c]{$1536\times640$} & \makecell[c]{\num{6182}} & \makecell[c]{\includegraphics[height=12mm]{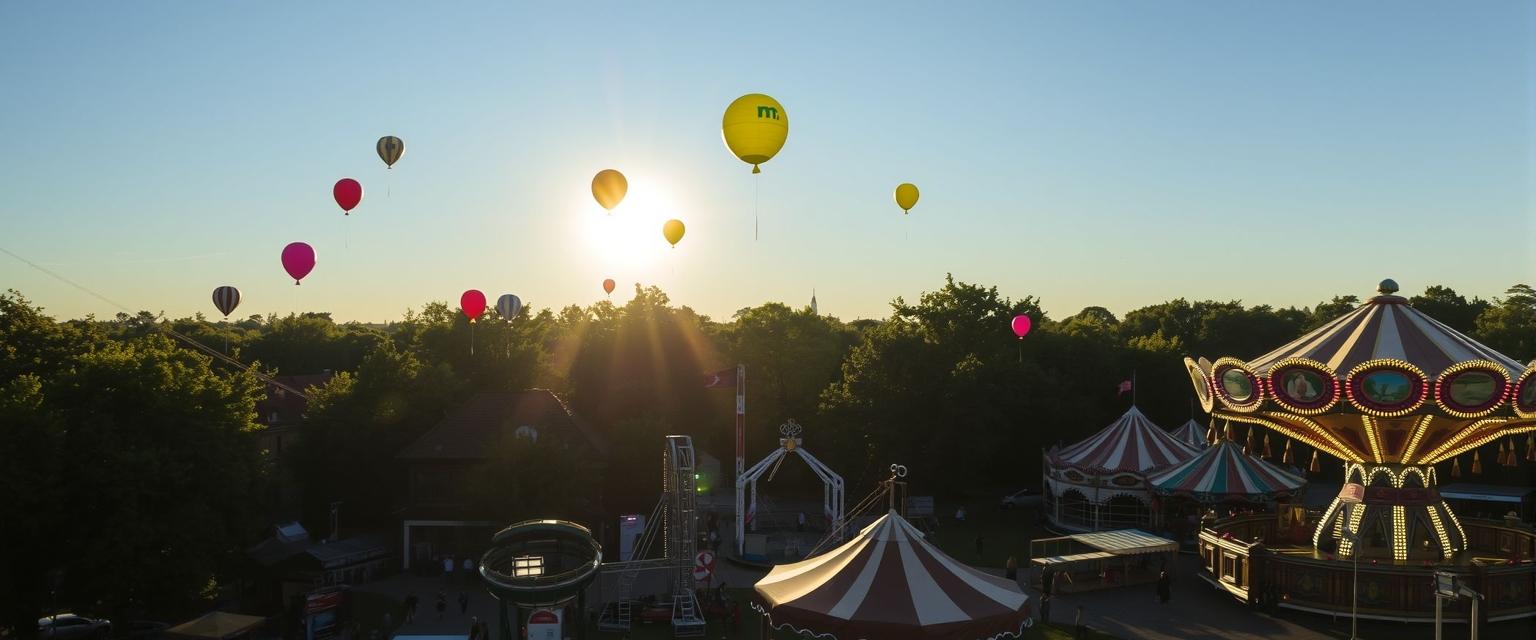}} \\
      
      \makecell[c]{$1024\times1024$} & \makecell[c]{\num{46619}} & \makecell[c]{\includegraphics[height=12mm]{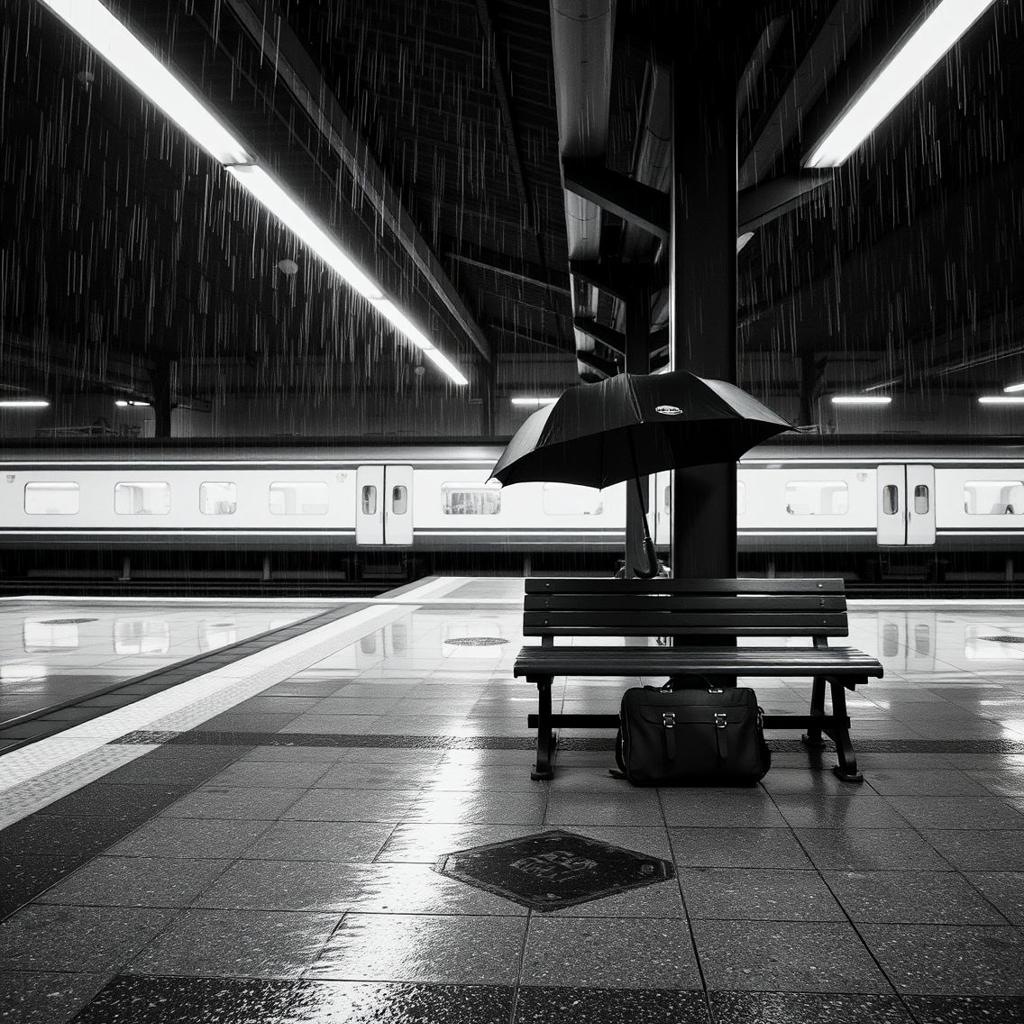}}
      & \makecell[c]{$1600\times640$} & \makecell[c]{\num{805}} & \makecell[c]{\includegraphics[height=12mm]{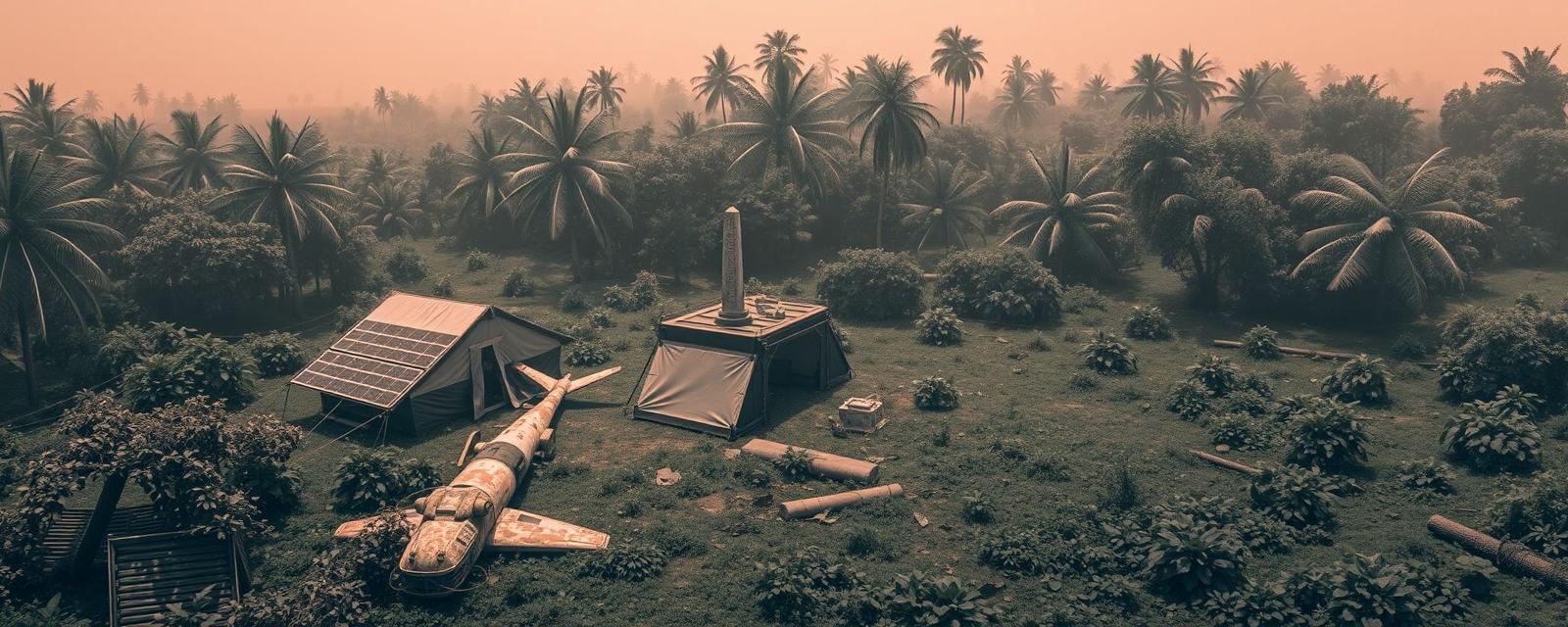}} \\
      
      \bottomrule
    \end{tabular}
  \label{tab:ar_distr}
\end{table*}

\begin{figure*}[!htbp]
  \centering
  \includegraphics[width=\linewidth]{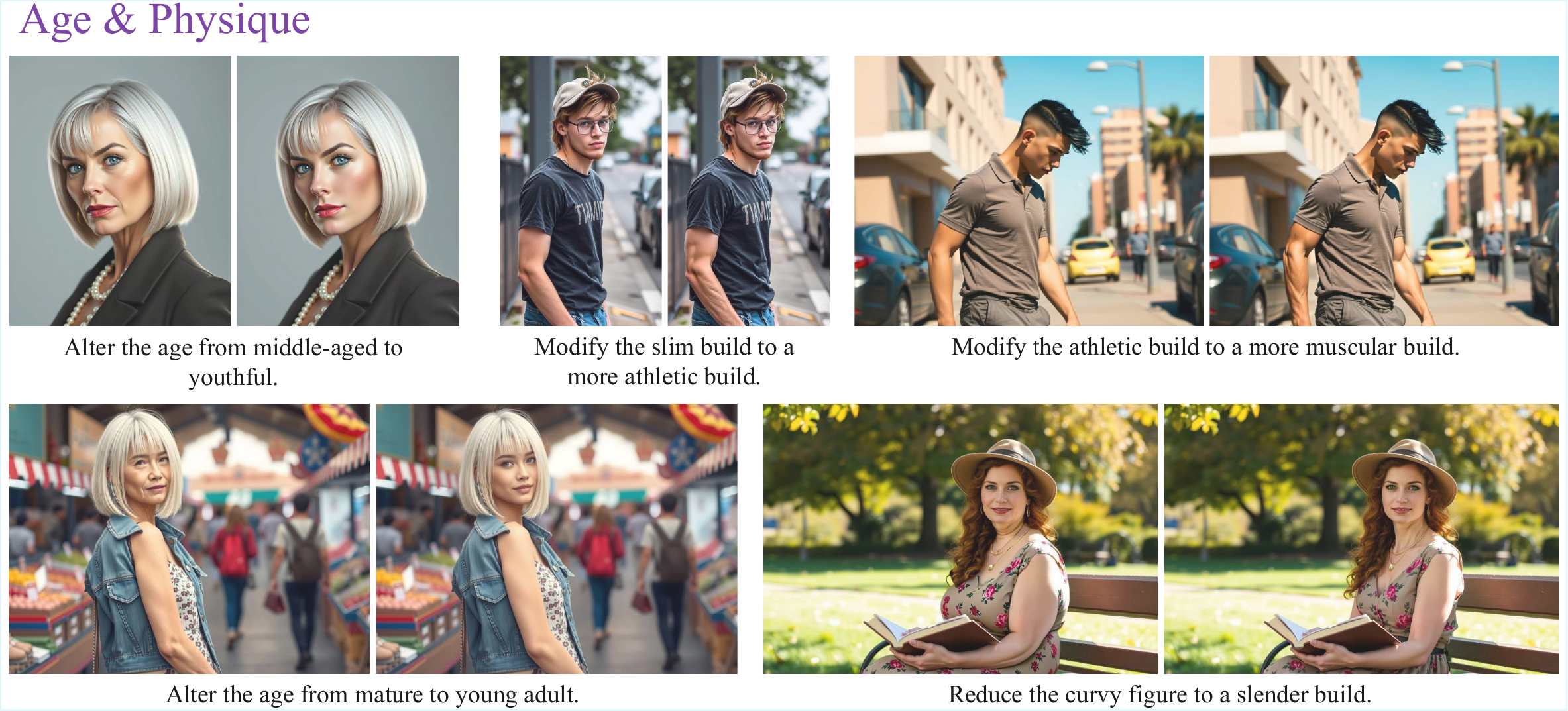}
  \caption{Edits involving age and physique transformations.}
  \label{fig:collage1}
\end{figure*}
\begin{figure*}[!htbp]
  \centering
  \includegraphics[width=\linewidth]{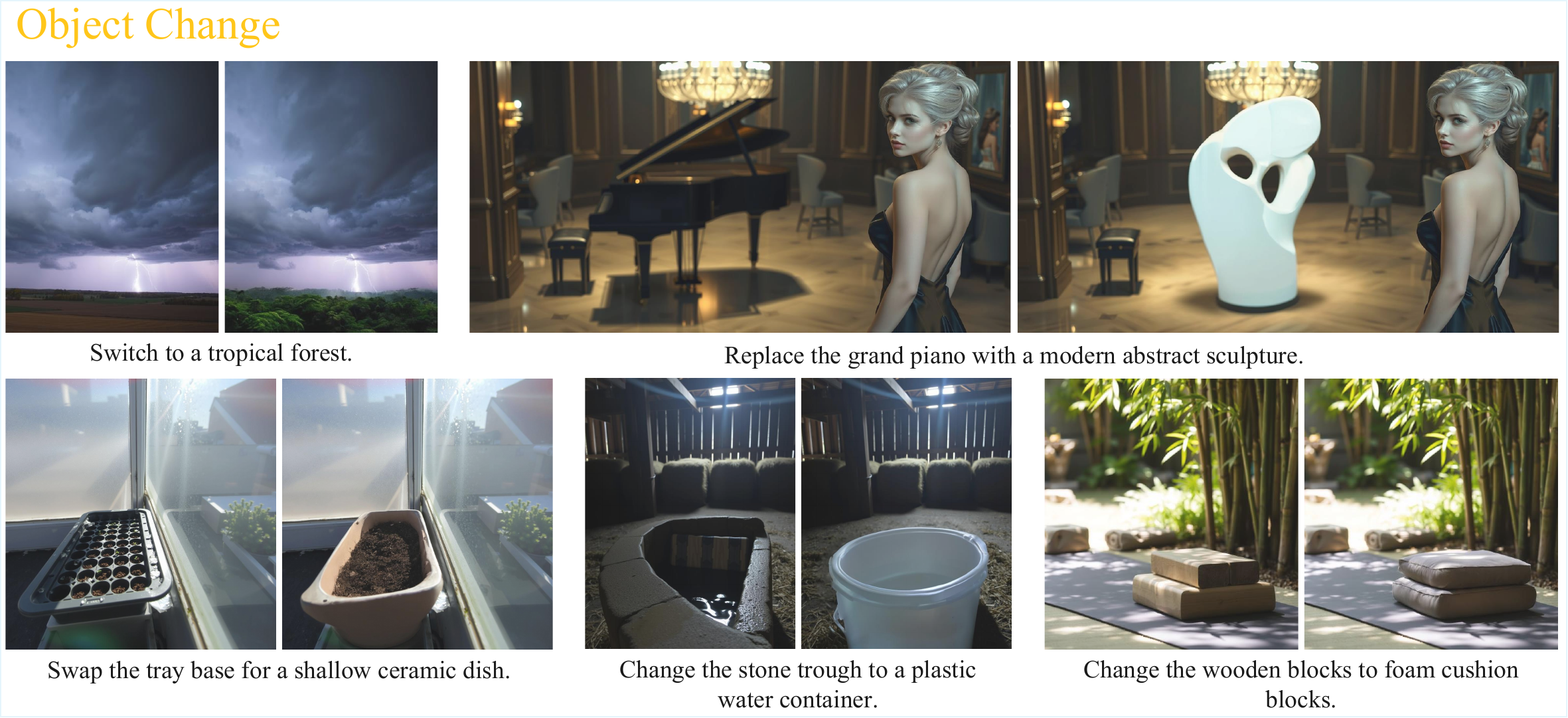}
  \caption{Edits dedicated to subject change.}
  \label{fig:collage3}
\end{figure*}
\begin{figure*}[!htbp]
  \centering
  \includegraphics[width=\linewidth]{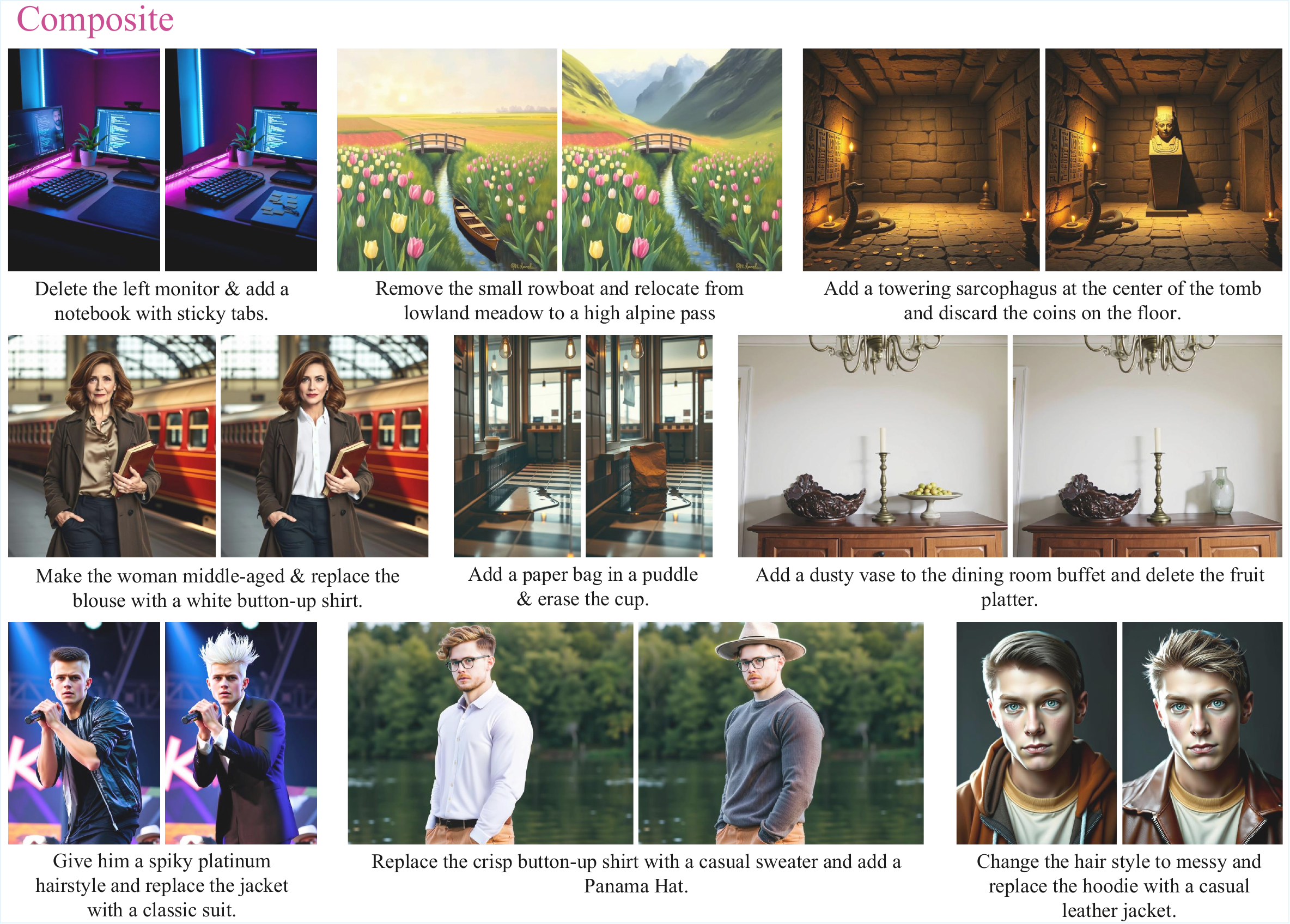}
  \caption{Composite edits with more than one change.}
  \label{fig:collage2}
\end{figure*}
\begin{figure*}[!htbp]
  \centering
  \includegraphics[width=\linewidth]{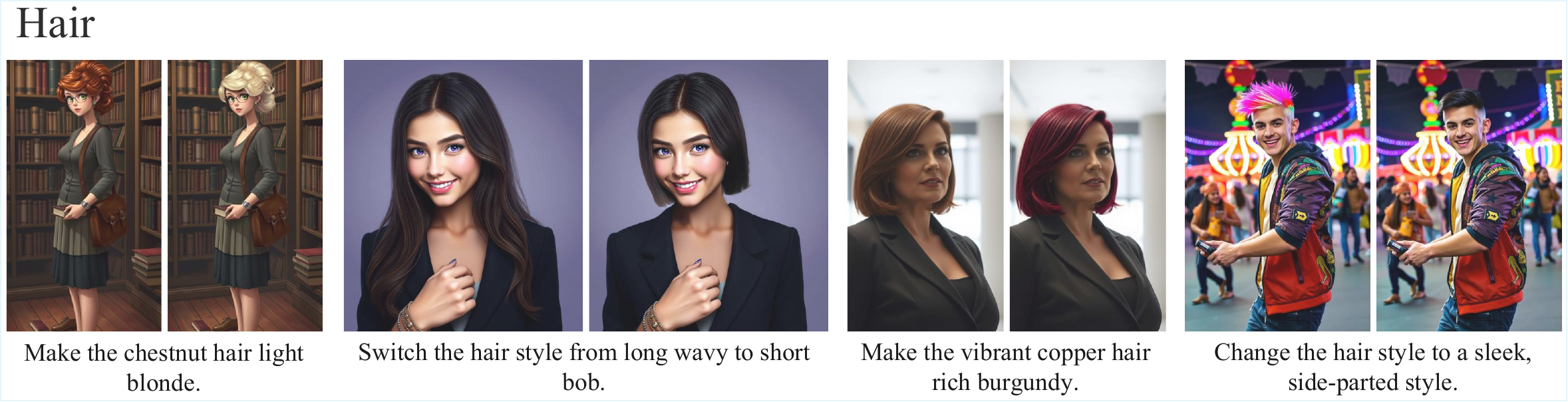}
  \caption{Examples if human hair changes.}
  \label{fig:collage6}
\end{figure*}
\begin{figure*}[!htbp]
  \centering
  \includegraphics[width=\linewidth]{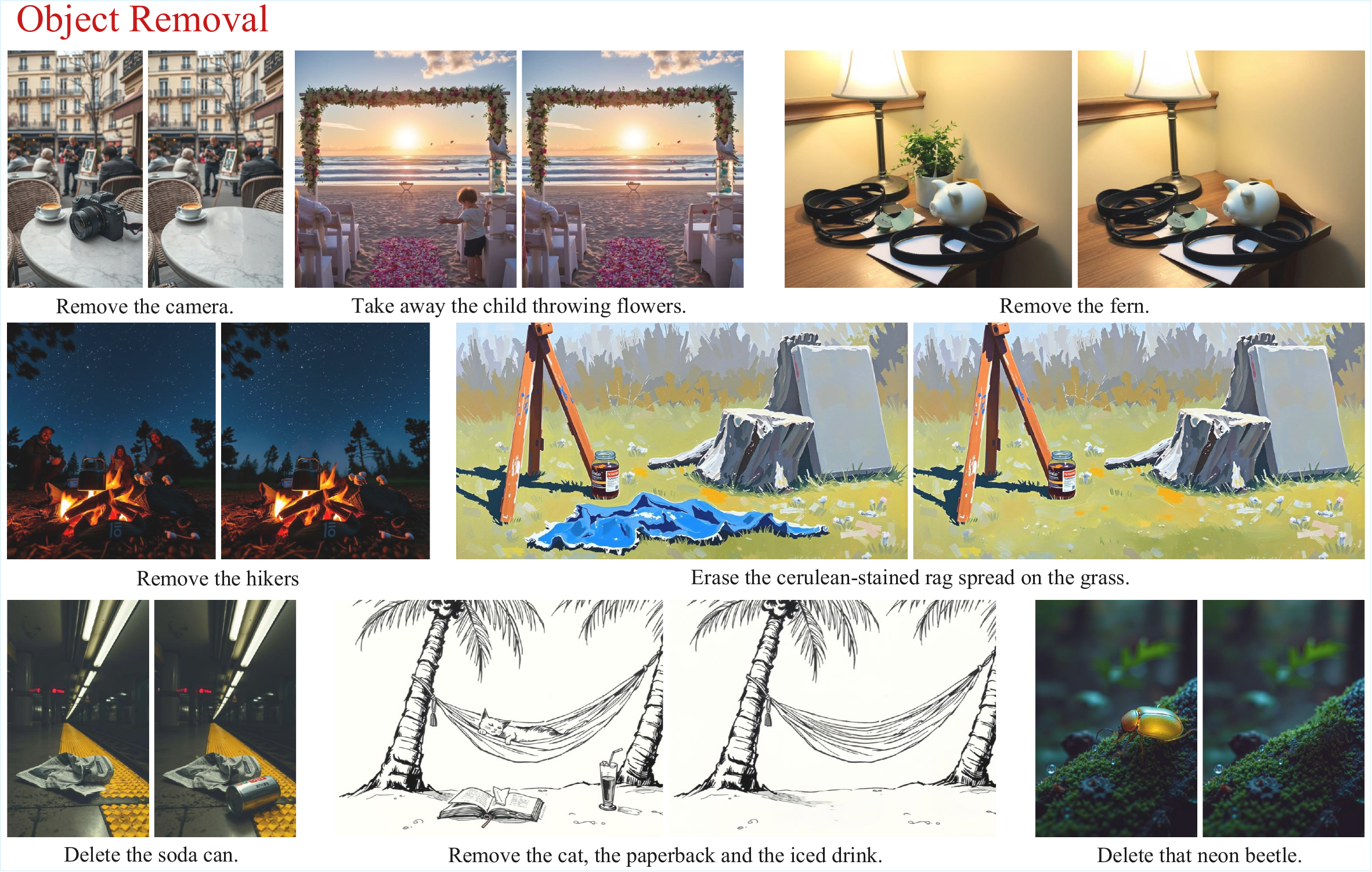}
  \caption{Edits dedicated to subject deletion operation.}
  \label{fig:collage4}
\end{figure*}
\begin{figure*}[!htbp]
  \centering
  \includegraphics[width=\linewidth]{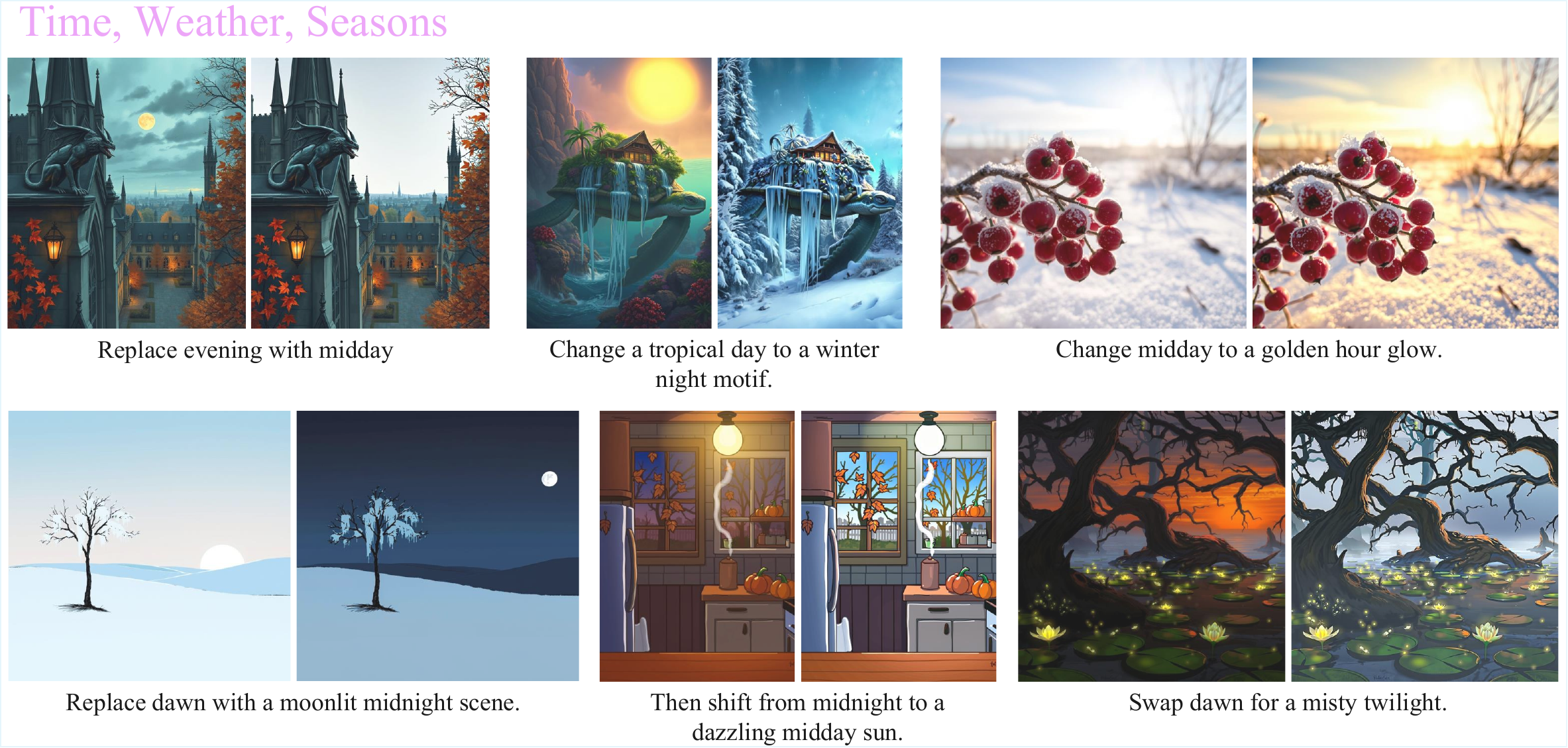}
  \caption{Showcases of global edits, they require to change a majority of image while preserving subjects identity from changes.}
  \label{fig:collage5}
\end{figure*}
\begin{figure*}[!htbp]
  \centering
  \includegraphics[width=\linewidth]{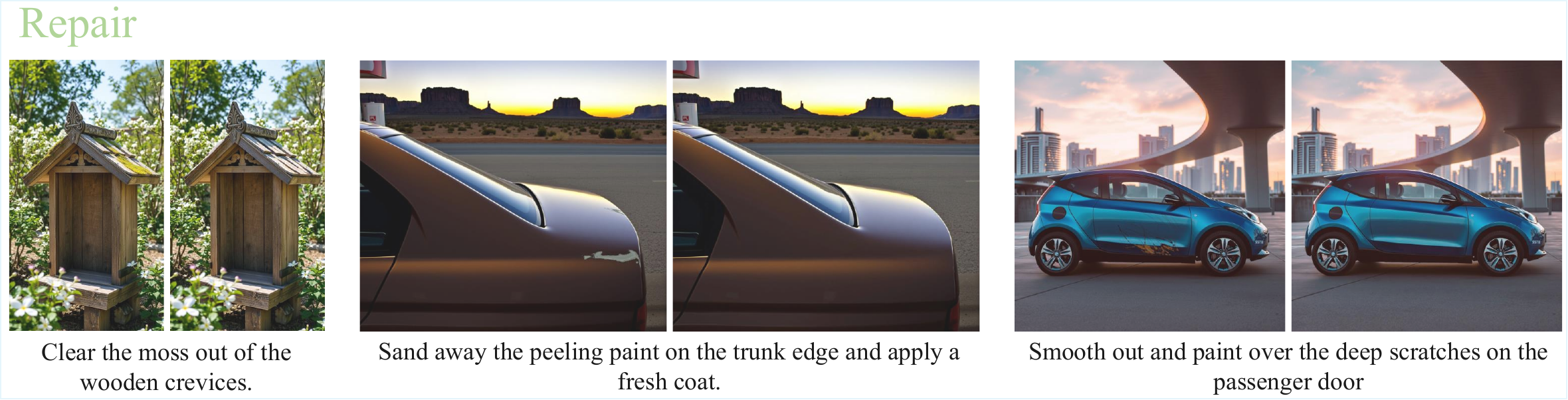}
  \caption{Object condition restoration cases.}
  \label{fig:collage7}
\end{figure*}
\begin{figure*}[!htbp]
  \centering
  \includegraphics[width=\linewidth]{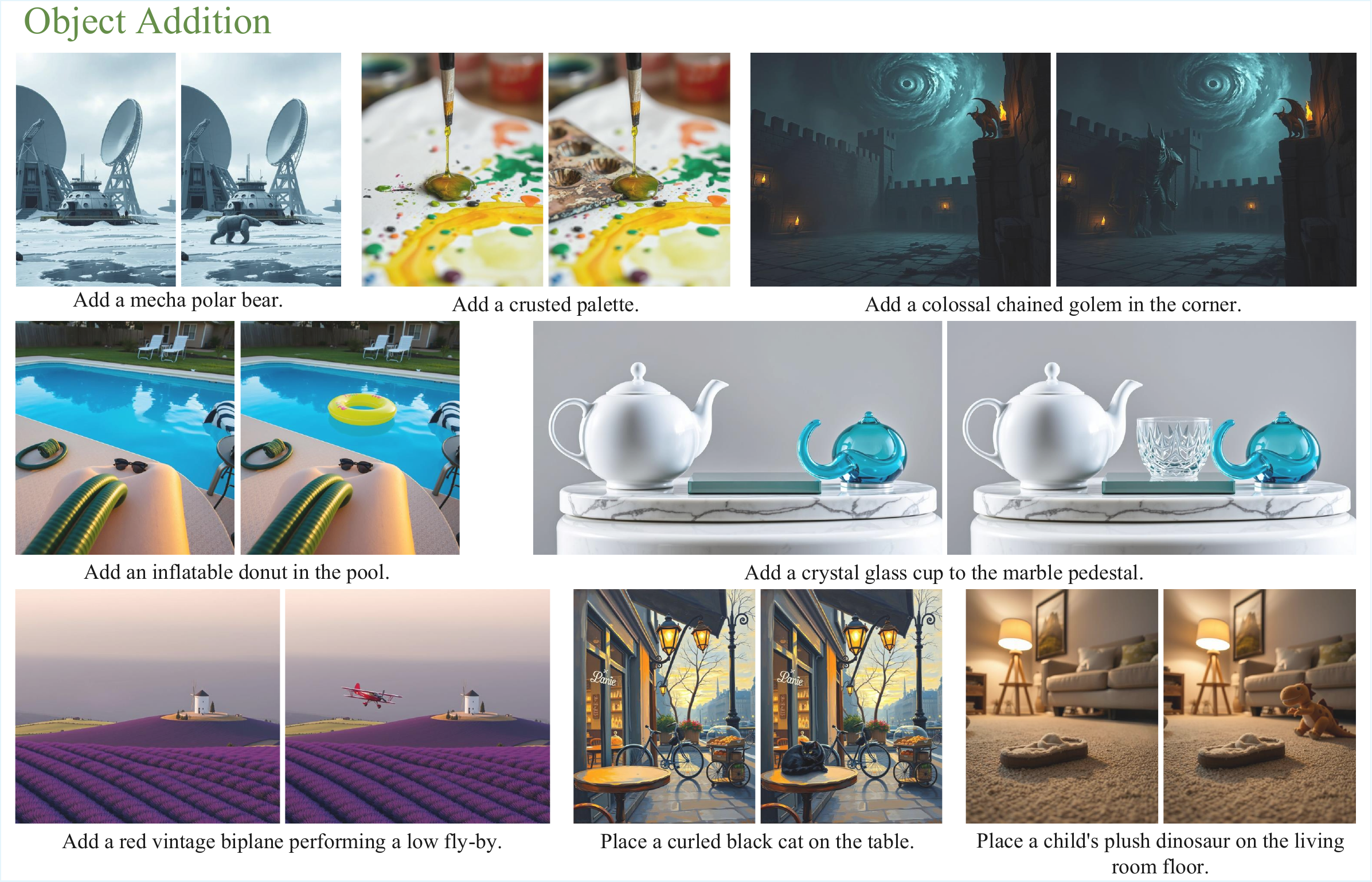}
  \caption{Introducing new objects and placing them harmonically.}
  \label{fig:collage12}
\end{figure*}
\begin{figure*}[!htbp]
  \centering
  \includegraphics[width=\linewidth]{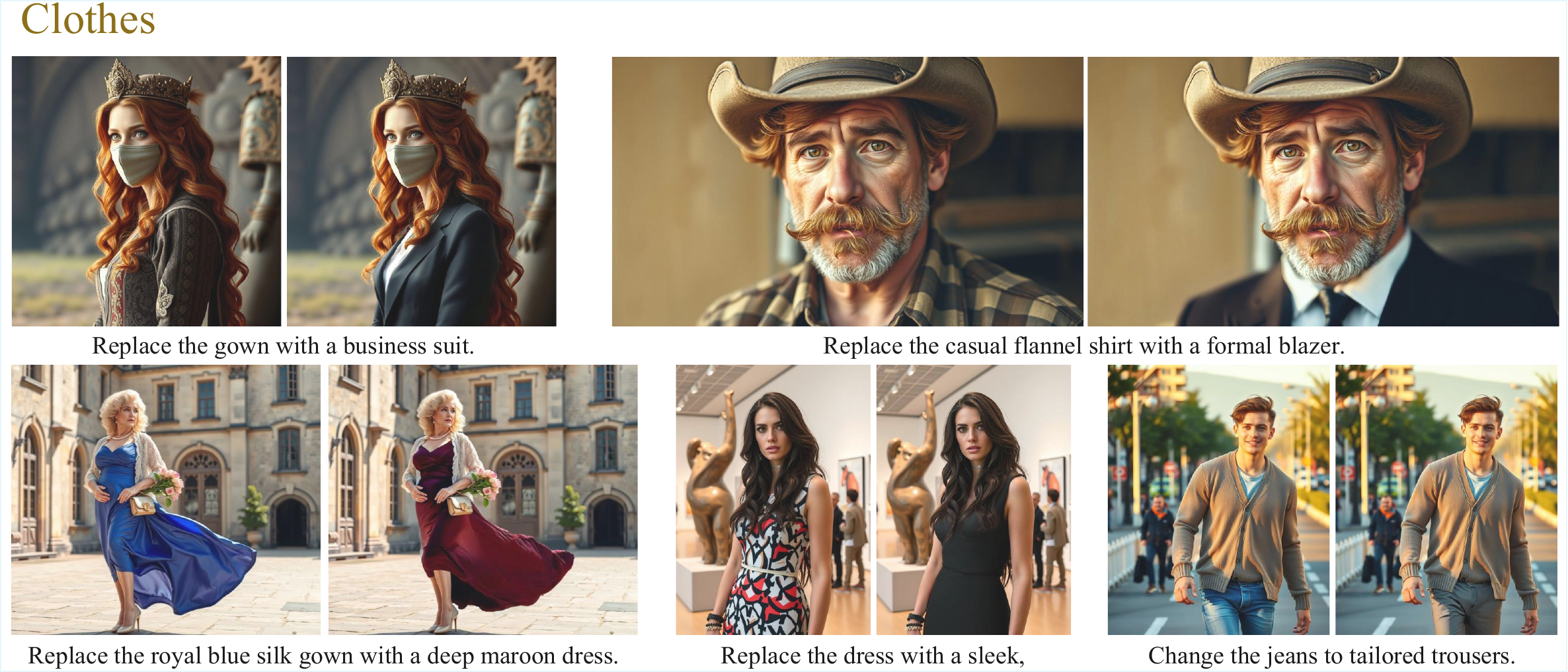}
  \caption{Edits that require human clothes change.}
  \label{fig:collage9}
\end{figure*}
\begin{figure*}[!htbp]
  \centering
  \includegraphics[width=\linewidth]{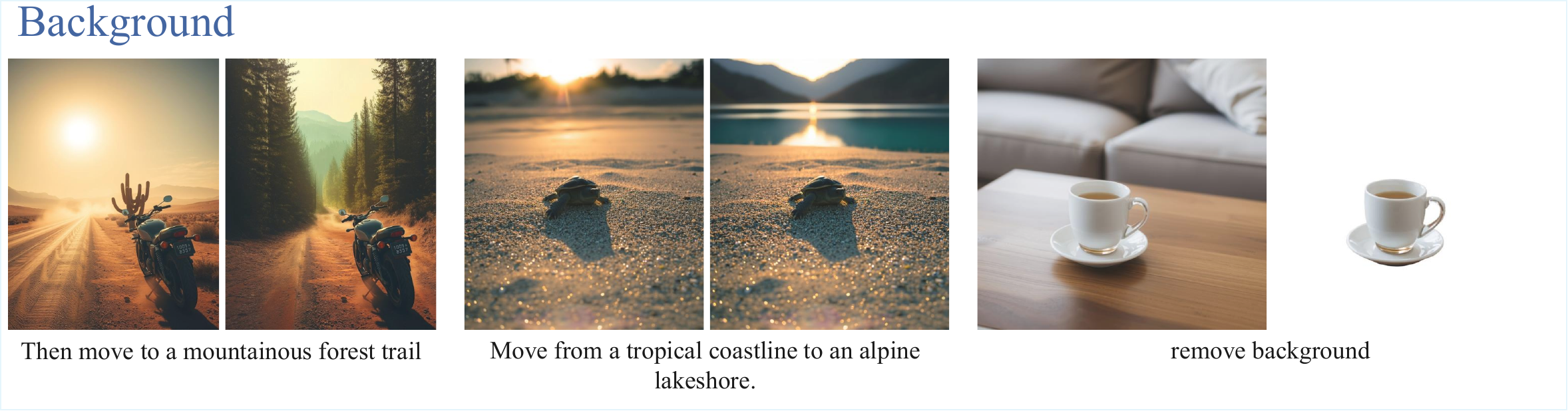}
  \caption{Background manipulations.}
  \label{fig:collage10}
\end{figure*}
\begin{figure*}[!htbp]
  \centering
  \includegraphics[width=\linewidth]{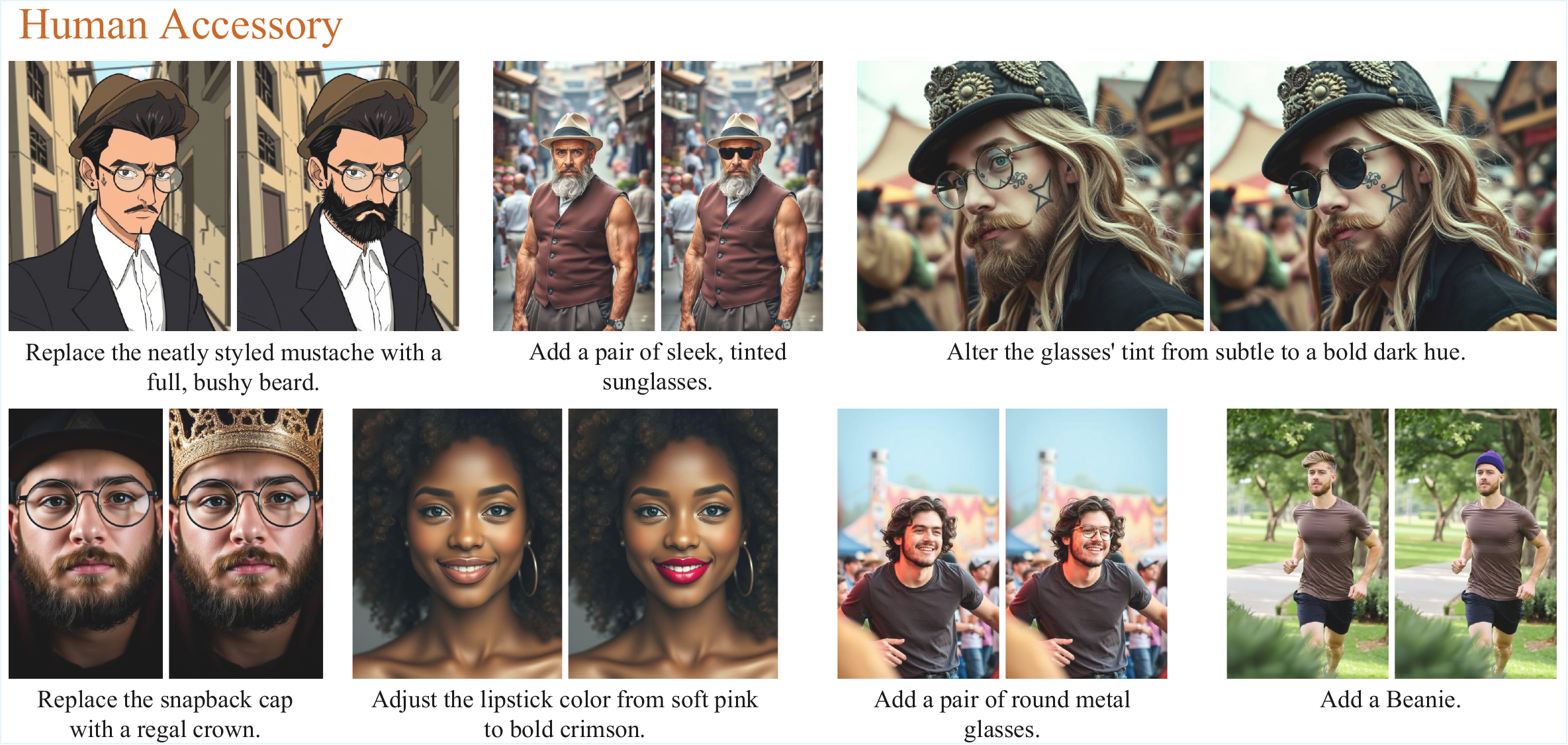}
  \caption{Changing accessories and adding new features to human appearance.}
  \label{fig:collage11}
\end{figure*}
\begin{figure*}[!htbp]
  \centering
  \includegraphics[width=\linewidth]{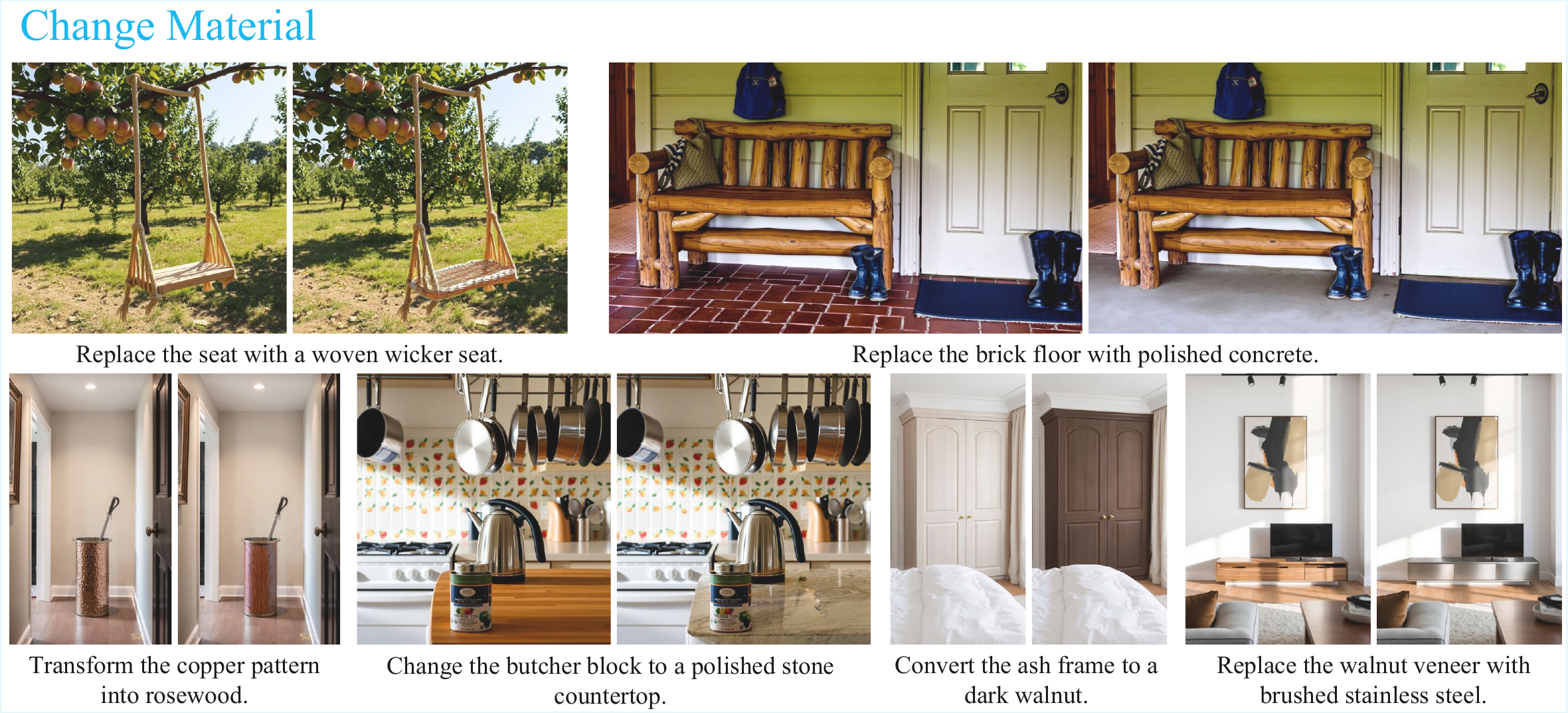}
  \caption{Material change showcases.}
  \label{fig:collage8}
\end{figure*}

\begin{table*}[!htbp]
    \centering
    \renewcommand{\arraystretch}{1.2}
    \setlength{\tabcolsep}{4pt}
    \caption{Example failure cases from the ablation study.}
    \begin{tabular}{
        >{\raggedright\arraybackslash}m{2.5cm} 
        >{\raggedright\arraybackslash}m{4.2cm} 
        >{\centering\arraybackslash}m{1.5cm} 
        >{\centering\arraybackslash}m{5.3cm}
    }
        \toprule
        \textbf{Shortcomings} & \makecell{\textbf{Explanation /}\\\textbf{Failure Mode}} & \makecell{\textbf{Inclusions}\\\textbf{found (300)}} & \textbf{Examples} \\
        \midrule
        
        \textbf{Initial image shortcomings} & The pipeline filters may occasionally miss problems in the original images, e.g., in scenes with dynamic human poses. & 15 & 
        \includegraphics[width=5.2cm]{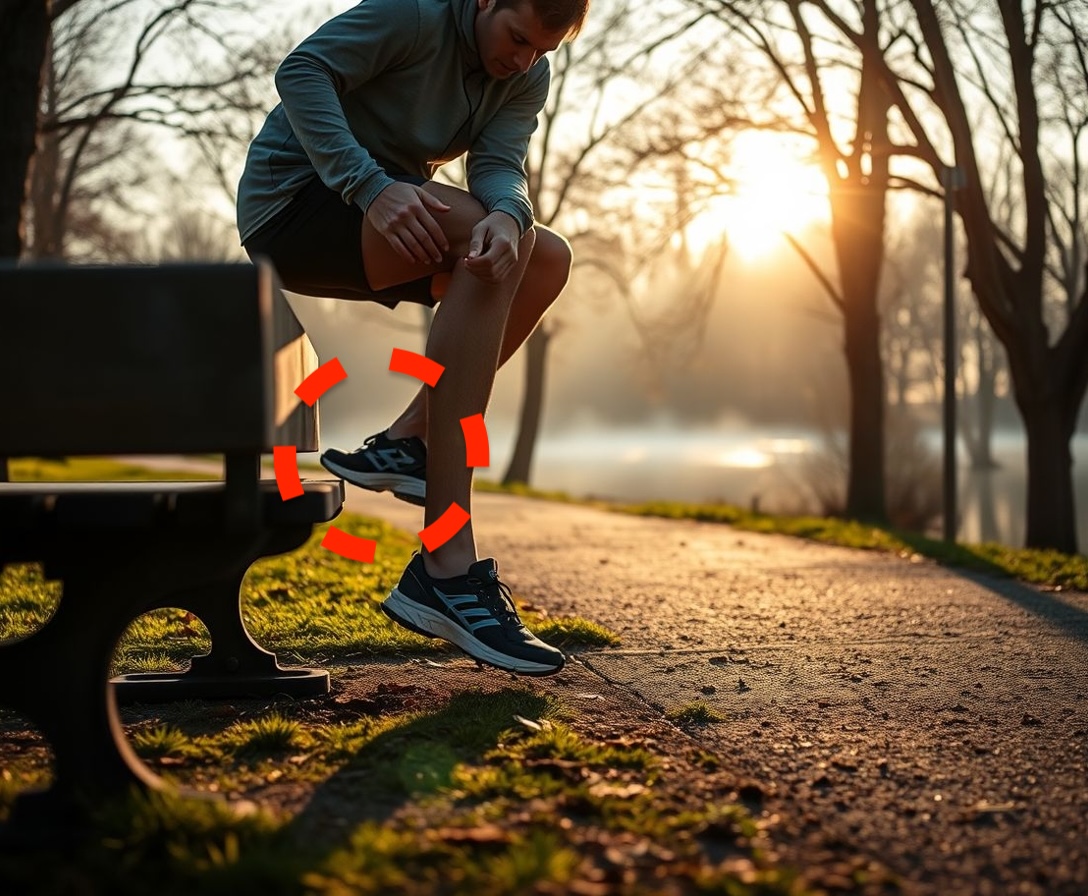} \\
        \midrule
        
        \textbf{Shadows, reflections, lighting} & Although the system usually removes or adds these effects correctly, some sophisticated (esp. lighting-related) cases remain challenging. & 13 &
        \begin{tabular}[c]{@{}c@{}}
            \includegraphics[width=2.5cm]{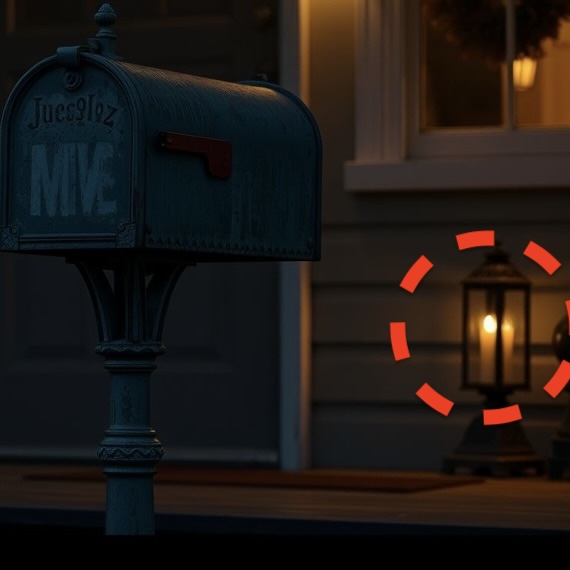}
            \includegraphics[width=2.5cm]{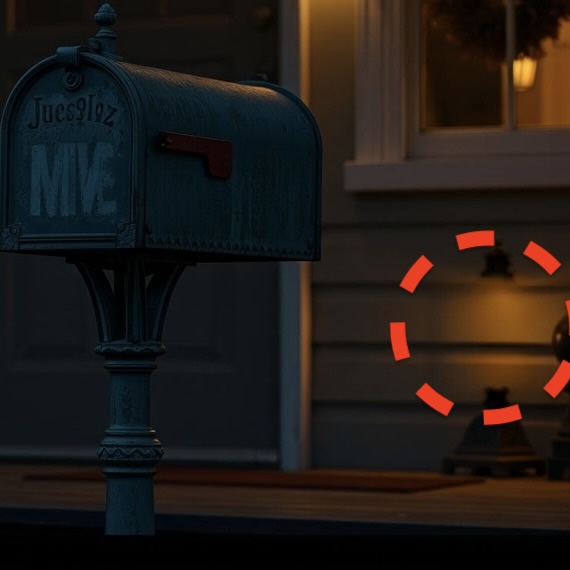}\\
            {Remove the flickering lantern} \\
            \addlinespace[6pt]
            \includegraphics[width=2.5cm]{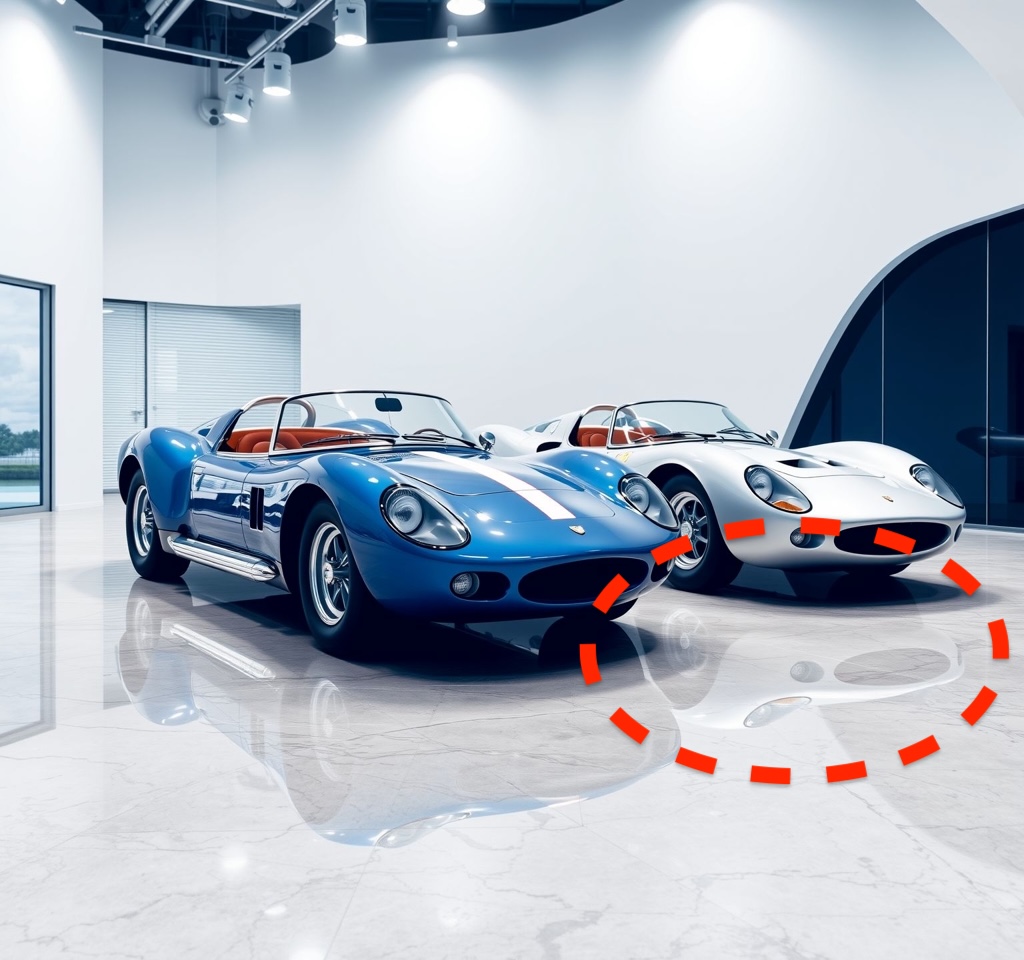}
            \includegraphics[width=2.5cm]{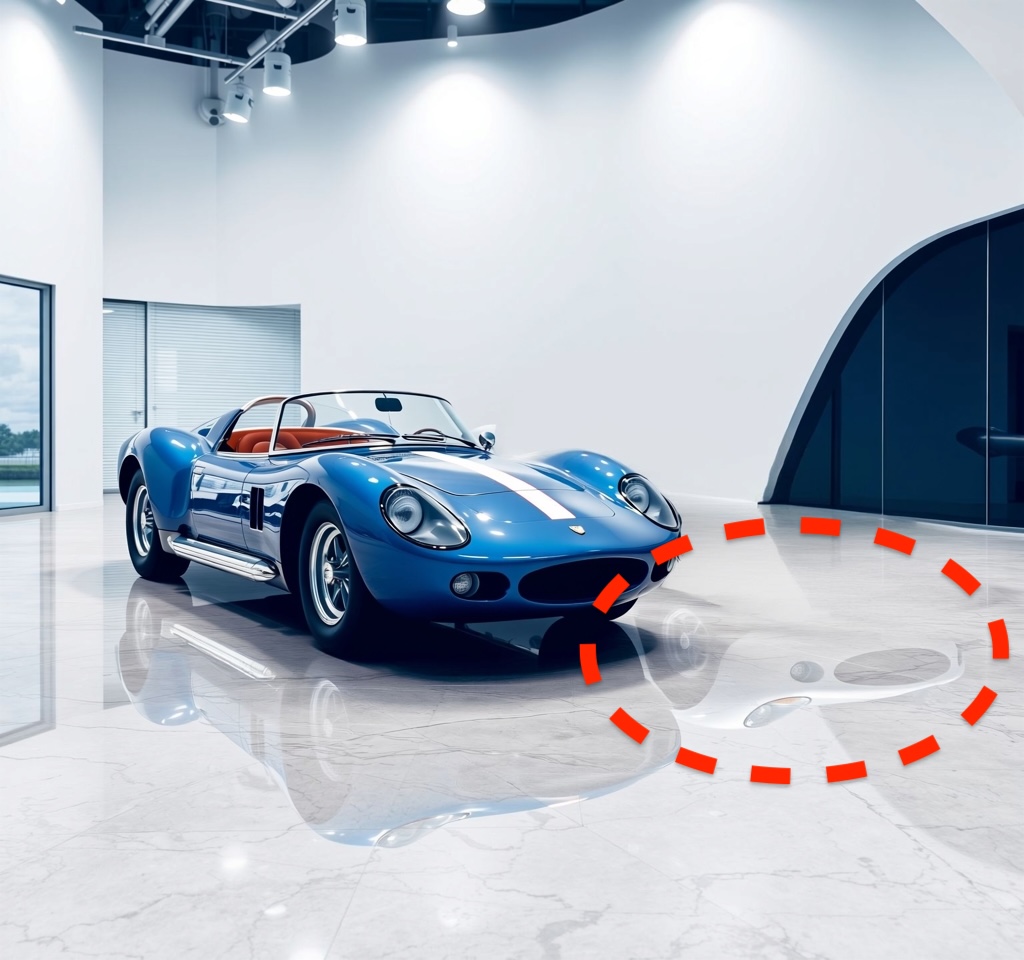}\\
            {Remove the car in the background}
        \end{tabular} \\
        \midrule
        
        \textbf{Target region detection} & Edits may \textit{over-affect} or \textit{under-affect} the image (e.g., failing to remove occluded object parts). & 10 & 
        \begin{tabular}[c]{@{}c@{}}
            \includegraphics[width=2.5cm]{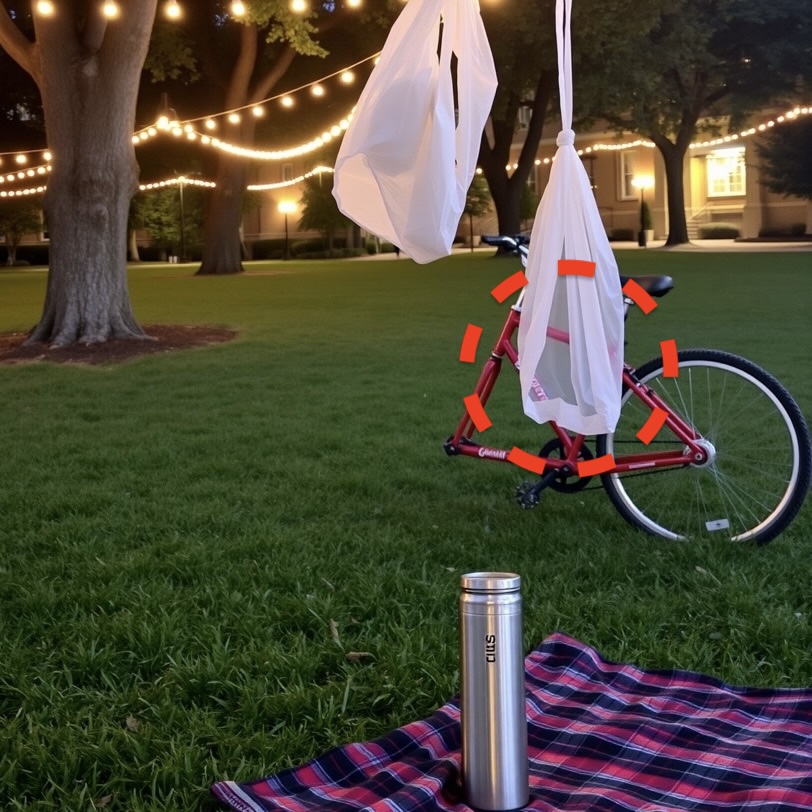}
            \includegraphics[width=2.5cm]{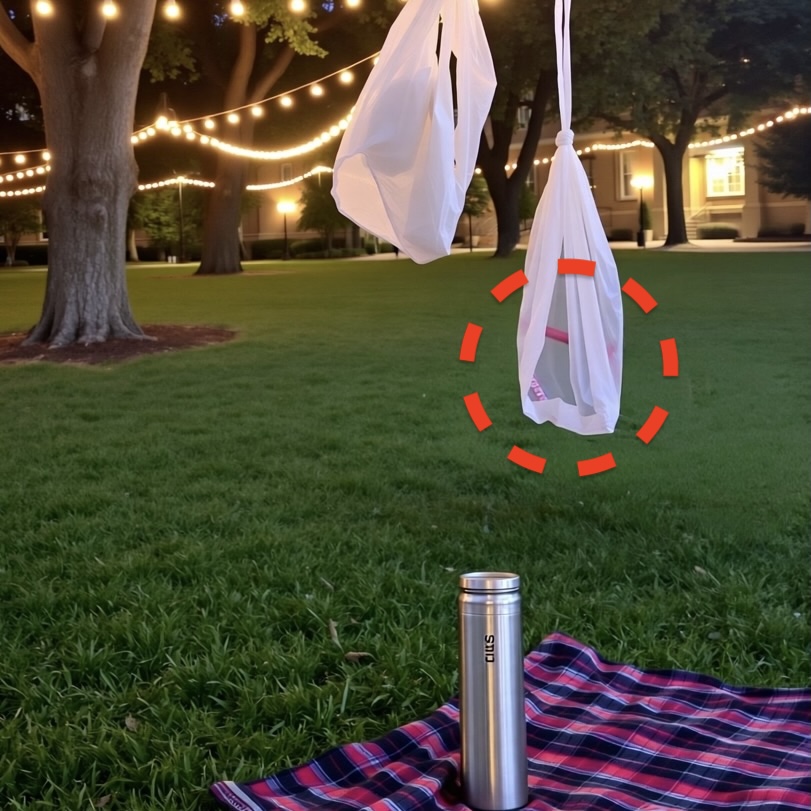}\\
            {Remove the red folding bike}
        \end{tabular} \\
        \midrule
        
        \textbf{Other issues} & Occasional errors such as imperfect inpainting after object removal. & 5 & 
        \begin{tabular}[c]{@{}c@{}}
            \includegraphics[width=2.5cm]{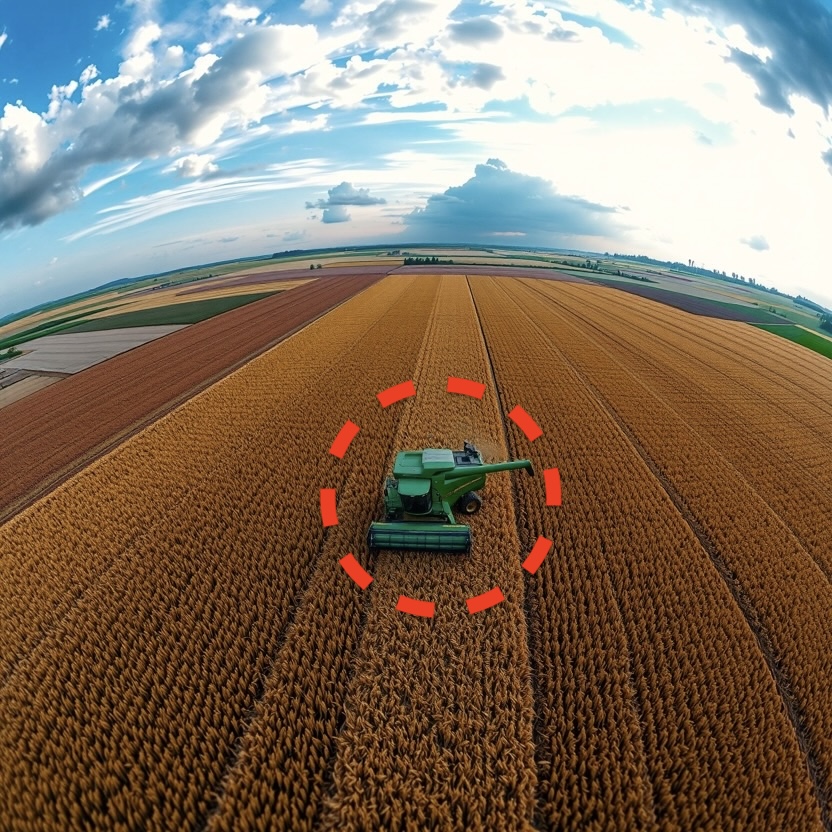}
            \includegraphics[width=2.5cm]{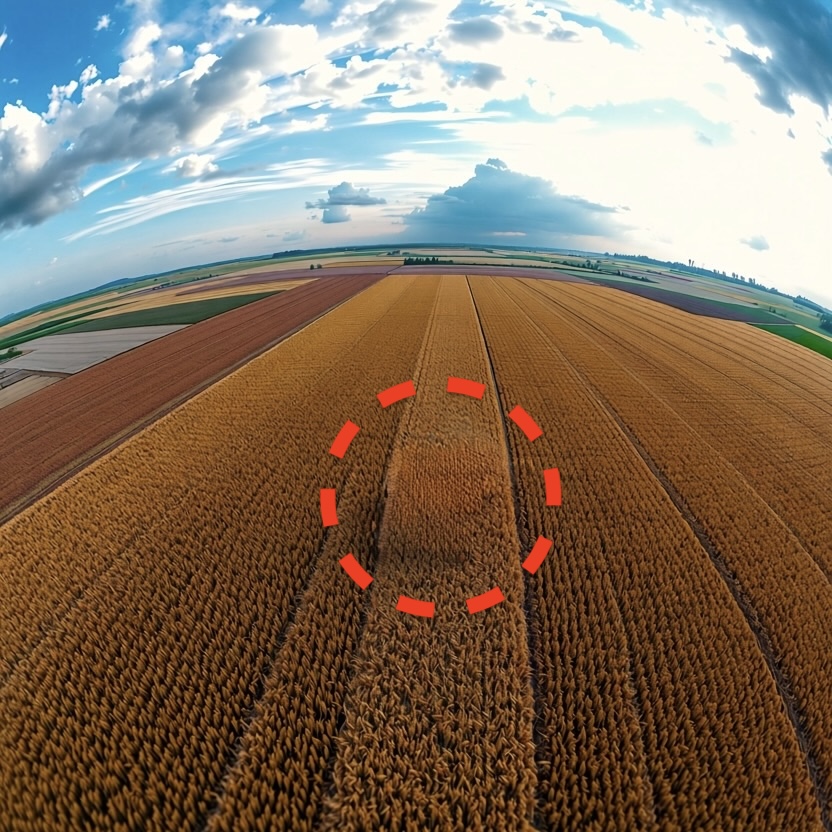}\\
            {Remove the combine harvester}
        \end{tabular} \\
        \bottomrule
        
    \end{tabular}
    \label{tab:appendix_humanaudit}
\end{table*}

\end{document}